\documentclass[a4paper,fleqn]{cas-sc}

\newif\ifshowauthors
\showauthorstrue   

\usepackage[numbers]{natbib}

\def\tsc#1{\csdef{#1}{\textsc{\lowercase{#1}}\xspace}}
\tsc{WGM}
\tsc{QE}

\usepackage{times}
\usepackage{latexsym}

\usepackage[T1]{fontenc}

\usepackage[utf8]{inputenc}

\usepackage{microtype}

\usepackage{inconsolata}

\usepackage{graphicx}

\usepackage{tikz}
\usetikzlibrary{arrows.meta,positioning,fit,backgrounds,shapes.geometric}
\usepackage{amsmath}
\usepackage{graphicx}
\usepackage{amsmath}
\usepackage{amssymb}
\usepackage{tcolorbox}
\usepackage{listings}
\usepackage{float}
\usepackage{algorithm}
\usepackage{algpseudocode}
\usepackage[section]{placeins}
\usepackage{capt-of}
\usepackage{url}
\usepackage{booktabs}
\usepackage{multirow}
\usepackage{amsfonts}
\usepackage{array}
\usetikzlibrary{arrows.meta}
\usetikzlibrary{positioning,fit,backgrounds,shapes.geometric,arrows.meta}

\usepackage{makecell}

\usepackage{xcolor}
\usepackage{pgfplots}
\usepgfplotslibrary{groupplots}
\pgfplotsset{compat=1.18}

\def\UpPct#1{{\scriptsize$\uparrow$\,#1\%}}

\usepackage{soul}
\tcbuselibrary{breakable}
\newif\ifREVISIONHIGHLIGHT
\REVISIONHIGHLIGHTfalse
\definecolor{revyellow}{RGB}{255,242,153}
\ifREVISIONHIGHLIGHT
  \newcommand{\rev}[1]{{\sethlcolor{revyellow}\hl{#1}}}
  \newcommand{\revmath}[1]{\colorbox{revyellow}{$\displaystyle #1$}}
  \newenvironment{revblock}%
    {\begin{tcolorbox}[colback=revyellow,colframe=revyellow!70!black,
       boxrule=0.3pt,left=2pt,right=2pt,top=2pt,bottom=2pt,breakable]}%
    {\end{tcolorbox}}
  \newenvironment{revtable}%
    {\begin{tcolorbox}[colback=revyellow,colframe=revyellow!70!black,
       boxrule=0.3pt,left=2pt,right=2pt,top=2pt,bottom=2pt]}%
    {\end{tcolorbox}}
\else
  \newcommand{\rev}[1]{#1}
  \newcommand{\revmath}[1]{#1}
  \newenvironment{revblock}{}{}
  \newenvironment{revtable}{}{}
\fi

\makeatletter
\AtEndDocument{%
  \addtocounter{page}{-1}%
  \immediate\write\@auxout{\string\csxdef{lastpage}{\thepage}}%
  \addtocounter{page}{1}%
}
\makeatother

\ExplSyntaxOn
\cs_set:Npn \__first_footerline:
{
  \group_begin:
  \small\sffamily
  \ifnum\theblind>0\relax
  \else
    \__short_authors: :~
  \fi
  { \rmfamily\itshape Published~in~Knowledge-Based~Systems }
  \group_end:
}
\ExplSyntaxOff

\AddToHookNext{shipout/foreground}{%
  \begin{tikzpicture}[remember picture,overlay]
    \node[anchor=south west,inner sep=0pt,text width=\textwidth,
          font=\footnotesize,align=left]
      at ([xshift=\dimexpr1in+\oddsidemargin\relax,yshift=22mm]current page.south west)
      {\textcopyright\ 2026. This manuscript version is made available under the CC BY-NC-ND 4.0 license\\
       \url{https://creativecommons.org/licenses/by-nc-nd/4.0/}\\
       The Version of Record is available at \url{https://doi.org/10.1016/j.knosys.2026.117071}};
  \end{tikzpicture}%
}

\begin{document}
\let\WriteBookmarks\relax
\raggedbottom
\renewcommand{\topfraction}{0.9}
\renewcommand{\bottomfraction}{0.8}
\renewcommand{\textfraction}{0.08}
\renewcommand{\floatpagefraction}{0.75}
\renewcommand{\dbltopfraction}{0.9}
\renewcommand{\dblfloatpagefraction}{0.75}
\setcounter{topnumber}{3}
\setcounter{bottomnumber}{2}
\setcounter{totalnumber}{5}

\shorttitle{Sensory-Aware Sequential Recommendation}

\ifshowauthors
\shortauthors{Y.-C. Yoon, C. Park and K. Koh}
\fi

\title{Sensory-Aware Sequential Recommendation via Review-Distilled Representations}



%

\ifshowauthors

\author[1]{Yeo-Chan Yoon}[orcid=0000-0002-5573-8964]

\fnmark[1]
\cormark[1]

\ead{ycyoon@jejunu.ac.kr}
\cortext[1]{Corresponding author}

\credit{Conceptualization of this study, Writing - review \& editing, Writing - original draft, Investigation, Methodology, Supervision}

\author[2]{Chanjun Park}

\fnmark[1]

\credit{Writing - review \& editing, Investigation}

\author[3]{Kyuhan Koh}[]
\credit{Writing - review \& editing}

\fi

\ifshowauthors

\affiliation[1]{
  organization={Jeju National University},
  city={Jeju},
  state={Jeju},
  country={Korea}
}

\affiliation[2]{
  organization={Soongsil University},
  city={Seoul},
  country={Korea}
}

\affiliation[3]{
  organization={California State University Stanislaus},
  city={Turlock},
  country={USA}
}

\fntext[1]{Yeo-Chan Yoon and Chanjun Park contributed equally to this research.}

\fi


\begin{abstract}
\rev{Sequential recommenders learn behavioral patterns from item identifiers, while the experiential properties that users describe in reviews, such as how products look, feel, smell, taste, or sound, rarely enter item representations in a controlled, auditable form.}
\rev{We present \mbox{\textsc{ASER}} (Attribute-based Sensory-Enhanced Representation), an offline pipeline that fine-tunes a large language model to extract evidence-grounded sensory attribute--value records, such as \textit{color: matte black} or \textit{scent: vanilla}, from review text and distills them into a compact student encoder that produces a frozen five-facet sensory bank for each item catalog.}
\rev{At recommendation time the pretrained backbone stays frozen: a lightweight relational metric between the user history and each candidate is learned over the bank, and its correction is applied within a validation-selected magnitude bound.}
\rev{Across five Amazon domains and four backbones, trained within a common experimental pipeline and evaluated by full-catalog leave-one-out ranking without sampled negatives, this integration improves HR@10 and NDCG@10 in all 20 domain--backbone pairs, with average relative gains of 6.1\% and 6.4\%.}
\rev{A matched non-sensory control channel, built with the same seed model, schema, and pipeline, separates the sources of the gain: the hit-rate improvement follows from structured, evidence-grounded extraction as such, whereas the sensory vocabulary yields a ranking-quality advantage in eight of nine matched comparisons.}
\rev{An audit of the Beauty evaluation catalog finds that 94.8\% of retained records are supported by their cited evidence spans, so the extracted signal remains inspectable against its source text.}
\end{abstract}


\begin{itemize}

\item Enriches sequential recommendation with review-distilled sensory attributes.

\item Distills LLM-extracted sensory records into compact item embeddings.

\item \rev{Improves HR@10 and NDCG@10 in all 20 domain--backbone settings.}

\item \rev{Structured extraction lifts hit rate; sensory content lifts ranking quality.}

\item \rev{Links recommendations to interpretable, human-perceivable descriptors.}

\end{itemize}


\maketitle

\section{Introduction}

Users often describe products in terms of how they \emph{look}, \emph{feel}, \emph{taste}, \emph{smell}, or \emph{sound} when writing reviews. These sensory descriptions, such as \rev{the smooth texture of a yoga mat or the vanilla scent of a candle}, convey fine-grained experiential information that is rarely captured by standard sequential recommender systems. Although modern sequence models achieve strong performance using item identifiers and interaction histories, they typically treat items as atomic symbols and overlook the rich semantic content embedded in review text.

The primary goal of this work is to enhance sequential recommendation by introducing a linguistically grounded representation layer derived from sensory information in user reviews. Instead of relying solely on behavioral signals, we construct item representations that explicitly encode experiential attributes such as color, texture, scent, and sound. \rev{We separate two claims that prior work tends to conflate. The first is that converting review text into structured, evidence-grounded attribute records, rather than into unconstrained text embeddings, is what brings additional correct items into the top-\mbox{$K$} list; a matched control channel built with the same pipeline over a non-sensory vocabulary reproduces this gain, so it is a property of the structuring and not of the sensory vocabulary. The second is that sensory content shows a consistent advantage in ordering items near the top of the list, an effect that the same control does not reproduce consistently. This paper makes the first claim for structured extraction in general and the second for the sensory channel in particular.} Sensory extraction is not treated as an end task in itself, but as a representation learning mechanism that transforms unstructured natural language into structured signals suitable for integration into sequence models.

To this end, we propose \textsc{ASER} (Attribute-based Sensory-Enhanced Representation), a framework that combines \rev{large-language-model-based} attribute extraction with knowledge distillation and sequential recommendation. We first fine-tune a teacher language model to extract structured sensory attribute--value pairs from product descriptions and reviews. These structured outputs provide supervision for training a compact student transformer that produces fixed-dimensional sensory embeddings for each item. The resulting embeddings encode distilled sensory semantics in a form that can be efficiently incorporated into downstream models.

We integrate the learned sensory embeddings into representative sequential recommendation architectures, including SASRec \citep{kang2018sasrec}, BERT4Rec \citep{sun2019bert4rec}, BSARec \citep{shin2024bsarec}\rev{,} and DIFF \citep{kim2025diff}, and evaluate them on five Amazon domains: Beauty, Sports, Toys, Grocery\rev{,} and Video Games. \rev{However, integrating review-derived sensory information into sequential recommendation is challenging, because sensory representations and collaborative representations capture different aspects of item semantics: directly combining them assumes that the two representations are aligned and that sensory information contributes consistently across users. To address this, we model sensory information as a relational signal rather than as a replacement for collaborative representations.} \rev{Integration is bounded and relational: the pretrained backbone is frozen, only the sensory relation between the user history and the candidate item is learned, and the correction it produces is applied with a magnitude bound whose strength and candidate depth are selected on validation data.} \rev{All 20 domain--backbone cells are trained and evaluated within one common experimental pipeline under the standard leave-one-out protocol: for each user, the last interaction is the test target and the second-to-last interaction is the validation target, and every model ranks the complete item catalog of its domain, with previously consumed items masked and without sampled negatives at evaluation time.} \rev{Under this protocol, sensory augmentation improves next-item prediction in all 20 domain--backbone combinations for both HR@10 and NDCG@10, and in all 120 reported metric cells.}

\rev{Beyond quantitative improvements, the extracted attributes keep the item-side signal auditable: every bank entry derives from evidence-grounded records whose cited spans can be checked against the source text, and Section~\mbox{\ref{sec:recall_audit}} audits both the precision and the recall of this evidence.}

Overall, this work demonstrates that \rev{structured, evidence-grounded attribute extraction provides a principled and scalable representation learning strategy for sequential recommendation, while the sensory channel within that family shows a consistent advantage in ranking quality rather than in hit rate alone.} By translating unstructured review text into structured sensory embeddings, we connect information extraction with recommender modeling and \rev{show both what linguistically grounded semantics contribute to sequence-based recommendation and the limits of that contribution.}

\section{Related Work}

\subsection{Sequential Recommendation with ID Signals and Language Signals}
Sequential recommendation models predict the next item from a temporally ordered interaction history. Early neural approaches used recurrent networks such as GRU4Rec, which improved over Markov baselines by modeling long-range dependencies in user sequences \citep{hidasi2015grurec}. Transformer-based models later became dominant because self-attention can represent both short-term intent and long-term preference more flexibly. SASRec models user histories with causal self-attention \citep{kang2018sasrec}, while BERT4Rec uses a masked item prediction objective to exploit bidirectional context \citep{sun2019bert4rec}. Recent work also explores architectural inductive biases and intention modeling beyond vanilla self-attention. BSARec, for example, combines complementary sequence modeling components and shows strong performance as a general backbone \citep{shin2024bsarec}, while AuriSRec explicitly models user intention with adversarial learning to separate current intent from historical behavior \citep{zhang2024aurisrec}. A common limitation of these ID-centric models is that they primarily learn from item identifiers and co-occurrence patterns, which makes it difficult to leverage review semantics that describe product experiences. ASER addresses this limitation by adding a compact sensory representation layer that can be \rev{used alongside these backbones through relational reranking}, without replacing the core sequential encoder.

Recent journal work further shows that sequential recommendation is moving beyond simple ID-sequence encoding toward efficient sequence modeling, temporal calibration, and robust augmentation. State-space and Mamba-style architectures improve the scalability of long behavior modeling through structured state-space duality and time-aware sequence processing \citep{qu2026ssd4rec,fan2025tim4rec}. Hybrid temporal architectures further combine local and global dependency modeling with sparse attention to capture both short-term and long-term interests efficiently \citep{song2026heterotemprec}. Diffusion and contrastive learning have also been used to strengthen sequential representations under sparse or noisy interaction patterns \citep{zhu2026jclrec}. These studies advance the sequential encoder itself. ASER is complementary: it does not redesign the sequence backbone, but instead constructs a reusable item-side sensory channel from unstructured reviews and \rev{applies this channel through a common relational interface across} multiple existing backbones.

A parallel line of work incorporates language signals into \rev{both recommendation in general and sequential recommendation in particular}. Earlier text-aware recommenders encode reviews or item descriptions and combine the resulting representations with collaborative signals \citep{mcauley2013hft,zheng2017deepconn,chen2018narre}. More recent NLP-oriented recommendation work leverages pretrained language models for explanation generation, personalized review generation, and aspect-aware retrieval, as in PETER, ERRA, XRec, and MAPLE \citep{li2021personalized,cheng2023explainable,ma2024xrec,yang2025maple}. These studies show that textual evidence can improve personalization and explanation quality, but they typically optimize generation or explanation faithfulness rather than learning a small, reusable item-side representation that can be cleanly fused into multiple sequential backbones. A common limitation is that these representations are trained for recommendation or explanation objectives without explicit constraints on what the text encoder should capture. As a result, the learned vectors can mix sensory cues with brand identifiers, compatibility statements, and general opinions, which makes the linguistic signal \rev{difficult to control and analyze}.

From the NLP side, structured information extraction has increasingly moved toward schema-conditioned text-to-structure generation. UIE unifies heterogeneous extraction tasks under a schema-guided generation framework \citep{lu2022unified}, and recent LLM-based IE work studies how to align extractors to schema-constrained outputs and how to perform schema-driven extraction over heterogeneous inputs \citep{qi2024adelie,bai2024schema}. ASER is closely related to this trend, but differs in two important ways. First, it narrows the target space to a sensory-only schema tailored to recommendation rather than attempting universal extraction. Second, each record isolates a fine-grained sensory attribute and an open-vocabulary value \rev{and provides} an evidence span that supports the extraction. This design makes the injected signal more interpretable and more controllable, and it enables systematic ablations by selectively injecting or removing specific sensory facets rather than relying on an entangled text embedding.

Work using LLMs for recommendation has expanded rapidly in 2024--2026. SeRALM augments sequential recommenders with knowledge produced by aligned large language models, which can improve performance but makes it hard to control which semantics are injected and can increase dependence on expensive LLM inference \citep{wang2024seralm}. LRD induces latent relations between items using a large language model, but the discovered relations are not guaranteed to be interpretable as stable item properties, and they are not grounded to explicit evidence spans in text \citep{yang2024lrd}. IDGenRec converts item identifiers into textual identifiers to better align with language models, but the resulting representations remain indirect proxies for item semantics and do not explicitly isolate experiential dimensions \citep{tan2024idgenrec}. ReLRec reconstructs pseudo labels from reviews for LLM-enhanced recommendation, but a reconstruction target can still emphasize non-experiential signals and does not guarantee a focused attribute space \citep{na2024relrec}. LLM2Rec studies large language models as embedding generators for recommendation, yet the embedding space is generally not constrained to a specific semantic subspace such as sensory experience, which limits interpretability and targeted fusion \citep{he2025llm2rec}. PatchRec improves efficiency through patch training of LLM recommenders, but the approach still treats the LLM as a central component rather than distilling a small controllable representation layer \citep{liao2025patchrec}. Distillation Matters and SLMRec focus on distilling LLM-level recommendation capability into smaller models, but the distilled target is typically a ranking behavior or teacher scores rather than an explicit linguistic attribute layer \citep{cui2024distillation,xu2025slmrec}. In parallel, recent papers increasingly formulate sequential recommendation itself as a language-centric or controllable reasoning problem. RecGPT proposes a foundation-model formulation for sequential recommendation based on textual item representations \citep{jiang2025recgpt}; Bi-Tuning and AGRec improve controllability and structural reasoning in LLM-based sequential recommendation \citep{zhang2025bituning,wang2025agrec}; and RDRec distills LLM-generated rationales into a compact recommender that can benefit both top-$N$ and sequential recommendation \citep{wang2024rdrec}. ASER is complementary to these lines by using LLMs primarily as scalable annotators for a constrained sensory schema and then distilling the output into a compact student that produces fixed sensory embeddings for downstream sequential models.

Recent journal studies also incorporate language-model semantics into recommender architectures from different angles. Some work evaluates how LLMs can improve sequential recommendations when used for \rev{reranking} or sequence understanding \citep{boz2025improvingllms}, while instruction-following formulations translate user preference, intention, and task context into natural-language prompts for recommendation \citep{zhang2025instructionrec}. Retrieval-enhanced LLM-based sequential recommendation further studies how to select collaborative and temporal evidence from long behavior histories before invoking an LLM \citep{hu2026react}. \rev{Collaborative signals can also flow in the opposite direction: CEILLM infuses graph-based collaborative embeddings into a large language model through lightweight adapters and combines the two signal families with a confidence-aware dynamic fusion gate for conversational recommendation \mbox{\citep{yoon2026ceillm}}.} These approaches demonstrate that LLM-derived semantics can enrich recommendation, but they typically use LLMs as a central inference, ranking, or behavior-comprehension component. ASER instead uses an LLM only as an offline annotator for a constrained sensory schema; downstream training and inference consume a compact student encoder and frozen item-level sensory bank without invoking an LLM.

\subsection{Attribute and Emotion Signals from Reviews, and the Sensory Gap}
Attribute-level modeling has long been used to make recommendation more semantically meaningful. HFT aligns latent factors with topics derived from reviews, but topics are coarse and can mix experiential cues with non-sensory content \citep{mcauley2013hft}. DeepCoNN and NARRE encode reviews with neural architectures, yet their representations remain opaque mixtures and often require heavy per-user and per-item review aggregation \citep{zheng2017deepconn,chen2018narre}. Explainable recommendation research further highlights the usefulness of aspect-aware text modeling, but models such as PETER, ERRA, XRec, and MAPLE are primarily designed to generate personalized explanations or reviews rather than to produce an auditable, reusable item-side sensory channel for sequential encoders \citep{li2021personalized,cheng2023explainable,ma2024xrec,yang2025maple}.

\rev{Side information has a long line of work in sequential recommendation. FDSA adds a feature-level self-attention block alongside the item-level one \mbox{\citep{zhang2019fdsa}}, S\textsuperscript{3}-Rec pretrains with mutual-information objectives that tie items to their attributes \mbox{\citep{zhou2020s3rec}}, NOVA keeps attribute signals out of the value path so they cannot distort the item representation \mbox{\citep{liu2021nova}}, and DIF-SR decouples attribute and identifier attention entirely \mbox{\citep{xie2022difsr}}; DIFF \mbox{\citep{kim2025diff}}, one of our backbones, continues this line. A parallel family replaces identifiers with text representations for transfer: UniSRec learns universal item representations from descriptions \mbox{\citep{hou2022unisrec}} and RecFormer models sequences directly as language \mbox{\citep{li2023recformer}}, while a systematic comparison finds identifier- and modality-based models competitive under matched settings rather than one dominating \mbox{\citep{yuan2023idmodality}}. These approaches take attributes as given, from catalog metadata or from unconstrained text; the question we ask is what happens when the item-side signal is a narrow, evidence-grounded slice of review language that can be audited, and it is applied without modifying the backbone at all.}

More recent approaches aim to move beyond generic text encoders by introducing explicit attribute signals for sequential recommendation. FineRec extracts fine-grained attribute--opinion pairs from reviews and reports consistent gains, but its extracted space is intentionally broad and unconstrained \citep{zhang2024finerec}. As a result, the mined attributes mix sensory cues with heterogeneous factors such as durability, compatibility, shipping condition, or vague quality statements, which vary widely across categories and are often weakly tied to \rev{stable item properties}. This makes the attribute channel hard to calibrate, difficult to filter reliably, and sensitive to domain-specific noise when injected into sequence models.

MARS learns attribute-aware matching representations, but it primarily treats attributes as auxiliary matching keys and does not provide an evidence-grounded, open-vocabulary attribute inventory that supports auditing, selective filtering, or error diagnosis \citep{chen2024mars}. In sequential recommendation, this matters because feature injection errors propagate across timesteps, and even a small fraction of noisy attributes can distort token representations under early fusion. Without explicit evidence spans, confidence, and negation handling, it is difficult to identify when and why an attribute signal harms performance. A related line models attributes as explicit relations to provide structure for sequential recommendation \citep{liu2025attribute}. These methods typically assume that reliable attributes already exist as metadata or curated taxonomies, and they do not address the central bottleneck in review-driven settings, namely how to extract consistent, comparable attributes from raw text at large scale while maintaining quality control.

Recent NLP work on fine-grained opinion extraction reinforces this concern. Aspect-sentiment triplet extraction models continue to improve the structured recovery of aspect, opinion, and polarity tuples \citep{naglik2024aste}, while Shoes-ACOSI shows that implicit opinion extraction in e-commerce reviews remains difficult even for modern LLMs and supervised systems \citep{peper2024shoes}. For recommendation, this matters because review signals are often indirect: users may imply a tactile or olfactory judgment without stating it in a canonical attribute form, or may mix experiential language with shipping or price complaints. ASER addresses this problem by restricting extraction to a sensory schema, grounding each extracted value to an explicit evidence span, and retaining polarity, negation, and confidence metadata so that noisy extractions can be filtered before \rev{downstream integration}.

Recent work on review- and aspect-based language signals is also relevant to review-derived sensory representation learning. Survey work on review-based recommender systems highlights the central role of textual reviews in extracting fine-grained preferences and item characteristics \citep{hasan2025reviewbased}, and review-centric recommendation models increasingly incorporate co-occurrence patterns, multimodal review signals, topic information, and implicit review relationships to improve personalization or explainability \citep{zhang2025beyondtexts,fang2025reviewsintegration,hao2025iregnn}. In parallel, LLM-based and graph-based aspect-level sentiment studies show the importance of data augmentation, confidence filtering, and structured aspect reasoning for noisy review text \citep{hellwig2025llmabsa,fan2026mda}. These studies reinforce the value of fine-grained review evidence and noise-aware language processing, but their targets are survey synthesis, conversational recommendation, multimodal/review integration, explainable recommendation, or aspect-level sentiment analysis rather than reusable item-level sensory features for sequential encoders. ASER adapts this principle to product reviews by restricting the output space to sensory attributes and using evidence-grounded teacher records as supervision for item-level sensory embeddings.

ASER further addresses these limitations by focusing on sensory descriptors as a deliberately constrained semantic slice of review text. Sensory language is both frequent and comparatively stable across product categories because it describes human-perceivable properties such as color, texture, scent, and sound that recur regardless of domain-specific functions or specifications. This makes sensory attributes a practical target for building a portable content channel that is less entangled with category-dependent jargon and less contaminated by logistical or transactional complaints. By enforcing a fixed sensory schema, extracting open-vocabulary values grounded by evidence spans, and attaching confidence and negation metadata, ASER produces a signal that can be audited and filtered before it is \rev{integrated with} sequential models. Distilling these structured outputs into compact sensory embeddings further reduces sparsity and enables efficient storage and \rev{relational integration}, while preserving the targeted semantics that broad attribute mining methods often struggle to isolate.

Emotion-aware recommendation is another active line, especially when user feedback is affective and domain dependent. Recent work in session-based news recommendation incorporates emotion-related signals to improve click prediction, but the emotion space is usually coarse and not tied to concrete item properties that persist across items and contexts \citep{du2025emotionnews}. In product reviews, sentiment and emotion can also be driven by shipping, price, or expectation mismatch rather than product experience. ASER reduces this confounding by targeting sensory descriptors that are closer to human-perceivable item properties and by attaching polarity and negation at the attribute level rather than at the whole-review level.

Despite substantial progress on attribute-aware and sentiment-aware recommendation, sensory semantics remain underexplored as a first-class representation space. Most prior attribute pipelines aim at functional aspects or generic opinions, and even recent LLM-enhanced methods typically do not constrain the extracted space to sensory-only descriptors with auditable evidence. ASER fills this gap by offering a scalable, controllable, and linguistically grounded sensory layer that can be \rev{integrated with} strong sequential recommenders such as SASRec, BERT4Rec, BSARec\rev{,} and DIFF without requiring the recommender itself to process raw text.

\section{Methodology}
\label{sec:method}

\begin{figure*}[t]
    \centering
    \includegraphics[width=\textwidth]{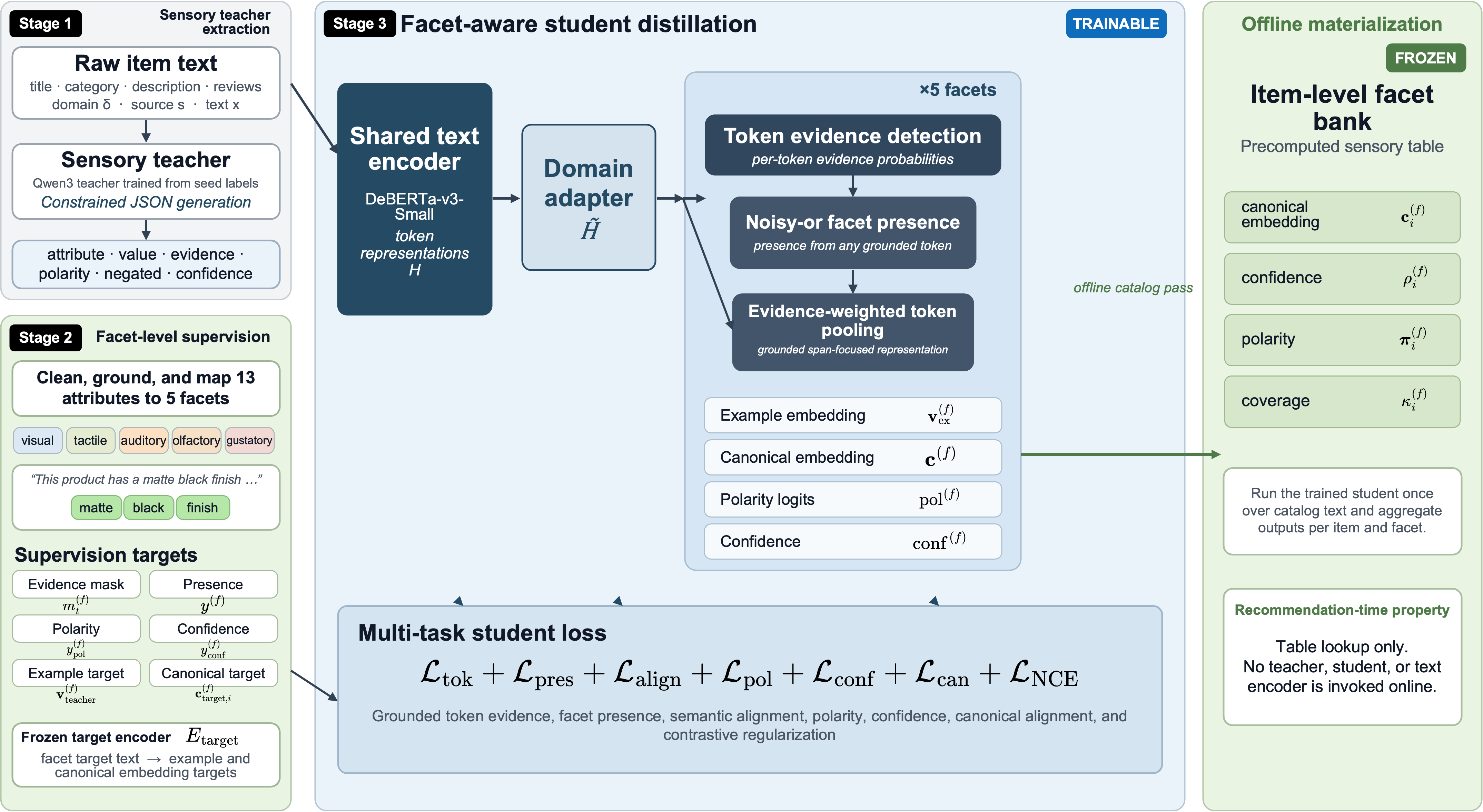}
    \caption[Offline stage of ASER]{\rev{%
    Offline stage of ASER: sensory extraction, facet supervision, and student distillation.
    Stage~1 applies a sensory teacher to raw item text and produces structured sensory JSON records containing attribute, value, evidence, polarity, negation, and confidence.
    Stage~2 converts the cleaned teacher outputs into facet-level supervision over five sensory facets, including token evidence masks, facet presence labels, polarity labels, confidence scores, example-level teacher embeddings, and canonical facet targets.
    Stage~3 trains a multi-facet sensory student.
    The student encodes text with DeBERTa-v3-Small, applies a per-domain bottleneck adapter, predicts token-level evidence for each facet, derives facet presence with noisy-or pooling, obtains grounded facet representations by evidence-weighted token pooling, and produces example embeddings, canonical embeddings, polarity logits, and confidence scores through facet-specific heads.
    The trained student is then run offline over catalog texts to construct an item-level facet bank containing canonical facet embeddings, confidence, polarity, and coverage statistics.
    All three stages run once, before recommender training; neither the teacher nor the student is invoked at recommendation time.}}
    \label{fig:offline-pipeline}
\end{figure*}

\begin{figure}[t]
    \centering
    \includegraphics[width=\columnwidth]{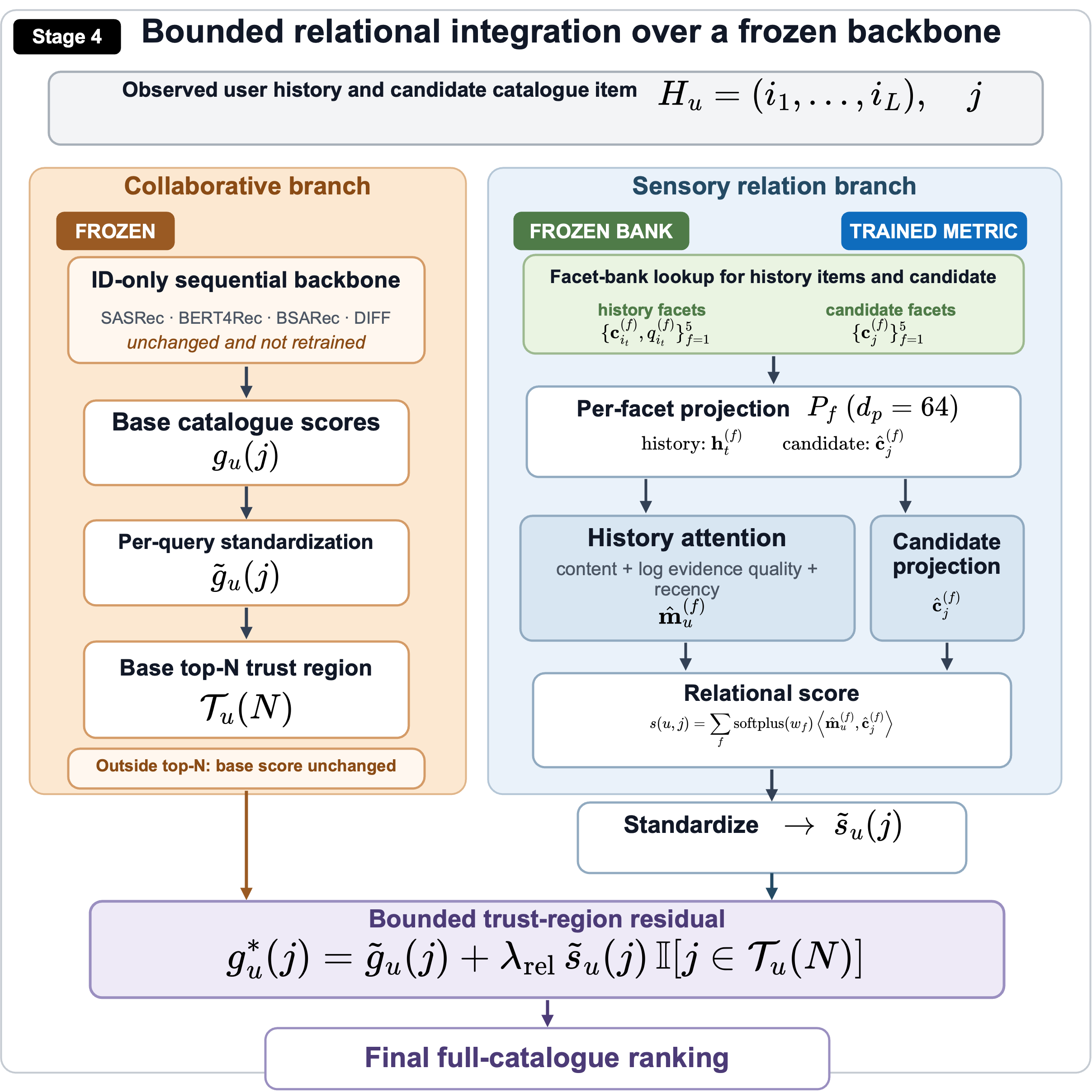}
    \caption[Runtime stage of ASER]{\rev{%
    Runtime stage of ASER (Stage~4): bounded trust-region reranking with the frozen facet bank.
The sequential backbone is frozen and scores the catalog exactly as its ID-only baseline does.
    In parallel, the precomputed item-level facet bank entry \mbox{$B_i$} is retrieved by item lookup for each history item, each stored canonical facet embedding is mapped by the learned per-facet projection \mbox{$P_f$} (Eq.~\mbox{\ref{eq:rel_candidate}}), and one attention pass per facet, biased by evidence quality and recency (Eqs.~\mbox{\ref{eq:runtime_quality_weight}}--\mbox{\ref{eq:rel_prototype}}), produces a per-facet user prototype.
    The relational score of a candidate is the facet-weighted sum of cosine agreements between prototype and candidate (Eq.~\mbox{\ref{eq:rel_score}}), which is bounded a priori by the learned facet weights.
    Base score and relational score are standardized per query and combined as a residual that is applied only inside the base top-\mbox{$N$} trust region (Eq.~\mbox{\ref{eq:rel_integration}}), so the residual is expressed in base-score standard deviations and cannot promote an item the backbone ranks below \mbox{$N$}.
    No language model or text encoder is invoked at recommendation time.}}
    \label{fig:online-integration}
\end{figure}

\rev{Figures~\mbox{\ref{fig:offline-pipeline}} and~\mbox{\ref{fig:online-integration}} summarize ASER, a sensory-aware sequential recommendation framework that separates sensory extraction, student distillation, and recommendation. Figure~\mbox{\ref{fig:offline-pipeline}} covers the offline pipeline (Stages~1--3) and Figure~\mbox{\ref{fig:online-integration}} the runtime integration (Stage~4).}
The teacher and student are used only in offline preprocessing.
At recommendation time, the model consumes item identifiers and a frozen item-level sensory table; neither the teacher LLM nor the text encoder is invoked during recommender training or inference.
This design allows the sensory pipeline to exploit rich catalog and review text offline, while preserving the efficiency and modularity of standard sequential recommendation backbones.

The methodology is organized into four global stages. Stage 1 extracts structured sensory records from item text
using a teacher language model. Stage 2 converts the cleaned teacher records into facet-level supervision. Stage 3
trains a compact multi-facet sensory student and materializes its outputs as an item-level facet bank. \rev{Stage 4 retrieves the frozen item-level facet bank at recommendation time, learns a sensory relation between the user history and the candidate item, and applies it as a bounded residual to the scores of a frozen sequential recommender.}

\begin{table}[htbp]
\centering
\footnotesize
\setlength{\tabcolsep}{4pt}
\renewcommand{\arraystretch}{1.06}
\begin{revtable}
\begin{tabular}{@{}ll@{}}
\toprule
Symbol & Meaning \\
\midrule
$\mathcal{F}$ & the five sensory facets (Table~\ref{tab:attr_facet_map}) \\
$\mathcal{C}$ & content-token mask (item text, excluding prefixes) \\
$\mathcal{T}_i$ & set of texts associated with item $i$ \\
$\mathcal{I}$, $\mathcal{U}$ & evaluated item catalog; evaluated user set \\
$L$ & encoder input sequence length; recommender history length \\
$d_h$ & text-encoder hidden size ($768$) \\
$d_b$ & per-domain adapter bottleneck width ($128$) \\
$D$ & stored facet-embedding dimension ($768$) \\
$d$ & recommender hidden size ($64$) \\
$d_p$ & per-facet sensory projection width ($64$) \\
$\delta$ & product-domain index \\
$\epsilon$ & small constant for numerical stability \\
$\epsilon_n$, $\epsilon_w$ & safe-normalization floor; evidence-mass threshold \\
$\tau$, $\tau_{\mathrm{pres}}$ & InfoNCE temperature ($0.07$); presence threshold ($0.5$) \\
$\lambda_f$ & per-facet loss weight \\
$w_{+}$, $w_{-}$ & presence-loss class weights ($1.0$, $0.35$) \\
$\theta$ & student parameters \\
$\sigma(\cdot)$ & logistic sigmoid \\
$\hat{\mathbf z}^{(f)}_i$, $\hat{\mathbf t}^{(f)}_i$ & L2-normalized student embedding; teacher target \\
$\omega^{(f)}_{ij}$, $\tilde\omega^{(f)}_{ij}$ & student-side and teacher-side aggregation weights \\
$E_{\mathrm{target}}$ & frozen target encoder used only for supervision \\
$q^{(f)}_i$ & item-level evidence-quality score of facet $f$ \\
$\lambda_{\mathrm{rel}}$, $N$ & residual weight and trust-region size (Stage~4) \\
\bottomrule
\end{tabular}
\end{revtable}
\caption[Notation]{\rev{Notation used throughout the paper.}}
\label{tab:notation}
\end{table}

\subsection{Stage 1: Sensory Teacher Extraction}
\label{sec:method_extraction}

We cast sensory attribute extraction as constrained structured generation.
For each item, we construct an input text by concatenating available catalog fields and review text.
The title, category, and description are placed first, followed by review texts separated by newline tokens.
This text is provided to the sensory teacher, which returns valid JSON under a fixed schema.

Each extracted record contains six fields:
\texttt{attribute}, \texttt{value}, \texttt{evidence}, \texttt{polarity}, \texttt{negated}, and \texttt{confidence}.
The \texttt{attribute} field is selected from a sensory inventory:
\texttt{color}, \texttt{pattern}, \texttt{shape}, \texttt{graphics}, \texttt{brightness}, \texttt{glossiness},
\texttt{transparency}, \texttt{finish}, \texttt{design}, \texttt{texture}, \texttt{scent}, \texttt{flavor}, and \texttt{sound}. \rev{Table~}\mbox{\ref{tab:attr_facet_map}}\rev{ gives the assignment of these 13 attributes to the five sensory facets.}
The \texttt{value} field is a short open-vocabulary phrase describing the sensory property.
The \texttt{evidence} field is a short text span copied from the input.
The \texttt{polarity} field records sentiment toward the extracted sensory value.
The \texttt{negated} flag indicates explicit absence, such as \emph{odorless} or \emph{no smell}.
The \texttt{confidence} field is a real-valued score in $[0,1]$.
An example output is provided in Appendix~\ref{app:json_example}.

\paragraph{Teacher construction.}
The teacher used in Stage~1 is constructed from seed supervision.
We generate a seed dataset of 10{,}882 items using GPT-5 Mini\footnote{\url{https://platform.openai.com/docs/models/gpt-5-mini}} with a sensory-only extraction prompt.
\rev{The items are drawn from the Amazon Reviews 2018 collections of the five product categories and are disjoint from the evaluation catalogs; the corpus construction is described in Section~\mbox{\ref{sec:method_supervision}}.}
The prompt and request schema are provided in Appendix~\ref{app:prompt_guidelines}.
The seed data are split into 10{,}337 training items and 545 validation items.

\paragraph{Seed label audit.}
To verify the reliability of the seed supervision, we manually audit a random 10\% sample of the 10{,}882 seed items.
Three annotators evaluate the extracted records, and final labels are aggregated by majority vote.
We report four criteria:
\emph{pair precision}, the fraction of extracted attribute--value records judged correct as sensory facts;
\emph{evidence faithfulness}, the fraction of records whose evidence span supports the extracted value;
\emph{polarity accuracy}, agreement between predicted and human polarity labels when polarity is not \texttt{unknown};
and \emph{negation accuracy}, correctness of the \texttt{negated} flag.
Table~\ref{tab:seed_audit} summarizes the audit results.

\begin{table}[t]
\centering
\small
\setlength{\tabcolsep}{3.5pt}
\renewcommand{\arraystretch}{1.05}
\caption{Seed label audit on a random 10\% sample.}
\label{tab:seed_audit}
\begin{tabular}{lcccc}
\toprule
 & Pair P & Evid.\ F & Pol.\ Acc & Neg.\ Acc \\
\midrule
Mean & 0.88 & 0.90 & 0.80 & 0.92 \\
\bottomrule
\end{tabular}
\end{table}

The audit suggests that most extracted records are valid sensory facts and that the cited evidence spans generally support the extracted values.
Polarity prediction is less reliable than fact and evidence correctness, which is expected because sentiment toward a specific sensory facet can be subtle and context-dependent.
Negation handling is strong, indicating that explicit absence cues are captured consistently.

\paragraph{Teacher fine-tuning.}
A Qwen3 model\footnote{\url{https://huggingface.co/Qwen/Qwen3-30B-A3B-Instruct-2507}} is fine-tuned with LoRA\rev{~\mbox{\citep{hu2022lora}}} to reproduce the GPT-5 Mini extraction behavior.
The base model specification follows the Qwen3 technical report~\citep{qwen2025qwen3}, and LoRA is used for parameter-efficient adaptation.
We evaluate semantic alignment between the resulting Qwen teacher and GPT-5 Mini on the validation split.
Alignment is measured by micro-averaged precision, recall, and F1 over extracted attribute--value pairs after normalizing \rev{letter case} and whitespace.
The resulting scores are reported in Table~\ref{tab:teacher_alignment}.

\subsection{Stage 2: Facet-Level Supervision Construction}
\label{sec:method_supervision}

The teacher produces attribute-level JSON records, but the student requires fixed supervision targets over a small number of sensory facets.
Stage~2 therefore converts cleaned teacher records into supervision over five sensory facets:
\[
\mathcal{F}
=
\{\textit{visual},\textit{tactile},\textit{auditory},\textit{olfactory},\textit{gustatory}\}.
\]
This conversion preserves the teacher's structured evidence while producing \rev{targets for training} a compact student encoder.

We use \emph{attribute} to denote the fine-grained teacher extraction labels (e.g., color, finish, scent, texture, and sound). We use \emph{facet} to denote the five coarse groups used by the student and downstream representation: visual, tactile, auditory, olfactory, and gustatory. When needed, \emph{taxonomy} refers to this five-facet mapping. \rev{The complete attribute-to-facet mapping is shown in Table~\mbox{\ref{tab:attr_facet_map}}.}

\begin{revblock}
\paragraph{Choice of the sensory schema.}
The five facets correspond to the five classical exteroceptive modalities considered in this work. We use them as a domain-independent operational partition in which every retained attribute is mapped to exactly one facet, allowing a single schema and a single student to transfer across all five Amazon domains without per-domain reannotation. The 13 fine-grained attributes in Table~\ref{tab:attr_facet_map} are not intended as a general linguistic inventory of sensory language. Instead, they are the attribute types that recur across all five domains frequently enough to support supervised training. They were selected using a pilot sample before any downstream evaluation and were then fixed. The inventory is deliberately imbalanced: nine of the 13 attributes are visual because visual descriptions are by far the most differentiated in product text, whereas the tactile, auditory, olfactory, and gustatory facets are each represented by a single attribute type whose surface realizations are much less taxonomically separable. Table~\ref{tab:modality_domain} illustrates the substantial variation in facet coverage across domains: gustatory coverage ranges from 0.05\% in Video Games to 94.53\% in Grocery, and olfactory coverage ranges from 4.06\% in Video Games to 66.12\% in Beauty. This variation motivates facet-specific modeling rather than uniform treatment of all sensory dimensions. Facet-specific heads with per-facet confidence thresholds (Table~\ref{tab:attr_facet_map}) allow one schema to accommodate this two-order-of-magnitude variation.

\paragraph{Limitations of the sensory schema.}
The schema prioritizes precision, and its cost in recall is real and measurable. On the seed annotation corpus of 10{,}882 annotated items, which is drawn from the Amazon Reviews 2018 supervision source rather than from the evaluation domains, an unconstrained extraction prompt caused the teacher to propose attribute types outside our 13-attribute inventory for 9.2\% of all extracted attributes (2{,}602 of 28{,}146). The excluded types are dominated by three families for which our inventory has no slot: weight and heft (1{,}422 attributes), comfort and ergonomics (1{,}051), and thermal sensation (129). Beyond these, the schema cannot represent interoceptive and nociceptive sensations that appear frequently in Beauty text, such as burning, tingling, stinging, and warmth, or proprioceptive properties such as balance and bendability. These are genuine sensory experiences that our five exteroceptive facets cannot express, and their absence limits the coverage rather than the precision of the sensory channel. The imbalance noted above is a second limitation: with nine visual attributes and one attribute for each of the remaining four facets, the schema resolves visual variation much more finely than nonvisual variation, which is reflected downstream in per-facet quality. We regard the inventory as extensible rather than final: the schema is a closed list associated with a fixed record format, so adding a sensation family requires new attribute types and a facet slot but does not require redesigning the pipeline.
\end{revblock}

\paragraph{Balanced multi-domain supervision corpus.}
Because sensory evidence differs across domains, we construct a balanced supervision corpus for student training.
We use Amazon Reviews 2018\rev{~\mbox{\citep{ni2019justifying}}} examples from five domains: Beauty, Toys, Sports, Video Games, and Grocery.
This choice is intentional: the student supervision corpus requires a large pool of item texts and reviews for teacher extraction, source balancing, and rare-facet coverage, and Amazon Reviews 2018 provides substantially more catalog and review content than Amazon Reviews 2014.
By contrast, Amazon Reviews 2014 is used for downstream recommendation evaluation because it remains a widely used sequential recommendation benchmark.
Thus, Amazon 2018 serves as a large-scale item-content source for training the sensory student, while Amazon 2014 serves as the behavioral benchmark for evaluating recommendation performance.

\rev{The corpus is balanced at the item level: 10{,}882 items are sampled in near-equal counts per domain (2{,}178 for Beauty and 2{,}176 for each of the other four domains), and each item contributes one annotation input that packs its title, category, description, and review excerpts into a single text. This exposes the student to both concise metadata-style cues and longer review-based evidence within every example.}

\paragraph{Teacher-label cleaning.}
We clean teacher outputs using source-grounded evidence filtering.
An extracted attribute is retained only when its facet is valid, its evidence string is non-empty, and the evidence span appears in the normalized source text.
We also apply facet-aware confidence thresholds:
0.60 for visual, tactile, and olfactory attributes, 0.72 for auditory attributes, and 0.75 for gustatory attributes.
Auditory and gustatory facets use stricter thresholds because their textual evidence is sparser and more ambiguous.
These thresholds were set conservatively based on pilot validation checks to favor precision of retained labels and were then held fixed across domains and downstream experiments.
For short titles and descriptions, we apply stricter source-specific margins and an additional margin for very short inputs.
This procedure removes unsupported labels while preserving a large supervision pool, \rev{reducing the total number of records from 105{,}952 to 93{,}675}.

\begin{table}[htbp]
\centering
\footnotesize
\setlength{\tabcolsep}{4pt}
\caption{\rev{Attribute records per domain as returned by the seed model and after evidence-grounded cleaning. Counts are over the 10{,}882 items of the seed corpus, one packed text per item.}}
\label{tab:domain_attr_cleaning}
\begin{tabular}{lrrr}
\toprule
Domain & Raw & Kept & Drop rate (\%) \\
\midrule
Beauty     & 21,196 & 19,612 & 7.47 \\
Grocery    & 20,983 & 17,804 & 15.15 \\
Sports     & 22,786 & 20,435 & 10.32 \\
Toys       & 22,174 & 19,913 & 10.20 \\
VideoGames & 18,813 & 15,911 & 15.43 \\
\midrule
Total      & 105,952 & 93,675 & 11.59 \\
\bottomrule
\end{tabular}
\end{table}

\begin{revblock}
\paragraph{Data flow.}
Three corpora play distinct roles in the pipeline.
First, the seed corpus consists of the 10{,}882 items above: GPT-5 Mini annotates each item's packed text once, returning at most twelve records per item and 105{,}952 records in total. The request format of Appendix~\mbox{\ref{app:prompt_format}} permits up to twenty attributes per item; the retained set is capped at twelve when the annotations are assembled, and that cap binds for 2{,}160 of the items. evidence-grounded cleaning retains 93{,}675 of them (Table~\mbox{\ref{tab:domain_attr_cleaning}}). Each item therefore contributes exactly one supervision example, giving 10{,}882 examples that are split into 10{,}337 for training and 545 for validation.
Second, the teacher of Stage~1 is fine-tuned on the training split of this corpus, and the facet-aware student of Stage~3 is trained on the same cleaned, capped records; the temporal control of Section~\mbox{\ref{sec:temporal_control}} filters exactly this corpus.
Third, the evaluation corpora are the Amazon Reviews 2014 5-core datasets of Table~\mbox{\ref{tab:dataset}}; the frozen student encodes item-side text of every evaluation item to produce the facet banks, and no evaluation item contributes supervision.
\end{revblock}

\paragraph{Facet mapping and token evidence masks.}
Each retained attribute is mapped to one of the five sensory facets \rev{according to Table~\mbox{\ref{tab:attr_facet_map}}}.

\begin{table}[htbp]
\centering
\footnotesize
\setlength{\tabcolsep}{4pt}
\renewcommand{\arraystretch}{1.08}
\begin{revtable}
\begin{tabular}{@{}llcc@{}}
\toprule
Facet & Fine-grained attributes & \# & Conf.\ thr. \\
\midrule
Visual     & \texttt{color}, \texttt{pattern}, \texttt{shape}, \texttt{graphics}, & 9 & 0.60 \\
           & \texttt{brightness}, \texttt{glossiness}, \texttt{transparency}, &  &  \\
           & \texttt{finish}, \texttt{design} &  &  \\
\addlinespace[2pt]
Tactile    & \texttt{texture}   & 1 & 0.60 \\
Auditory   & \texttt{sound}     & 1 & 0.72 \\
Olfactory  & \texttt{scent}     & 1 & 0.60 \\
Gustatory  & \texttt{flavor}    & 1 & 0.75 \\
\midrule
Total      &                    & 13 & \\
\bottomrule
\end{tabular}
\end{revtable}
\caption[Attribute-to-facet mapping]{\rev{Mapping from the 13 fine-grained teacher attributes to the five sensory facets, with the per-facet confidence threshold applied during teacher-label cleaning. The mapping is fixed a priori, is shared by all five domains, and is the only attribute-to-facet routing used anywhere in the pipeline.}}
\label{tab:attr_facet_map}
\end{table}

For each training example and each facet $f \in \mathcal{F}$, we construct a binary token-level evidence mask \rev{over the \mbox{$L$} tokens of the encoded input}
\[
\mathbf{m}^{(f)} = [m_1^{(f)},\ldots,m_L^{(f)}],
\]
where $m_t^{(f)}=1$ if token $t$ belongs to a teacher-provided evidence span for facet $f$, and $m_t^{(f)}=0$ otherwise.
Evidence spans are grounded to token positions by substring matching against the original text span and then projected to DeBERTa token positions.
We also define the content mask $\mathcal{C}$ as the set of token positions corresponding to the item text, excluding domain and source prefix tokens.

\paragraph{Facet-level labels.}
For each example and facet, we construct a binary presence label $y^{(f)}$.
The label is positive when at least one retained teacher record belongs to facet $f$.
We also construct a polarity label $y_{\mathrm{pol}}^{(f)}$ and a confidence target $y_{\mathrm{conf}}^{(f)}$ from the retained teacher records.
When multiple records exist for the same facet, we aggregate confidence scores and select the dominant polarity after confidence weighting.
Teacher polarity values that cannot be mapped to the three modeled classes \rev{(positive, negative, and neutral)} are masked from the polarity loss.

\paragraph{Example and canonical embedding targets.}
The student uses two dense targets. The first target supervises the
example facet embedding, and the second supervises the canonical facet
embedding. For each example and facet, we form a facet target text
$g^{(f)}$ by concatenating up to three high-confidence
attribute--value pairs within the same facet. For example, a visual target
text may be \texttt{color = neon blue ; transparency = see-through}.

To avoid undefined normalization for inactive facets, we use a safe
$L_2$-normalization operator:
\begin{equation}
\operatorname{safe}\text{-}\ell_2(\mathbf{x}, a)
=
\begin{cases}
\dfrac{\mathbf{x}}{\max(\lVert \mathbf{x} \rVert_2, \epsilon_n)},
& a = 1,\\[6pt]
\mathbf{0},
& a = 0,
\end{cases}
\label{eq:safe_l2}
\end{equation}
where $a$ is an active-facet indicator and $\epsilon_n$ is a small
numerical constant.

For an active example-facet pair, a frozen target encoder \rev{\mbox{$E_{\mathrm{target}}$}, instantiated as the general-purpose sentence encoder all-mpnet-base-v2 \mbox{\citep{reimers2019sentence}} with 768-dimensional output,} produces an
example-level teacher embedding:
\begin{equation}
\mathbf{v}^{(f)}_{\mathrm{teacher}}
=
\begin{cases}
\operatorname{safe}\text{-}\ell_2
\left(E_{\mathrm{target}}\left(g^{(f)}\right), 1\right),
& y^{(f)} = 1,\\[4pt]
\mathbf{0},
& y^{(f)} = 0.
\end{cases}
\label{eq:example_teacher_embedding}
\end{equation}
For the canonical target, we aggregate teacher embeddings associated
with the same item and facet. We first define the active canonical
facet indicator:
\begin{equation}
a^{(f)}_{\mathrm{target},i}
=
\mathbb{I}
\left[
\sum_{j \in \mathcal{T}_i}
\tilde{\omega}^{(f)}_{ij}
>
\epsilon_w
\right],
\label{eq:canonical_target_active}
\end{equation}
where $\mathcal{T}_i$ is the set of texts associated with item $i$,
$\tilde{\omega}^{(f)}_{ij}$ is the teacher-side confidence weight, \rev{defined as the product of the facet-presence indicator and the teacher-reported confidence of text \mbox{$j$} for facet \mbox{$f$}, \mbox{$\tilde{\omega}^{(f)}_{ij} = \mathbb{I}[\,y^{(f)}_{ij}=1\,]\cdot\mathrm{conf}^{(f)}_{ij}$}, so that it is the teacher-side counterpart of the student aggregation weight \mbox{$\omega^{(f)}_{ij}$} in Eq.~\mbox{\ref{eq:item_bank_weight}},} and
$\epsilon_w$ is a small evidence-mass threshold\rev{, and \mbox{$\epsilon$} denotes a small positive constant added to denominators for numerical stability}. The canonical target is
then
\begin{equation}
\mathbf{c}^{(f)}_{\mathrm{target},i}
=
\operatorname{safe}\text{-}\ell_2
\left(
\frac{
\sum_{j \in \mathcal{T}_i}
\tilde{\omega}^{(f)}_{ij}
\mathbf{v}^{(f)}_{\mathrm{teacher},ij}
}{
\sum_{j \in \mathcal{T}_i}
\tilde{\omega}^{(f)}_{ij}
+
\epsilon
},
a^{(f)}_{\mathrm{target},i}
\right).
\label{eq:canonical_target_embedding}
\end{equation}

The example target $\mathbf{v}^{(f)}_{\mathrm{teacher}}$ supervises the
example embedding head, while the canonical target
$\mathbf{c}^{(f)}_{\mathrm{target},i}$ supervises the canonical head, which
aggregates several texts of an item into one facet representation.
\rev{In the supervision corpus each item contributes a single packed text, so the
two targets coincide during training; the canonical head is exercised at bank
construction (Section~\mbox{\ref{sec:bank}}), where an evaluation item is
represented by many separate review texts.}
Embedding-alignment and canonical-alignment losses are applied only to
active facet slots. When the corresponding active indicator is zero, the
target vector is set to $\mathbf{0}$ and the dense alignment loss for that
facet slot is masked out. The frozen target encoder is used only for
supervision construction and is not used by the downstream recommender.

\paragraph{Static item-content setting and leakage scope.}
We adopt a static item-content recommendation setting.
The catalog fields and review text used to construct the sensory supervision and the item-level facet bank are treated as item-side content, analogous to metadata or item descriptions, and are available independently of the user--item interaction split.
The leave-one-out split is applied only to behavioral interaction sequences: for each user, the final interaction is held out for testing, the second-to-last interaction is used for validation, and all earlier interactions are used for training.
Thus, the recommender never observes the held-out user--item interaction during training.

To avoid item-level textual overlap between student supervision and downstream recommendation, we remove from the Amazon Reviews 2018 supervision corpus any items that also appear in the Amazon Reviews 2014 5-core domains used for evaluation.
The student is therefore trained on a disjoint item set and is used only as a frozen encoder when constructing the downstream sensory table. The student never observes user identifiers, user histories, held-out targets, or recommendation losses.
Likewise, the downstream recommender uses only the precomputed item-level facet bank together with lightweight
runtime projection and mixing modules, and does not receive teacher outputs, raw reviews, or online text encodings at
evaluation time.
Therefore, the evaluation measures whether static sensory item content improves next-item prediction from user histories, rather than whether the model can exploit future behavioral interactions.
\rev{Item disjointness alone, however, addresses textual overlap rather than temporal precedence; Section~\mbox{\ref{sec:temporal_control}} provides an empirical temporal control in which supervision from items lacking any pre-cutoff review is removed.} \rev{The rule also governs supervision rather than the bank: the facet bank of an evaluation domain is built from the review text of that domain's own catalog, so an item's representation can draw on reviews that belong to held-out interactions. In Beauty, 44{,}720 of the 198{,}475 review texts that feed the bank, or 22.5\%, are the test or validation interaction of some user. This is item content in the sense used here, since no user identifier, interaction order, or label is available to the encoder, but it is not independent of the evaluated interactions, and the Limitations section records it as such.}
We use Amazon Reviews 2018 for sensory student supervision because its larger item and review coverage provides a richer source of item-side text than Amazon Reviews 2014, while Amazon Reviews 2014 is retained as the downstream sequential recommendation benchmark.

\subsection{Stage 3: Multi-Facet Sensory Student Distillation}
\label{sec:method_student}

Stage~3 trains a compact multi-facet student on the supervision constructed in Stage~2.
The student is designed to replace repeated teacher LLM inference with efficient text encoding.
It also produces an item-level facet bank that can be precomputed offline and consumed by lightweight runtime
integration modules during recommendation.

\paragraph{Student supervision corpus statistics.}
Before describing the student architecture, we summarize the supervision corpus as consumed by the DeBERTa-based student.
Table~\ref{tab:text_stats} reports token-length statistics after DeBERTa tokenization.
\rev{Because title, category, description, and review excerpts are packed into a single text per item, the inputs are long: the median is 542 tokens and 72.4\% of items reach the 512-token limit, so the encoder sees a truncated prefix for most items. Section~\mbox{\ref{sec:recall_audit}} measures what this truncation costs in extraction recall.}

\begin{table}[htbp]
\centering
\caption{\rev{Token-length statistics of the packed item texts that form the student training corpus, measured with the DeBERTa tokenizer before the 512-token input limit is applied. Each item contributes one such text.}}
\label{tab:text_stats}
\begin{tabular}{lrrrr}
\toprule
Statistic & Avg Tokens & Median & P95 & Max \\
\midrule
Packed item text & 497.6 & 542 & 632 & 1{,}011 \\
\bottomrule
\end{tabular}
\end{table}

Table~\ref{tab:modality_domain} summarizes the coverage of the five sensory facets in the cleaned supervision corpus.
This statistic is computed from Stage~2 labels, but it directly motivates the Stage~3 design.
Visual and tactile facets appear frequently in Beauty, Toys, and Sports.
Olfactory attributes are concentrated in Beauty and Grocery.
Gustatory attributes are dominant in Grocery.
\rev{Auditory attributes are the most domain-dependent: they are present for roughly half of the Video Games items and 43\% of Toys items, but for only 3\% of Grocery items and 10\% of Beauty items.} Games exhibits relatively low overall retained coverage across the five facets, although its auditory share is the highest of the five domains. We therefore interpret Games as a domain with sparse but informative sensory evidence rather than uniformly dense sensory language. This imbalance motivates a single all-domain encoder with facet-specific heads rather than a single undifferentiated sensory-vector regression target.

\begin{table}[htbp]
\centering
\footnotesize
\setlength{\tabcolsep}{2.6pt}
\renewcommand{\arraystretch}{1.05}
\caption{\rev{Facet coverage across domains in the student training corpus, given as the percentage of supervision examples in that domain for which the facet is present. Each example is one item.}}
\label{tab:modality_domain}
\begin{tabular}{lccccc}
\toprule
Facet & Beauty & Toys & Sports & Games & Groc. \\
\midrule
Visual    & 85.45 & 97.19 & 97.34 & 87.91 & 56.56 \\
Tactile   & 89.71 & 88.24 & 95.40 & 70.59 & 83.84 \\
Auditory  & 10.29 & 42.86 & 23.13 & 49.64 & 3.00 \\
Olfactory & 66.12 & 19.79 & 29.90 & 4.06  & 52.06 \\
Gustatory & 8.51  & 1.50  & 10.89 & 0.05  & 94.53 \\
\bottomrule
\end{tabular}
\end{table}

\paragraph{Student input and shared encoder.}
Each student input is formatted as
\[
\texttt{domain equals } \delta \ ; \ 
\texttt{source equals } s \ ; \ 
\texttt{text equals } x,
\]
where $\delta$ denotes the product domain and $s$ denotes the source type of the text.
\rev{The input format admits several source types; in the corpus used here each item contributes one packed text that already combines its title, category, description, and reviews, so \mbox{$s$} takes a single value throughout.}
We use DeBERTa-v3-Small~\citep{he2021debertav3} as the shared text encoder.
Given an input sequence of length $L$, the encoder produces contextual token representations
\rev{of hidden size \mbox{$d_h$}, the DeBERTa-v3-Small encoder width (\mbox{$d_h=768$}),}
\[
  H = [\mathbf{h}_1,\ldots,\mathbf{h}_L] \in \mathbb{R}^{L \times d_h}.
\]
A per-domain bottleneck adapter applies a residual correction:
\begin{equation}
  \tilde{H}
  =
  H + \mathrm{Adapter}_{\delta}(H),
  \label{eq:domain_adapter}
\end{equation}
where $\mathrm{Adapter}_{\delta}$ is a two-layer bottleneck MLP
$d_h \to d_b \to d_h$ with $d_b=128$ and GELU activation. \rev{Throughout, \mbox{$\sigma(\cdot)$} denotes the logistic sigmoid.}
The adapter is specific to domain $\delta$, while the DeBERTa encoder is shared across domains. \rev{The symbol \mbox{$\delta$} is reserved for the domain index; \mbox{$d$} denotes the recommender hidden size.}

The student processes five sensory facets
$\mathcal{F}=\{\textit{visual},\textit{tactile},\textit{auditory},\textit{olfactory},\textit{gustatory}\}$.
For each facet, the forward pass follows four internal phases:
token-level evidence detection,
facet presence pooling,
evidence-weighted representation pooling,
and facet-specific prediction.
Algorithm~\ref{alg:student-forward} summarizes this computation. \rev{In the algorithm, \mbox{$D$} is the dimensionality of the facet embeddings produced by the student heads (\mbox{$D=768$}, see Eq.~\mbox{\ref{eq:canonical_embedding}}).}

\begin{algorithm}[t]
\caption{Forward Pass of Multi-Facet Sensory Student}
\label{alg:student-forward}
\begin{algorithmic}[1]
\Require Input tokens $x$, attention mask, content mask $\mathcal{C}$, domain id $\delta$
\Ensure Per-facet outputs: $\{p_t^{(f)},\, p^{(f)},\, v_{\mathrm{ex}}^{(f)},\, c^{(f)},\, \mathrm{pol}^{(f)},\, \mathrm{conf}^{(f)}\}_{f \in \mathcal{F}}$

\State $H \gets \textsc{DeBERTa}(x)$ \Comment{$H \in \mathbb{R}^{L \times d_h}$}
\State $\tilde{H} \gets H + \mathrm{Adapter}_{\delta}(H)$ \Comment{Per-domain bottleneck residual}

\For{each facet $f \in \mathcal{F}$}
  \Statex \hspace{\algorithmicindent}\textcolor{gray}{\textit{// Phase 1: Token-level evidence detection}}
  \State $p_t^{(f)} \gets \sigma\!\bigl(w_f^\top \tilde{h}_t\bigr) \cdot \mathbb{I}[t \in \mathcal{C}]$
    \quad $\forall\, t \in \{1,\ldots,L\}$

  \Statex \hspace{\algorithmicindent}\textcolor{gray}{\textit{// Phase 2: Facet presence via noisy-or pooling}}
  \State $p^{(f)} \gets 1 - \prod_{t \in \mathcal{C}} \bigl(1 - p_t^{(f)}\bigr)$

  \Statex \hspace{\algorithmicindent}\textcolor{gray}{\textit{// Phase 3: Evidence-weighted token pooling}}
  \State $\alpha_t^{(f)} \leftarrow p_t^{(f)} / \left(\sum_{t' \in \mathcal{C}} p_{t'}^{(f)} + \epsilon\right)$
  \If{$\sum_{t' \in \mathcal{C}} p_{t'}^{(f)} \le \epsilon$}
    \State $\alpha_t^{(f)} \gets 1 / |\mathcal{C}|$ \quad $\forall\, t \in \mathcal{C}$
    \Comment{Uniform fallback}
  \EndIf
  \State $r^{(f)} \gets \sum_{t \in \mathcal{C}} \alpha_t^{(f)}\, \tilde{h}_t$
    \Comment{$r^{(f)} \in \mathbb{R}^{d_h}$}

  \Statex \hspace{\algorithmicindent}\textcolor{gray}{\textit{// Phase 4: Facet-specific prediction heads}}
  \State $v_{\mathrm{ex}}^{(f)} \gets \ell_2\text{-norm}\!\bigl(\mathrm{MLP}_{\mathrm{ex}}^{(f)}(r^{(f)})\bigr)$
    \Comment{Example embedding, $\in \mathbb{R}^D$}
  \State $c^{(f)} \gets \ell_2\text{-norm}\!\bigl(\mathrm{MLP}_{\mathrm{can}}^{(f)}(r^{(f)})\bigr)$
    \Comment{Canonical embedding, $\in \mathbb{R}^D$}
  \State $\mathrm{pol}^{(f)} \gets W_{\mathrm{pol}}^{(f)}\, v_{\mathrm{ex}}^{(f)}$
    \Comment{Polarity logits, $\in \mathbb{R}^3$}
  \State $\mathrm{conf}^{(f)} \gets \sigma\!\bigl(w_{\mathrm{conf}}^{(f)\top} v_{\mathrm{ex}}^{(f)}\bigr)$
    \Comment{Confidence, $\in (0,1)$}
\EndFor
\end{algorithmic}
\end{algorithm}

\begin{algorithm}[t]
\caption{Training Procedure of Multi-Facet Sensory Student}
\label{alg:student-training}
\begin{algorithmic}[1]
\Require Dataset $\mathcal{D}$ with teacher annotations
 $\{m_t^{(f)},\, y^{(f)},\, v_{\mathrm{teacher}}^{(f)},\, c_{\mathrm{target}}^{(f)},\, y_{\mathrm{pol}}^{(f)},\, y_{\mathrm{conf}}^{(f)}\}_{f \in \mathcal{F}}$

\Statex \textcolor{gray}{\textit{// Phase A: Token evidence + presence + light alignment}}
\For{each mini-batch $(x, \delta) \sim \mathcal{D}$}
  \State Compute forward pass (Algorithm~\ref{alg:student-forward})
  \For{each facet $f \in \mathcal{F}$}
    \State $\mathcal{L}_{\mathrm{tok}}^{(f)} \gets \mathrm{FocalBCE}\!\bigl(p_t^{(f)},\; m_t^{(f)}\bigr)$
    \State $\mathcal{L}^{(f)}_{\mathrm{pres}} \gets \mathrm{WeightedBCE}\!\bigl(p^{(f)},\; y^{(f)}\bigr)$
    \State $\mathcal{L}_{\mathrm{align}}^{(f)} \gets y^{(f)} \cdot \bigl(1 - \cos\!\bigl(v_{\mathrm{ex}}^{(f)},\; v_{\mathrm{teacher}}^{(f)}\bigr)\bigr)$
  \EndFor
  \State $\mathcal{L}_A \gets \sum_{f} \lambda_f \bigl(\mathcal{L}_{\mathrm{tok}}^{(f)} + \mathcal{L}_{\mathrm{pres}}^{(f)} + \mathcal{L}_{\mathrm{align}}^{(f)}\bigr)$
  \State Update $\theta$ via AdamW on $\mathcal{L}_A$
\EndFor

\Statex \textcolor{gray}{\textit{// Phase B: All objectives enabled}}
\For{each mini-batch $(x, \delta) \sim \mathcal{D}$}
  \State Compute forward pass (Algorithm~\ref{alg:student-forward})
  \For{each facet $f \in \mathcal{F}$}
    \State $\mathcal{L}_{\mathrm{tok}}^{(f)},\; \mathcal{L}_{\mathrm{pres}}^{(f)},\; \mathcal{L}_{\mathrm{align}}^{(f)}$ \quad (same as Phase A)
    \State $\mathcal{L}_{\mathrm{pol}}^{(f)} \gets \mathrm{CE}\!\bigl(\mathrm{pol}^{(f)},\; y_{\mathrm{pol}}^{(f)}\bigr)$
    \State $\mathcal{L}_{\mathrm{conf}}^{(f)} \gets \mathrm{MSE}\!\bigl(\mathrm{conf}^{(f)},\; y_{\mathrm{conf}}^{(f)}\bigr)$
    \State $\mathcal{L}_{\mathrm{can}}^{(f)} \gets a_{\mathrm{target}}^{(f)} \cdot \bigl(1 - \cos\!\bigl(c^{(f)},\; c_{\mathrm{target}}^{(f)}\bigr)\bigr)$
    \State $\mathcal{L}_{\mathrm{nce}}^{(f)} \gets \mathrm{InfoNCE}\!\bigl(v_{\mathrm{ex}}^{(f)},\; v_{\mathrm{teacher}}^{(f)}\bigr)$ \Comment{In-batch contrastive}
  \EndFor
  \State $\mathcal{L}_B \gets \sum_{f} \lambda_f \bigl(\mathcal{L}_{\mathrm{tok}}^{(f)} + \mathcal{L}_{\mathrm{pres}}^{(f)} + \mathcal{L}_{\mathrm{align}}^{(f)} + \mathcal{L}_{\mathrm{pol}}^{(f)} + \mathcal{L}_{\mathrm{conf}}^{(f)} + \mathcal{L}_{\mathrm{can}}^{(f)} + \mathcal{L}_{\mathrm{nce}}^{(f)}\bigr)$
  \State Update $\theta$ via AdamW on $\mathcal{L}_B$
\EndFor
\end{algorithmic}
\end{algorithm}

\paragraph{Phase 1: Token-level evidence detection.}
For each facet $f \in \mathcal{F}$ and token position $t$, a dedicated linear head predicts the probability that token $t$ is sensory evidence for facet $f$:
\begin{equation}
  p_t^{(f)}
  =
  \sigma
  \left(
  \mathbf{w}_f^{\top}\tilde{\mathbf{h}}_t
  \right)
  \cdot \mathbb{I}[t \in \mathcal{C}],
  \qquad t=1,\ldots,L.
  \label{eq:token_evidence}
\end{equation}
Here, $\mathcal{C}$ is the content-token mask, excluding domain and source prefix tokens.
The supervision target is the teacher-derived binary evidence mask $m_t^{(f)}$.

\paragraph{Phase 2: Facet presence via noisy-or pooling.}
Facet presence is derived from token-level evidence probabilities rather than predicted by a separate sentence-level classifier.
For each facet, we apply noisy-or pooling over content tokens:
\begin{equation}
  p^{(f)}
  =
  1 -
  \prod_{t \in \mathcal{C}}
  \left(1 - p_t^{(f)}\right).
  \label{eq:noisy_or}
\end{equation}
This reflects the assumption that a facet is present when at least one content token provides evidence for that facet.

\paragraph{Phase 3: Evidence-weighted token pooling.}
The student constructs a grounded facet representation by using token evidence probabilities as soft attention weights.
We first normalize the evidence probabilities:
\begin{equation}
  \alpha_t^{(f)}
  =
  \frac{p_t^{(f)}}
  {\sum_{t' \in \mathcal{C}} p_{t'}^{(f)} + \epsilon},
  \qquad t \in \mathcal{C}.
  \label{eq:evidence_weight}
\end{equation}
Algorithm~1 uses the same $\epsilon$-stabilized normalization and applies the uniform fallback when the total evidence mass is at most $\epsilon$.
If the total evidence mass is negligible, the weights fall back to uniform pooling:
\[
  \alpha_t^{(f)} = 1/|\mathcal{C}|,
  \qquad t \in \mathcal{C}.
\]
The grounded facet representation is
\begin{equation}
  \mathbf{r}^{(f)}
  =
  \sum_{t \in \mathcal{C}}
  \alpha_t^{(f)} \tilde{\mathbf{h}}_t,
  \qquad
  \mathbf{r}^{(f)} \in \mathbb{R}^{d_h}.
  \label{eq:evidence_pooling}
\end{equation}
Thus, the dense representation is focused on predicted evidence spans rather than on the entire input sequence.

\paragraph{Phase 4: Facet-specific prediction heads.}
The evidence-weighted representation $\mathbf{r}^{(f)}$ feeds two embedding heads.
The first produces an example facet embedding:
\begin{equation}
  \mathbf{v}_{\mathrm{ex}}^{(f)}
  =
  \ell_2\text{-norm}
  \left(
  \mathrm{MLP}_{\mathrm{ex}}^{(f)}
  \left(\mathbf{r}^{(f)}\right)
  \right),
  \qquad
  \mathbf{v}_{\mathrm{ex}}^{(f)} \in \mathbb{R}^{D}.
  \label{eq:example_embedding}
\end{equation}
The second produces a canonical facet embedding:
\begin{equation}
  \mathbf{c}^{(f)}
  =
  \ell_2\text{-norm}
  \left(
  \mathrm{MLP}_{\mathrm{can}}^{(f)}
  \left(\mathbf{r}^{(f)}\right)
  \right),
  \qquad
  \mathbf{c}^{(f)} \in \mathbb{R}^{D}.
  \label{eq:canonical_embedding}
\end{equation}
We use $D=768$.

Polarity and confidence are predicted from the example embedding, not directly from $\mathbf{r}^{(f)}$.
The polarity head produces three-class logits:
\begin{equation}
  \mathrm{pol}^{(f)}
  =
  W_{\mathrm{pol}}^{(f)}
  \mathbf{v}_{\mathrm{ex}}^{(f)}
  \in \mathbb{R}^{3},
  \label{eq:polarity_head}
\end{equation}
corresponding to positive, negative, and neutral.
The confidence head predicts a scalar confidence score:
\begin{equation}
  \mathrm{conf}^{(f)}
  =
  \sigma
  \left(
  \mathbf{w}_{\mathrm{conf}}^{(f)\top}
  \mathbf{v}_{\mathrm{ex}}^{(f)}
  \right)
  \in (0,1).
  \label{eq:confidence_head}
\end{equation}

The example embedding is optimized to match example-level teacher semantics, while the canonical embedding is optimized to form a stable item-level facet representation shared across texts associated with the same item.
This separation is important because review-level examples may emphasize different local cues, whereas the recommender requires a reusable item-level sensory feature.

\paragraph{Training objective.}
Student training follows Algorithm~\ref{alg:student-training}.
For each facet, we use seven losses:
token evidence tagging,
facet presence prediction,
example embedding alignment,
polarity classification,
confidence regression,
canonical embedding alignment,
and in-batch contrastive regularization.

The token evidence loss is focal binary cross-entropy\rev{~\mbox{\citep{lin2017focal}}}:
\begin{equation}
  \mathcal{L}_{\mathrm{tok}}^{(f)}
  =
  \mathrm{FocalBCE}
  \left(
  p_t^{(f)}, m_t^{(f)}
  \right).
  \label{eq:loss_tok}
\end{equation}
The presence loss is weighted BCE:
\begin{equation}
\mathcal{L}^{(f)}_{\mathrm{pres}}
=
- w_{+} y^{(f)} \log p^{(f)}
- w_{-}(1-y^{(f)}) \log(1-p^{(f)}).
\end{equation}
\rev{Here \mbox{$w_{+}=1.0$} and \mbox{$w_{-}=0.35$} are the positive and negative class weights, set so that the abundant negative facet slots do not dominate the sparse positives.}
The example embedding alignment loss is
\begin{equation}
\mathcal{L}^{(f)}_{\mathrm{align}}
=
y^{(f)}
\left[
1 -
\cos
\left(
\mathbf{v}^{(f)}_{\mathrm{ex}},
\mathbf{v}^{(f)}_{\mathrm{teacher}}
\right)
\right].
\label{eq:student_example_alignment}
\end{equation}
The polarity loss is cross-entropy:
\begin{equation}
\mathcal{L}^{(f)}_{\mathrm{pol}}
=
\operatorname{CE}
\left(
\mathrm{pol}^{(f)},
y^{(f)}_{\mathrm{pol}}
\right).
\label{eq:student_polarity_loss}
\end{equation}
The confidence loss is mean squared error:
\begin{equation}
  \mathcal{L}_{\mathrm{conf}}^{(f)}
  =
  \mathrm{MSE}
  \left(
  \mathrm{conf}^{(f)}, y_{\mathrm{conf}}^{(f)}
  \right).
  \label{eq:loss_conf}
\end{equation}
The canonical embedding alignment loss is
\begin{equation}
\mathcal{L}^{(f)}_{\mathrm{can}}
=
a^{(f)}_{\mathrm{target},i}
\left[
1 -
\cos
\left(
\mathbf{c}^{(f)},
\revmath{\mathbf{c}^{(f)}_{\mathrm{target},i}}
\right)
\right].
\label{eq:student_canonical_alignment}
\end{equation}
The in-batch contrastive loss is
\begin{revblock}
\begin{equation}
\mathcal{L}_{\mathrm{nce}}^{(f)}
=
-\frac{1}{|I_f|}
\sum_{i\in I_f}
\log
\frac{
\exp\!\left(\frac{\langle \hat{\mathbf z}_i^{(f)}, \hat{\mathbf t}_i^{(f)}\rangle}{\tau}\right)
}{
\sum_{j\in I_f}
\exp\!\left(\frac{\langle \hat{\mathbf z}_i^{(f)}, \hat{\mathbf t}_j^{(f)}\rangle}{\tau}\right)
}
.
\label{eq:loss_nce}
\end{equation}
\end{revblock}

Here,
\[
I_f=\{\,i \mid a_{i,f}=1\,\},
\]
where \rev{the active-facet indicator} $a_{i,f}=\mathbb{I}[y_i^{(f)}=1]$ indicates whether facet $f$ is active for example $i$ according to the cleaned teacher annotations, that is, whether at least one retained teacher record is mapped to facet $f$ for that example.
$\hat{\mathbf z}_i^{(f)}$ is the L2-normalized student example embedding for facet $f$, \mbox{$\hat{\mathbf t}_i^{(f)}$}\rev{ is the L2-normalized teacher target }\mbox{$\mathbf{v}^{(f)}_{\mathrm{teacher},i}$}\rev{ of Eq.~}\mbox{\ref{eq:example_teacher_embedding}}\rev{ for the same example,}
and $\tau$ is the temperature ($\tau=0.07$ in our implementation)\rev{, following the InfoNCE formulation~\mbox{\citep{oord2018representation}}}.
For each facet $f$, the positive pair is the diagonal pair $(\hat{\mathbf z}_i^{(f)}, \hat{\mathbf t}_i^{(f)})$, and negatives are in-batch samples $(\hat{\mathbf z}_i^{(f)}, \hat{\mathbf t}_j^{(f)})$ for $j\neq i$.
Thus, $\mathcal{L}_{\mathrm{nce}}^{(f)}$ is computed only over examples in which facet $f$ is present, which avoids contrasting inactive facet slots against teacher targets that are absent for that facet.
This term encourages geometrically consistent example embeddings while preserving facet-specific separation.

\paragraph{Two-phase training.}
Training proceeds in two phases.
\rev{Phase~A provides a warm-up stage for grounding and lightweight semantic alignment:}
\begin{equation}
  \mathcal{L}_{A}
  =
  \sum_{f \in \mathcal{F}}
  \lambda_f
  \left(
  \mathcal{L}_{\mathrm{tok}}^{(f)}
  +
  \mathcal{L}_{\mathrm{pres}}^{(f)}
  +
  \mathcal{L}_{\mathrm{align}}^{(f)}
  \right).
  \label{eq:loss_phase_a}
\end{equation}
Phase~B enables the full objective:
\begin{equation}
  \mathcal{L}_{B}
  =
  \sum_{f \in \mathcal{F}}
  \lambda_f
  \left(
  \mathcal{L}_{\mathrm{tok}}^{(f)}
  +
  \mathcal{L}_{\mathrm{pres}}^{(f)}
  +
  \mathcal{L}_{\mathrm{align}}^{(f)}
  +
  \mathcal{L}_{\mathrm{pol}}^{(f)}
  +
  \mathcal{L}_{\mathrm{conf}}^{(f)}
  +
  \mathcal{L}_{\mathrm{can}}^{(f)}
  +
  \mathcal{L}_{\mathrm{nce}}^{(f)}
  \right).
  \label{eq:loss_phase_b}
\end{equation}
The parameters are optimized with AdamW.
The facet weights $\lambda_f$ control the relative importance of the five sensory facets.
These weights are global coefficients for a single all-domain student and are not intended to mirror per-domain marginal facet coverage.

\paragraph{Balanced all-domain optimization.}
Because the supervision corpus is imbalanced across domains and facets, we use domain-balanced mini-batch sampling.
\rev{This prevents overrepresented domains from dominating student training.}
We also apply domain-token dropout by replacing the domain token with an unknown-domain token for a subset of training examples.
This encourages the shared encoder to learn transferable sensory expressions rather than relying excessively on domain-specific surface forms.

The facet weights $\lambda_f$ are not inverse-frequency class weights.
Class imbalance within token evidence detection and facet presence
prediction is handled by focal BCE and weighted BCE. Instead,
$\lambda_f$ controls the contribution of each facet-level task to the
shared all-domain student objective. We therefore use conservative
weights for auditory and gustatory facets because their supervision is
either sparse, noisy, or highly concentrated in a small number of domains.
Without this conservative weighting, these facets can introduce
high-variance gradients or bias the shared encoder toward
domain-specific lexical patterns, especially flavor expressions that are
concentrated in Grocery.

\paragraph{Item-level facet bank construction.}
\label{sec:bank}
After student training, we run the student offline over all available item texts.
For each item $i$, text example $j$, and facet $f$, the student produces
\[
\left(
p_{ij}^{(f)},
\mathbf{v}_{\mathrm{ex},ij}^{(f)},
\mathbf{c}_{ij}^{(f)},
\mathrm{pol}_{ij}^{(f)},
\mathrm{conf}_{ij}^{(f)}
\right).
\]
We aggregate these example-level outputs into an item-level facet bank.
The aggregation weight is
\begin{equation}
\omega^{(f)}_{ij}
=
p^{(f)}_{ij}
\cdot
\operatorname{conf}^{(f)}_{ij}.
\label{eq:item_bank_weight}
\end{equation}
We define an item-level active-facet indicator:
\begin{equation}
a^{(f)}_i
=
\mathbb{I}
\left[
\sum_j
\omega^{(f)}_{ij}
>
\epsilon_w
\right].
\label{eq:item_bank_active}
\end{equation}
The item-level canonical facet embedding is then computed with safe
normalization:
\begin{equation}
\mathbf{c}^{(f)}_i
=
\operatorname{safe}\text{-}\ell_2
\left(
\frac{
\sum_j
\omega^{(f)}_{ij}
\mathbf{c}^{(f)}_{ij}
}{
\sum_j
\omega^{(f)}_{ij}
+
\epsilon
},
a^{(f)}_i
\right).
\label{eq:item_bank_embedding}
\end{equation}
The item-level confidence score is the evidence-weighted mean confidence:
\begin{equation}
\rho^{(f)}_i
=
\frac{
\sum_j
p^{(f)}_{ij}
\operatorname{conf}^{(f)}_{ij}
}{
\sum_j
p^{(f)}_{ij}
+
\epsilon
}.
\label{eq:item_bank_confidence}
\end{equation}
This uses predicted facet evidence mass as the aggregation weight and
\rev{avoids using confidence to weight itself}. The item-level polarity
distribution is obtained by confidence-weighted logit aggregation:
\begin{equation}
\boldsymbol{\pi}^{(f)}_i
=
\begin{cases}
\operatorname{softmax}
\left(
\dfrac{
\sum_j
\omega^{(f)}_{ij}
\operatorname{pol}^{(f)}_{ij}
}{
\sum_j
\omega^{(f)}_{ij}
+
\epsilon
}
\right),
& a^{(f)}_i = 1,\\[10pt]
\left(\frac{1}{3}, \frac{1}{3}, \frac{1}{3}\right),
& a^{(f)}_i = 0.
\end{cases}
\label{eq:item_bank_polarity}
\end{equation}
We also store a coverage statistic
\begin{equation}
\kappa^{(f)}_i
=
\sum_j
\mathbb{I}
\left[
p^{(f)}_{ij}
>
\tau_{\mathrm{pres}}
\right],
\label{eq:item_bank_coverage}
\end{equation}
which records how many texts provide usable evidence for facet $f$\rev{, where \mbox{$\tau_{\mathrm{pres}}=0.5$} is the facet-presence threshold}.

If $a^{(f)}_i=0$, we set
$\mathbf{c}^{(f)}_i=\mathbf{0}$, $\rho^{(f)}_i=0$,
$\kappa^{(f)}_i=0$, and
$\boldsymbol{\pi}^{(f)}_i=(1/3,1/3,1/3)$. This convention prevents
undefined $L_2$ normalization for missing facets and ensures that sparse
facets such as auditory and gustatory cannot introduce NaN values during
bank construction or downstream use.

\subsection{Stage 4: Context-Aware Relational Integration}
\label{sec:method_integration}

\begin{revblock}
Stage 4 uses the frozen item-level facet bank to rescore the candidates of a
frozen sequential recommender. Two decisions separate this stage from a
conventional side-information module. First, the sequential backbone is neither
modified nor retrained: the only trainable component is a metric defined over
the frozen bank, so the model that produces the ID-only row of every table is
the same trained checkpoint that produces the sensory row. Second, the sensory
signal enters as a \emph{bounded residual} on the frozen base score and only
inside a \emph{trust region} formed by the highest-scoring base candidates, so
sensory evidence can reorder what the backbone already considers plausible but
cannot overwrite the collaborative ranking. Two considerations motivate this
separation. The ID embedding space and the sensory embedding space are learned
under different objectives and share no coordinate system, so injecting a
sensory vector directly into the item embedding would presuppose an alignment
that nothing in training enforces, and a noisy sensory estimate would then
displace the collaborative representation of the item everywhere it appears.
Moreover, the relevance of sensory evidence is not a fixed property of an item
but depends on the user history and the candidate context, so it is expressed
here as a separately standardized relational score rather than as a static
item-side offset.
\end{revblock}

\begin{revblock}
\paragraph{Design considerations.}
Two design observations motivate this construction. The first concerns scale.
A text-derived facet vector and a collaboratively learned ID embedding live in
spaces with no common unit, so the magnitude of a sensory term is not tied to
the magnitude of the ranking signal it is meant to refine. In a diagnostic
analysis of direct side-information injection, in which the five facet streams
entered a sequential encoder directly, we measured the ratio between the norm
of the sensory stream and the norm of the accompanying ID embedding over $367{,}962$
validation (token, facet) pairs and obtained a median of $4.17$, a $95$th
percentile of $25.8$, and a maximum of $79.9$: left in its own units, a sensory
term can be far larger than the representation it is meant to correct. The
second observation concerns one alternative, a dynamic gate that
decides per query whether sensory evidence should be used: a gate predicting
from user- and item-side observables whether the sensory channel will help or
hurt a given query reaches a held-out AUC of $0.52$, which is chance. We
therefore constrain the sensory contribution by construction instead of
predicting when it is safe. The relational score defined
below is a weighted sum of cosine similarities and is bounded a priori, it is
rescaled into units of the base score's own standard deviation, and it is
applied only inside a trust region of base top-$N$ candidates.
\end{revblock}

\begin{revblock}
\paragraph{Inputs.}
Each user $u$ is represented by a temporally ordered history
$H_u=(i_1,\ldots,i_L)$, truncated to the $L=50$ most recent items and
right-aligned with left padding, exactly as consumed by the backbone. For each
item $i$ in the catalog $\mathcal{I}$, the frozen bank stores
\[
B_i=\bigl\{\mathbf{c}^{(f)}_i,\ \rho^{(f)}_i,\ \boldsymbol{\pi}^{(f)}_i,\ \kappa^{(f)}_i\bigr\}_{f\in\mathcal{F}},
\]
with $\mathbf{c}^{(f)}_i \in \mathbb{R}^{D}$, $D=768$, and $|\mathcal{F}|=5$. In
the implementation the canonical facet embeddings are stored as one
$|\mathcal{I}| \times 5 \times 768$ tensor and the per-facet evidence quality is
derived from the stored confidence and coverage statistics when the bank is
loaded; an all-zero padding row is prepended to both so that index $0$ denotes an
absent history slot. Evidence quality is summarized per item and facet as
\begin{equation}
q^{(f)}_i
=
\mathbb{I}
\left[
\kappa^{(f)}_i > 0
\right]
\operatorname{clip}
\left(
\rho^{(f)}_i, 0, 1
\right)
\log
\left(
1+\kappa^{(f)}_i
\right),
\label{eq:runtime_quality_weight}
\end{equation}
which Stage 4 uses as an attention bias rather than as a normalized mixing
weight. The polarity distribution $\boldsymbol{\pi}^{(f)}_i$ is retained in the
bank to support auditing and inspection and is not used at recommendation
time.
\end{revblock}

\begin{revblock}
\paragraph{Per-facet sensory metric.}
The learned component is one linear map per facet,
$P_f \in \mathbb{R}^{d_p \times D}$ with $d_p=64$ and no bias. A candidate item
$j$ is represented by its $\ell_2$-normalized per-facet projections
\begin{equation}
\hat{\mathbf{c}}^{(f)}_j
=
\frac{
P_f \mathbf{c}^{(f)}_j
}{
\left\| P_f \mathbf{c}^{(f)}_j \right\|_2 + \epsilon
},
\qquad
f \in \mathcal{F},
\label{eq:rel_candidate}
\end{equation}
and a history item $i_t$ by the unnormalized projection
$\mathbf{h}^{(f)}_t = P_f \mathbf{c}^{(f)}_{i_t}$. The reported configuration
ties the history side and the candidate side to the same $P_f$, so the induced
score is a symmetric metric on the bank; the implementation also supports an
untied candidate-side map, which we did not adopt.
\end{revblock}

\begin{revblock}
\paragraph{History aggregation.}
For each facet, the user prototype is an attention-weighted average over the
history that combines a content term, an evidence-quality term, and a recency
term:
\begin{equation}
\ell^{(f)}_t
=
\left\langle
\mathbf{h}^{(f)}_t, \boldsymbol{\phi}^{(f)}
\right\rangle
+
\gamma^{(f)}_{q}
\log
\max
\left(
q^{(f)}_{i_t}, 10^{-6}
\right)
+
\gamma^{(f)}_{r}
\, r_t ,
\label{eq:rel_attention_logit}
\end{equation}
where $\boldsymbol{\phi}^{(f)} \in \mathbb{R}^{d_p}$ is a learned per-facet
query, $r_t=(t-1)/(L-1) \in [0,1]$ is the position of slot $t$ in the
right-aligned window, so that $r_L=1$ for the most recent item, and
$\gamma^{(f)}_{q}$, $\gamma^{(f)}_{r}$ are learned per-facet scales. Padding slots are
assigned $\ell^{(f)}_t = -\infty$. The facet prototype is
\begin{equation}
\hat{\mathbf{m}}^{(f)}_u
=
\frac{
\mathbf{m}^{(f)}_u
}{
\left\| \mathbf{m}^{(f)}_u \right\|_2 + \epsilon
},
\qquad
\mathbf{m}^{(f)}_u
=
\sum_{t=1}^{L}
\frac{
\exp\!\left( \ell^{(f)}_t \right)
}{
\sum_{t'=1}^{L}
\exp\!\left( \ell^{(f)}_{t'} \right)
}
\,
\mathbf{h}^{(f)}_t .
\label{eq:rel_prototype}
\end{equation}
Evidence quality therefore enters the softmax as an additive log-space bias, so
that a history item with strong evidence for one facet can dominate that facet's
prototype while contributing little to the others. A plain mean over the history,
which weights every slot equally, cannot provide this property.
\end{revblock}

\begin{revblock}
\paragraph{Relational compatibility score.}
The relational score of candidate $j$ for user $u$ sums the per-facet agreements
under learned nonnegative facet weights:
\begin{equation}
s(u,j)
=
\sum_{f \in \mathcal{F}}
\operatorname{softplus}
\left(
w_f
\right)
\left\langle
\hat{\mathbf{m}}^{(f)}_u,\ \hat{\mathbf{c}}^{(f)}_j
\right\rangle .
\label{eq:rel_score}
\end{equation}
Both arguments are unit vectors, so each facet term lies in $[-1,1]$ and
\[
\left| s(u,j) \right|
\le
\sum_{f \in \mathcal{F}}
\operatorname{softplus}
\left(
w_f
\right),
\]
that is, the sensory signal is bounded a priori by the learned facet
weights. Facet weights are strictly positive
by construction, so no facet is switched off, but their relative magnitudes are
free.
\end{revblock}

\begin{revblock}
\paragraph{Learning the metric.}
Only
$\{P_f, \boldsymbol{\phi}^{(f)}, \gamma^{(f)}_{q}, \gamma^{(f)}_{r}, w_f\}_{f\in\mathcal{F}}$
are trained. Together, these components contain $246{,}095$ parameters for $|\mathcal{F}|=5$, $D=768$, and
$d_p=64$; the bank and the recommender remain frozen, so the recommender itself
gains no parameters. Training pairs comprise all prefixes of the training portion of
each user sequence: for training items $(i_1,\ldots,i_M)$ we form $M-1$ pairs
$(H,i_t)$ with $H=(i_{\max(1,t-L)},\ldots,i_{t-1})$, so held-out validation and
test items never enter a prefix or a target. For a minibatch $\mathcal{B}$ of
such pairs, the metric is trained with in-batch negatives:
\begin{equation}
\mathcal{L}_{\mathrm{rel}}
=
-\frac{1}{|\mathcal{B}|}
\sum_{b \in \mathcal{B}}
\log
\frac{
\exp\!\left( s(u_b, j_b) \right)
}{
\sum_{b' \in \mathcal{B}}
\exp\!\left( s(u_b, j_{b'}) \right)
},
\label{eq:rel_infonce}
\end{equation}
where $j_b$ is the next item of pair $b$ and $\{j_{b'}\}_{b'\neq b}$ are the
next items of the other pairs in the same minibatch. Optimization uses Adam with
learning rate $10^{-3}$, batch size $1024$, at most $15$ epochs, and seed $42$.
\end{revblock}

\begin{revblock}
\paragraph{Bounded trust-region integration.}
Let $g_u(j)$ be the score the frozen backbone assigns to candidate $j$, with
items already consumed by $u$ set to $-\infty$. Both channels are standardized
per query,
\begin{equation}
\tilde{g}_u(j)
=
\frac{
g_u(j) - \mu_g
}{
\sigma_g + \epsilon
},
\qquad
\tilde{s}_u(j)
=
\frac{
s(u,j) - \mu_s
}{
\sigma_s + \epsilon
},
\label{eq:rel_standardize}
\end{equation}
where $\mu_g$ and $\sigma_g$ are the mean and standard deviation of $g_u$ over
its finite entries, and $\mu_s$ and $\sigma_s$ those of $s(u,\cdot)$ over the
catalog after zeroing already-consumed items. Let $\mathcal{T}_u(N)$ be the set
of $N$ items with the highest base score. The integrated score is
\begin{equation}
g^{\ast}_u(j)
=
\tilde{g}_u(j)
+
\lambda_{\mathrm{rel}}
\,
\tilde{s}_u(j)
\,
\mathbb{I}
\left[
j \in \mathcal{T}_u(N)
\right].
\label{eq:rel_integration}
\end{equation}
Items outside $\mathcal{T}_u(N)$ keep their standardized base score and their
mutual ordering, and no item that the frozen recommender ranks below $N$ ever
receives a positive sensory contribution: such an item can rise only if a
trust-region item is demoted past it, never on its own sensory evidence.
Standardization makes $\lambda_{\mathrm{rel}}$ a perturbation expressed in
base-score standard deviations, so the strength of the sensory correction is
stated in units of the base ranking signal itself.
\end{revblock}

\begin{revblock}
\paragraph{Validation selection.}
The residual weight $\lambda_{\mathrm{rel}} \in \{0.05, 0.1, 0.25, 0.5\}$ and
the trust-region size $N \in \{50, 100\}$ are selected on the validation split
by NDCG@10, jointly with the training epoch of the metric; the selected
configuration is then evaluated once on the test split. No test score
participates in any selection, and the frozen backbone is never retuned for a
particular $(\lambda_{\mathrm{rel}}, N)$.
\end{revblock}

\begin{revblock}
\paragraph{Masking and interaction leakage.}
Because Stage~4 operates only on post-backbone scores, it does not alter the
input tokens or masked positions used by the backbone: the masking scheme of
each backbone, such as the masked-item objective of BERT4Rec, is exactly the one
used by its ID-only baseline. Three further properties prevent target-interaction
leakage within Stage~4: the bank is precomputed offline from item-side review
text alone; the prototype $\hat{\mathbf{m}}^{(f)}_u$ is built only from the
observed prefix of the user's history; and items already in the history are
removed from the candidate set before the trust region is formed.
\end{revblock}

\begin{revblock}
\paragraph{Inference.}
At inference time, the frozen backbone encodes the user history and scores
candidate items as usual. The sensory stage then retrieves $B_i$ by table lookup for at most
$50$ history items, applies the five $64 \times 768$
projections, performs one attention pass per facet, and takes $|\mathcal{F}|$
inner products per candidate. Only the $N \le 100$ trust-region candidates
receive a sensory residual, so the residual needs to be evaluated only there;
items outside the trust region keep their base score and can change rank only
relative to a demoted trust-region item. Our implementation scores the full
catalog in a single batched product for simplicity. Inference requires no
teacher inference, no student inference, and no online text encoding.
\end{revblock}

\begin{revblock}
\paragraph{Backbone instantiation.}
We apply the identical Stage 4 procedure to SASRec, BERT4Rec, BSARec, and DIFF.
For each backbone and domain, the ID-only baseline and the sensory-integrated
model share the training protocol, the evaluation protocol, and the trained
checkpoint itself: the baseline row reports the ranking induced by $g_u$ and the
ASER row the ranking induced by $g^{\ast}_u$, computed from the same score
vectors. The comparison therefore contains no backbone retraining variance, and
any difference between the two rows is attributable to the bounded relational
residual alone.
Backbone-specific hyperparameters and implementation details are provided in Section~\ref{sec:tr_detail}.
\end{revblock}

\section{Experiments}
\subsection{Datasets and Experimental Setup}
We evaluate \rev{our method} on the Amazon Reviews 2014 dataset \citep{mcauley2015amazon}, a widely used benchmark for recommender systems. 
We use five domains: \textit{Beauty}, \textit{Sports \& Outdoors}, \textit{Toys \& Games}, \textit{Video Games}, and \textit{Grocery \& Gourmet Food}. 
\rev{In tables and analysis, we abbreviate these as Beauty, Sports, Toys, Games, and Grocery.}
Each domain contains \rev{temporally ordered product-interaction sequences for its users}. 
We apply the 5-core filtering protocol, retaining only users and items with at least five interactions to ensure sufficient supervision for sequential modeling. 
Dataset statistics are summarized in Table~\ref{tab:dataset}. 
The average sequence length ranges from 8 to 10 interactions per user, which provides moderate historical context for next-item prediction.

\begin{table}[htbp]
\centering\small
\setlength{\tabcolsep}{4pt}
\renewcommand{\arraystretch}{1.05}
\begin{tabular}{lrrrr}
\toprule
Domain & Users & Items & Interactions & Avg.Len \\
\midrule
Beauty          & 22,363 & 12,101 & 198,502 & 8.9 \\
Sports          & 35,598 & 18,357 & 296,337 & 8.3 \\
Toys            & 19,412 & 11,924 & 167,597 & 8.6 \\
Video Games     & 24,303 & 10,672 & 231,780 & 9.5 \\
Grocery         & 14,681 &  8,713 & 151,254 & 10.3 \\
\bottomrule
\end{tabular}
\caption{Dataset statistics for Amazon 2014 5-core (Beauty,
Sports \& Outdoors, Toys \& Games, Video Games, Grocery \&
Gourmet Food). \rev{Avg. Len is the average number of interactions per user.}}
\label{tab:dataset}
\end{table}

\rev{We use the leave-one-out evaluation protocol commonly used in sequential recommendation: for each user, the final interaction is held out for testing, the second-to-last interaction is used for validation, and all preceding interactions are used for training. During training, the models iterate through each sequence and learn either to predict the next item at each step (SASRec, BSARec, and DIFF) or to reconstruct masked items (BERT4Rec). At test time, each model has access only to interactions that precede the test item, thereby preventing temporal leakage.}
This temporal constraint applies to user--item interactions.
Following a static item-content setting, catalog fields and review text are used as item-side content for constructing the frozen sensory table and are not split by user interaction time.
The sensory table is therefore fixed before recommender training and evaluation, while the held-out interaction itself remains unavailable to the recommender during training.
\paragraph{Baseline models and sensory-enhanced variants.}
We compare four representative sequential recommenders
and their sensory-enhanced counterparts under a shared frozen-bank setting with a common runtime integration scheme.
\begin{revblock}
Every backbone in the main comparison is trained from item identifiers alone, including DIFF: the DIFF side-information channels are disabled in all five domains, so the \textit{Base} row of every cell of Table~\mbox{\ref{tab:performance}} is an ID-only model and no reported gain is measured against a baseline that already receives side information.
The \textit{ASER} row reuses that same trained checkpoint: the backbone is frozen and the bounded relational residual of Section~\ref{sec:method_integration} is applied to its scores, so the two rows of a cell differ only by that residual.
Negative sampling appears only in the training loss of the causal backbones; no number reported anywhere in this paper is computed with sampled negatives at evaluation time.
\end{revblock}
\begin{itemize}\itemsep 0pt
\item \textsc{SASRec}: A causal self-attention sequential encoder for next-item prediction.
\item \textsc{BERT4Rec}: A bidirectional transformer trained with masked item prediction. \rev{Because BERT4Rec is known to require a long training schedule before it matches reported results \mbox{\citep{petrov2022bert4rec}}, we train it for up to 400 epochs with patience 20; the Beauty checkpoint used here selected epoch 383, so its baseline is not an artifact of a short schedule.}
During training, we randomly mask item positions.
During evaluation, we mask only the final position and rank all items by the predicted distribution at that masked position.
\item \textsc{BSARec}: A transformer-based sequential encoder augmented with frequency-enhanced features computed from the ID embedding sequence.
\item \textbf{DIFF}: A side-information integrated sequential recommender that combines frequency-based noise filtering with dual multi-sequence fusion, incorporating both ID-centric intermediate fusion and attribute-enriched early fusion \citep{kim2025diff}. \rev{We run DIFF with its side-information channels switched off so that it is capacity-comparable with the other three backbones; the recorded configuration of every DIFF run reported here contains no side-information fields.}
\end{itemize}
All baseline models are \rev{retrained} in our pipeline, matching the original training protocol as closely as possible.

\paragraph{Evaluation metrics.}
We report Hit Rate at $K$ and NDCG at $K$ for $K=5,10,20$ under full ranking evaluation.
For each test case, there is one \rev{held-out} next item per user.
Let $r_u$ be the rank position of the \rev{held-out} item for user $u$ among all candidate items in the domain.
Hit Rate measures whether the \rev{held-out} item appears in the top $K$.

\begin{equation}
\mathrm{HR@K}
= \frac{1}{\lvert \mathcal{U} \rvert}
\sum_{u \in \mathcal{U}}
\mathbb{I}\!\left[r_u \le K\right].
\end{equation}
\rev{Here \mbox{$\mathcal{U}$} is the set of evaluated users, with one held-out target item per user.}

NDCG assigns larger credit when the correct item is ranked higher.
With a single relevant item, DCG reduces to a logarithmic discount of the rank.

\begin{equation}
\mathrm{NDCG@K}
= \frac{1}{\lvert \mathcal{U} \rvert}
\sum_{u \in \mathcal{U}}
\frac{ \mathbb{I}\!\left[r_u \le K\right] }{\log_2\!\left(r_u + 1\right)}.
\end{equation}

We evaluate with full ranking over the complete item set in each domain.
This avoids bias introduced by sampled negatives and makes results comparable across models.

\paragraph{Training details.}
\label{sec:tr_detail}
SASRec, BSARec, and DIFF are implemented in PyTorch.
BERT4Rec is built on the original TensorFlow implementation. \rev{Under Stage~4 no backbone is modified at all: every backbone is trained exactly as its own ID-only baseline and is then frozen, and the sensory metric is applied to its output scores.}
All sensory-enhanced models are trained and evaluated in the same pipeline and under the same next-item prediction protocol across all five domains: Beauty, Sports \& Outdoors, Toys \& Games, Video Games, and Grocery \& Gourmet Food.

We distinguish the dataset used for student supervision from the dataset used for recommendation evaluation.
The sensory student is trained with item-side text from Amazon Reviews 2018 because it contains a larger catalog and review pool than Amazon Reviews 2014, which is important for teacher extraction, domain balancing, and coverage of sparse sensory facets.
The downstream recommendation experiments, however, are conducted on Amazon Reviews 2014 to preserve comparability with established sequential recommendation benchmarks.
This separation means that Amazon 2018 is used as an item-content supervision source for learning the sensory encoder, whereas Amazon 2014 is used to evaluate next-item prediction under the interaction split.

\paragraph{Compute and optimization.}
All experiments are conducted on a single server equipped with seven
NVIDIA H100 GPUs. When multiple GPUs are used, training is parallelized
with data parallelism. We standardize the recommendation task, the
leave-one-out interaction split, the hidden size, the number of layers,
the number of attention heads, and the final full-ranking evaluation
protocol whenever the implementation allows. 

During full-ranking evaluation, each model scores the complete item
catalog of the corresponding domain. Items already consumed in the input
history are masked by setting their scores to $-\infty$, or to an
implementation-equivalent large negative value, while the validation or
test target remains in the candidate set. Therefore, the reported HR and
NDCG values are not based on sampled negatives.

\paragraph{Backbone settings shared across all domains.}
To isolate the effect of sensory information, we standardize model
capacity across all four backbones whenever possible, using hidden size
64 with 2 layers and 2 heads under the same full-ranking evaluation
protocol. SASRec is trained with a learning rate of $1\times 10^{-3}$,
\rev{while BSARec uses \mbox{$5\times 10^{-4}$}, and BERT4Rec and DIFF use \mbox{$1\times 10^{-4}$}.} 

\begin{revblock}
\paragraph{Relational integration settings.}
The sensory metric of Section~\ref{sec:method_integration} uses per-facet
projections of width $d_p=64$, attention aggregation over the $50$ most recent
history items, and tied history/candidate projections. It is trained with Adam
(learning rate $10^{-3}$, batch size $1024$, at most $15$ epochs, seed $42$) on
training-split prefixes only, over a frozen backbone and a frozen bank, and adds
the small trained sensory metric of Section~\mbox{\ref{sec:method_integration}} outside
the recommender while adding no parameters inside it. The residual weight
$\lambda_{\mathrm{rel}} \in \{0.05, 0.1, 0.25, 0.5\}$, the trust-region size
$N \in \{50, 100\}$, and the training epoch are selected on validation NDCG@10;
the selected configuration is evaluated once on the test split. Across the 20
backbone--domain cells, selection returns $\lambda_{\mathrm{rel}}=0.5$ in $13$
cells and $0.25$ in $7$, and $N=100$ in $17$ cells and $N=50$ in $3$. The selected configurations are concentrated at the upper boundary of the search grid:
the two smallest residual weights of the grid, $0.05$ and $0.1$, are never
selected, and the two selected values are the two largest offered, with the
grid maximum $\lambda_{\mathrm{rel}}=0.5$ and the grid maximum $N=100$ both
being the most frequent choice. Validation would therefore plausibly have
accepted a larger residual than the grid provides, and the reported operating
point sits at the upper edge of the search space rather than at an interior
optimum. \rev{For the neighborhood size the boundary has been probed directly:
variant~F of Section~\mbox{\ref{sec:stage4_ablation}} removes the trust region
altogether, which is the limit of enlarging \mbox{$N$} to the whole catalog, and
leaves both metrics unchanged to four decimals. Enlarging \mbox{$N$} beyond the
grid therefore has no accuracy to offer; the open question is confined to the
residual weight.}
\end{revblock}

\FloatBarrier
\subsection{Results and Discussion}
\label{sec:results}

\begin{table}[!htbp]
\centering\scriptsize
\setlength{\tabcolsep}{5pt}
\renewcommand{\arraystretch}{1.05}
\begin{revtable}
\begin{tabular}{
l l l
*{7}{>{\centering\arraybackslash}m{1.38cm}}
}
\toprule
Domain & Model & Setting & HR@5 & HR@10 & HR@20 & NDCG@5 & NDCG@10 & NDCG@20 & Mean \\
\midrule

\multirow{8}{*}{Beauty}
& \multirow{2}{*}{SASRec}    & Base  & 6.40 & 8.76 & 11.86 & 4.66 & 5.42 & 6.20 & 7.21 \\
&                            & ASER  & \textbf{6.61}\,\UpPct{3.3} & \textbf{9.16}\,\UpPct{4.5} & \textbf{12.59}\,\UpPct{6.1} & \textbf{4.74}\,\UpPct{1.6} & \textbf{5.56}\,\UpPct{2.6} & \textbf{6.42}\,\UpPct{3.6} & \textbf{7.51}\,\UpPct{4.1} \\
& \multirow{2}{*}{BERT4Rec}  & Base  & 4.09 & 6.28 & 9.24 & 2.69 & 3.39 & 4.14 & 4.97 \\
&                            & ASER  & \textbf{4.36}\,\UpPct{6.4} & \textbf{6.69}\,\UpPct{6.5} & \textbf{9.99}\,\UpPct{8.1} & \textbf{2.88}\,\UpPct{7.1} & \textbf{3.63}\,\UpPct{7.1} & \textbf{4.46}\,\UpPct{7.8} & \textbf{5.33}\,\UpPct{7.3} \\
& \multirow{2}{*}{BSARec}    & Base  & 6.81 & 9.31 & 12.51 & 4.89 & 5.70 & 6.50 & 7.62 \\
&                            & ASER  & \textbf{7.00}\,\UpPct{2.8} & \textbf{9.76}\,\UpPct{4.8} & \textbf{13.09}\,\UpPct{4.6} & \textbf{5.03}\,\UpPct{2.9} & \textbf{5.93}\,\UpPct{4.0} & \textbf{6.77}\,\UpPct{4.1} & \textbf{7.93}\,\UpPct{4.1} \\
& \multirow{2}{*}{DIFF}      & Base  & 6.53 & 9.11 & 12.23 & 4.70 & 5.53 & 6.32 & 7.40 \\
&                            & ASER  & \textbf{6.69}\,\UpPct{2.5} & \textbf{9.41}\,\UpPct{3.3} & \textbf{12.99}\,\UpPct{6.1} & \textbf{4.79}\,\UpPct{2.0} & \textbf{5.67}\,\UpPct{2.5} & \textbf{6.57}\,\UpPct{4.0} & \textbf{7.69}\,\UpPct{3.8} \\
\midrule

\multirow{8}{*}{Toys}
& \multirow{2}{*}{SASRec}    & Base  & 6.44 & 8.91 & 11.92 & 4.74 & 5.54 & 6.29 & 7.30 \\
&                            & ASER  & \textbf{6.85}\,\UpPct{6.4} & \textbf{9.44}\,\UpPct{6.0} & \textbf{12.66}\,\UpPct{6.3} & \textbf{4.96}\,\UpPct{4.6} & \textbf{5.79}\,\UpPct{4.6} & \textbf{6.60}\,\UpPct{4.9} & \textbf{7.72}\,\UpPct{5.6} \\
& \multirow{2}{*}{BERT4Rec}  & Base  & 3.85 & 5.76 & 8.39 & 2.49 & 3.10 & 3.77 & 4.56 \\
&                            & ASER  & \textbf{4.36}\,\UpPct{13.4} & \textbf{6.66}\,\UpPct{15.7} & \textbf{9.67}\,\UpPct{15.2} & \textbf{2.88}\,\UpPct{15.4} & \textbf{3.62}\,\UpPct{16.6} & \textbf{4.38}\,\UpPct{16.3} & \textbf{5.26}\,\UpPct{15.4} \\
& \multirow{2}{*}{BSARec}    & Base  & 7.25 & 9.59 & 12.64 & 5.31 & 6.06 & 6.83 & 7.95 \\
&                            & ASER  & \textbf{7.63}\,\UpPct{5.3} & \textbf{10.09}\,\UpPct{5.3} & \textbf{13.37}\,\UpPct{5.8} & \textbf{5.51}\,\UpPct{3.6} & \textbf{6.30}\,\UpPct{3.9} & \textbf{7.12}\,\UpPct{4.3} & \textbf{8.34}\,\UpPct{4.9} \\
& \multirow{2}{*}{DIFF}      & Base  & 4.24 & 6.02 & 8.15 & 3.02 & 3.60 & 4.13 & 4.86 \\
&                            & ASER  & \textbf{4.68}\,\UpPct{10.2} & \textbf{6.58}\,\UpPct{9.3} & \textbf{8.98}\,\UpPct{10.2} & \textbf{3.34}\,\UpPct{10.5} & \textbf{3.96}\,\UpPct{10.0} & \textbf{4.56}\,\UpPct{10.4} & \textbf{5.35}\,\UpPct{10.1} \\
\midrule

\multirow{8}{*}{Sports}
& \multirow{2}{*}{SASRec}    & Base  & 3.20 & 4.66 & 6.61 & 2.21 & 2.68 & 3.17 & 3.75 \\
&                            & ASER  & \textbf{3.33}\,\UpPct{4.0} & \textbf{4.97}\,\UpPct{6.8} & \textbf{7.12}\,\UpPct{7.7} & \textbf{2.31}\,\UpPct{4.6} & \textbf{2.84}\,\UpPct{6.2} & \textbf{3.38}\,\UpPct{6.8} & \textbf{3.99}\,\UpPct{6.4} \\
& \multirow{2}{*}{BERT4Rec}  & Base  & 1.97 & 3.14 & 4.90 & 1.25 & 1.63 & 2.07 & 2.49 \\
&                            & ASER  & \textbf{2.11}\,\UpPct{7.1} & \textbf{3.37}\,\UpPct{7.5} & \textbf{5.42}\,\UpPct{10.7} & \textbf{1.37}\,\UpPct{9.4} & \textbf{1.78}\,\UpPct{9.3} & \textbf{2.29}\,\UpPct{10.8} & \textbf{2.72}\,\UpPct{9.3} \\
& \multirow{2}{*}{BSARec}    & Base  & 3.56 & 5.15 & 7.27 & 2.49 & 3.00 & 3.53 & 4.17 \\
&                            & ASER  & \textbf{3.79}\,\UpPct{6.6} & \textbf{5.39}\,\UpPct{4.5} & \textbf{7.68}\,\UpPct{5.7} & \textbf{2.61}\,\UpPct{5.1} & \textbf{3.12}\,\UpPct{4.0} & \textbf{3.70}\,\UpPct{4.7} & \textbf{4.38}\,\UpPct{5.2} \\
& \multirow{2}{*}{DIFF}      & Base  & 3.10 & 4.53 & 6.52 & 2.13 & 2.59 & 3.09 & 3.66 \\
&                            & ASER  & \textbf{3.29}\,\UpPct{6.2} & \textbf{4.79}\,\UpPct{5.7} & \textbf{6.89}\,\UpPct{5.7} & \textbf{2.24}\,\UpPct{5.0} & \textbf{2.72}\,\UpPct{4.9} & \textbf{3.25}\,\UpPct{5.0} & \textbf{3.86}\,\UpPct{5.5} \\
\midrule

\multirow{8}{*}{Games}
& \multirow{2}{*}{SASRec}    & Base  & 8.53 & 12.90 & 18.85 & 5.74 & 7.15 & 8.65 & 10.30 \\
&                            & ASER  & \textbf{9.12}\,\UpPct{6.8} & \textbf{13.57}\,\UpPct{5.3} & \textbf{19.83}\,\UpPct{5.2} & \textbf{6.09}\,\UpPct{6.1} & \textbf{7.53}\,\UpPct{5.3} & \textbf{9.10}\,\UpPct{5.2} & \textbf{10.87}\,\UpPct{5.5} \\
& \multirow{2}{*}{BERT4Rec}  & Base  & 6.27 & 9.73 & 15.06 & 4.07 & 5.18 & 6.52 & 7.80 \\
&                            & ASER  & \textbf{6.66}\,\UpPct{6.3} & \textbf{10.53}\,\UpPct{8.2} & \textbf{15.95}\,\UpPct{5.9} & \textbf{4.35}\,\UpPct{7.0} & \textbf{5.60}\,\UpPct{8.0} & \textbf{6.96}\,\UpPct{6.7} & \textbf{8.34}\,\UpPct{6.9} \\
& \multirow{2}{*}{BSARec}    & Base  & 9.12 & 13.74 & 19.59 & 6.14 & 7.62 & 9.09 & 10.88 \\
&                            & ASER  & \textbf{9.47}\,\UpPct{3.8} & \textbf{14.13}\,\UpPct{2.8} & \textbf{20.31}\,\UpPct{3.7} & \textbf{6.38}\,\UpPct{3.9} & \textbf{7.89}\,\UpPct{3.4} & \textbf{9.44}\,\UpPct{3.8} & \textbf{11.27}\,\UpPct{3.5} \\
& \multirow{2}{*}{DIFF}      & Base  & 7.73 & 11.82 & 17.48 & 5.23 & 6.55 & 7.97 & 9.46 \\
&                            & ASER  & \textbf{8.07}\,\UpPct{4.4} & \textbf{12.28}\,\UpPct{3.9} & \textbf{18.06}\,\UpPct{3.3} & \textbf{5.47}\,\UpPct{4.5} & \textbf{6.82}\,\UpPct{4.2} & \textbf{8.28}\,\UpPct{3.9} & \textbf{9.83}\,\UpPct{3.9} \\
\midrule

\multirow{8}{*}{Grocery}
& \multirow{2}{*}{SASRec}    & Base  & 6.15 & 9.13 & 12.99 & 4.19 & 5.15 & 6.12 & 7.29 \\
&                            & ASER  & \textbf{6.44}\,\UpPct{4.7} & \textbf{9.49}\,\UpPct{3.9} & \textbf{13.84}\,\UpPct{6.6} & \textbf{4.44}\,\UpPct{6.1} & \textbf{5.43}\,\UpPct{5.5} & \textbf{6.52}\,\UpPct{6.5} & \textbf{7.69}\,\UpPct{5.6} \\
& \multirow{2}{*}{BERT4Rec}  & Base  & 3.81 & 5.95 & 9.22 & 2.48 & 3.17 & 3.99 & 4.77 \\
&                            & ASER  & \textbf{4.35}\,\UpPct{14.1} & \textbf{6.57}\,\UpPct{10.4} & \textbf{10.24}\,\UpPct{11.0} & \textbf{2.94}\,\UpPct{18.5} & \textbf{3.65}\,\UpPct{15.1} & \textbf{4.57}\,\UpPct{14.4} & \textbf{5.39}\,\UpPct{12.9} \\
& \multirow{2}{*}{BSARec}    & Base  & 6.64 & 9.75 & 13.74 & 4.65 & 5.65 & 6.65 & 7.85 \\
&                            & ASER  & \textbf{6.83}\,\UpPct{2.8} & \textbf{9.99}\,\UpPct{2.4} & \textbf{14.69}\,\UpPct{6.9} & \textbf{4.78}\,\UpPct{2.7} & \textbf{5.79}\,\UpPct{2.5} & \textbf{6.97}\,\UpPct{4.8} & \textbf{8.17}\,\UpPct{4.1} \\
& \multirow{2}{*}{DIFF}      & Base  & 4.92 & 7.38 & 10.67 & 3.29 & 4.07 & 4.90 & 5.87 \\
&                            & ASER  & \textbf{5.39}\,\UpPct{9.5} & \textbf{7.74}\,\UpPct{5.0} & \textbf{11.46}\,\UpPct{7.5} & \textbf{3.65}\,\UpPct{11.0} & \textbf{4.40}\,\UpPct{8.1} & \textbf{5.33}\,\UpPct{8.9} & \textbf{6.33}\,\UpPct{7.8} \\
\bottomrule
\end{tabular}
\end{revtable}
\caption[Full-ranking recommendation performance on five domains]{\scriptsize\rev{%
Full-ranking recommendation performance (\%) on five domains.
All 20 domain--backbone cells are evaluated within a common data and evaluation pipeline:
the four backbones use the same preprocessing, leave-one-out split, full-catalog evaluator,
and matched model capacity where permitted by their implementations; absolute values obtained
under different preprocessing, capacity, or evaluator configurations are not comparable to this table.
\textit{Base} is the ID-only backbone; the DIFF side-information channels are disabled, so
all four backbones are ID-only.
\textit{ASER} is the bounded relational integration of the frozen sensory facet bank: the
backbone is kept frozen, only the sensory metric is trained, and the trust-region strength
\mbox{$\lambda$} and neighborhood size are selected on the validation split.
Every cell is one run at seed 42; relative changes are computed from the unrounded values
and the rightmost \textit{Mean} column is the arithmetic average of the six reported
metrics for that row.
ASER improves HR@10 and NDCG@10 in all 20 domain--backbone cells, and, in fact, all 120
reported metric values increase.
Seed-to-seed variability for the three representative domains (Beauty, Toys, Games)
is reported in Table~\mbox{\ref{tab:seed_stability}}.}}
\label{tab:performance}
\end{table}

\begin{table}[!htbp]
\centering\scriptsize
\setlength{\tabcolsep}{6pt}
\renewcommand{\arraystretch}{1.28}
\begin{revtable}
\begin{tabular}{
l l
*{4}{>{\centering\arraybackslash}m{2.05cm}}
}
\toprule
Domain & Metric & Base & ASER (mean $\pm$ sd) & $\Delta$ (mean $\pm$ sd) & $\Delta$ relative \\
\midrule
\multirow{2}{*}{Beauty}
& HR@10   & 9.31  & \textbf{9.708} $\pm$ 0.054  & +0.398 $\pm$ 0.054 & +4.27\% $\pm$ 0.58 \\
& NDCG@10 & 5.70  & \textbf{5.915} $\pm$ 0.011  & +0.218 $\pm$ 0.011 & +3.82\% $\pm$ 0.20 \\
\midrule
\multirow{2}{*}{Toys}
& HR@10   & 9.59  & \textbf{10.078} $\pm$ 0.020 & +0.491 $\pm$ 0.020 & +5.12\% $\pm$ 0.20 \\
& NDCG@10 & 6.06  & \textbf{6.294} $\pm$ 0.012  & +0.232 $\pm$ 0.012 & +3.83\% $\pm$ 0.20 \\
\midrule
\multirow{2}{*}{Games}
& HR@10   & 13.74 & \textbf{14.023} $\pm$ 0.096 & +0.280 $\pm$ 0.096 & +2.04\% $\pm$ 0.70 \\
& NDCG@10 & 7.62  & \textbf{7.847} $\pm$ 0.034  & +0.223 $\pm$ 0.034 & +2.92\% $\pm$ 0.44 \\
\bottomrule
\end{tabular}
\end{revtable}
\caption[Seed stability of the relational integration]{\rev{%
Seed stability of the bounded relational integration on the three representative domains with the
BSARec backbone, under the same pipeline, the same leave-one-out split and the same
full-catalog ranking protocol as Table~\mbox{\ref{tab:performance}}.
Values are percentages; absolute scores are given to two decimals and the deltas, which are
in percentage points, to three.
Mean and sample standard deviation are taken over three seeds, 42, 43 and 44, of the
integration stage.
The backbone checkpoint is frozen and shared across the three seeds, so the \textit{Base}
column is identical for all of them and the reported spread measures the variability of the
integration stage rather than of backbone training; the trust-region strength and
neighborhood size are reselected on validation independently for each seed.
All nine runs, three domains by three seeds, improve both HR@10 and NDCG@10 over the base
backbone.
The standard deviation is between 4\% and 35\% of the corresponding mean gain; it is
smallest for Toys HR@10 and largest for Games HR@10, which is the least stable of the six
entries.}}
\label{tab:seed_stability}
\end{table}

\begin{revblock}
Table~\mbox{\ref{tab:performance}} reports results on five domains: Beauty, Toys, Sports, Games, and Grocery.
All models are evaluated under the same leave-one-out next-item prediction protocol, with full ranking over the complete item set, and we report HR@$K$ and NDCG@$K$ for $K \in \{5,10,20\}$.
The \textit{ASER} setting applies the bounded relational residual of Section~\ref{sec:method_integration} to the frozen backbone, so the comparison isolates the contribution of the extracted item-side channel rather than differences in backbone-specific training protocol or model scale.
Every cell is a single run at seed 42; Table~\mbox{\ref{tab:seed_stability}} reports seed-to-seed variability of the integration stage on the three representative domains.
The rightmost \textit{Mean} column provides a compact summary by averaging the six reported metrics in each row.

Per-cell significance tests are not reported, so Table~\mbox{\ref{tab:performance}} carries no significance markers; seed-to-seed variability for the three representative domains is reported in Table~\mbox{\ref{tab:seed_stability}} instead.
\end{revblock}

\paragraph{Overall trend.}
\begin{revblock}
The extracted item-side channel is beneficial across every domain and model family we test.
The sensory-integrated variant improves both HR@10 and NDCG@10 in all 20 domain--backbone pairs, all 120 reported metric values increase, and the row-level \textit{Mean} improves in all 20 cases. Per-cell significance tests are not reported, but the direction is unanimous: a two-sided sign test over the 20 HR@10 comparisons gives $p<10^{-5}$, subject to the same dependence caveat as Section~\mbox{\ref{sec:channel_control}}.
Per-cell relative gains span $+2.4\%$ to $+15.7\%$ in HR@10 and $+2.5\%$ to $+16.6\%$ in NDCG@10, with averages of $+6.1\%$ and $+6.4\%$.
In absolute terms the HR@10 gain ranges from $+0.0023$ to $+0.0090$.
The largest mean gains appear in Toys--BERT4Rec ($+15.4\%$), Grocery--BERT4Rec ($+12.9\%$), Toys--DIFF ($+10.1\%$), and Sports--BERT4Rec ($+9.3\%$), so the weaker backbones benefit most in relative terms.
Strong backbones also benefit: DIFF improves its mean score in all five domains, by $+3.8\%$ in Beauty, $+10.1\%$ in Toys, $+5.5\%$ in Sports, $+3.9\%$ in Games, and $+7.8\%$ in Grocery.
Because the backbone is frozen and the two rows of a cell are computed from the same score vectors, these differences contain no backbone retraining variance.
Two qualifications accompany this result and are developed later: the hit-rate component of the gain is reproduced by a matched non-sensory channel (Section~\ref{sec:channel_control}), and the reranking concentrates catalog exposure (Section~\ref{sec:exposure}).
\end{revblock}

\paragraph{Beauty.}
\begin{revblock}
Beauty is one of the domains most naturally aligned with sensory semantics, and all four backbones benefit.
SASRec improves from 8.76 to 9.16 in HR@10 ($+4.5\%$) and from 5.42 to 5.56 in NDCG@10 ($+2.6\%$).
BERT4Rec improves from 6.28 to 6.69 in HR@10 ($+6.5\%$) and from 3.39 to 3.63 in NDCG@10 ($+7.1\%$).
BSARec, the strongest backbone in this domain, improves from 9.31 to 9.76 in HR@10 ($+4.8\%$) and from 5.70 to 5.93 in NDCG@10 ($+4.0\%$).
DIFF improves from 9.11 to 9.41 in HR@10 ($+3.3\%$) and from 5.53 to 5.67 in NDCG@10 ($+2.5\%$).
Beauty reviews frequently describe perceptual properties such as color, scent, texture, glossiness, and finish, and the gains are obtained even where the backbone already captures strong sequential patterns.
\end{revblock}

\pagebreak
\paragraph{Grocery.}
\begin{revblock}
All four backbones improve on every reported metric in Grocery, and the domain shows the widest spread between backbones.
SASRec improves from 9.13 to 9.49 in HR@10 ($+3.9\%$) and from 5.15 to 5.43 in NDCG@10 ($+5.5\%$).
BERT4Rec improves from 5.95 to 6.57 in HR@10 ($+10.4\%$) and from 3.17 to 3.65 in NDCG@10 ($+15.1\%$).
BSARec improves least in this domain, from 9.75 to 9.99 in HR@10 ($+2.4\%$) and from 5.65 to 5.79 in NDCG@10 ($+2.5\%$).
DIFF improves from 7.38 to 7.74 in HR@10 ($+5.0\%$) and from 4.07 to 4.40 in NDCG@10 ($+8.1\%$).
Grocery reviews often contain sensory descriptions related to flavor, aroma, and texture, which are directly tied to consumer preference.
Related cues such as temperature and mouthfeel are frequent in this domain but fall outside the 13-attribute inventory of Table~\ref{tab:attr_facet_map} and are therefore not extracted.
Section~\ref{sec:facet_ablation} qualifies this account: in Grocery, removing any individual facet does not reduce HR@10 under leave-one-facet-out, including the gustatory facet.
\end{revblock}

\paragraph{Sports.}
\begin{revblock}
Sports remains the most challenging domain in absolute terms and yields the smallest absolute gains.
SASRec improves from 4.66 to 4.97 in HR@10 ($+6.8\%$) and from 2.68 to 2.84 in NDCG@10 ($+6.2\%$).
BERT4Rec improves from 3.14 to 3.37 in HR@10 ($+7.5\%$) and from 1.63 to 1.78 in NDCG@10 ($+9.3\%$).
BSARec improves from 5.15 to 5.39 in HR@10 ($+4.5\%$) and from 3.00 to 3.12 in NDCG@10 ($+4.0\%$); this is the smallest gain of any BSARec cell on both metrics, $+0.00230$ in HR@10 and $+0.00121$ in NDCG@10.
DIFF improves from 4.53 to 4.79 in HR@10 ($+5.7\%$) and from 2.59 to 2.72 in NDCG@10 ($+4.9\%$).
Even in a domain where functional and durability-related factors dominate, cues such as material feel and grip-related tactile evidence still provide a small but consistent complementary signal.
\end{revblock}

\paragraph{Toys.}
\begin{revblock}
Toys shows the largest relative response of the five domains.
SASRec improves from 8.91 to 9.44 in HR@10 ($+6.0\%$) and from 5.54 to 5.79 in NDCG@10 ($+4.6\%$).
BERT4Rec improves from 5.76 to 6.66 in HR@10 ($+15.7\%$) and from 3.10 to 3.62 in NDCG@10 ($+16.6\%$), the largest relative gain anywhere in the table.
BSARec improves from 9.59 to 10.09 in HR@10 ($+5.3\%$) and from 6.06 to 6.30 in NDCG@10 ($+3.9\%$).
DIFF improves from 6.02 to 6.58 in HR@10 ($+9.3\%$) and from 3.60 to 3.96 in NDCG@10 ($+10.0\%$).
All four Toys comparisons show positive gains, including BSARec ($+0.00505$ in HR@10).
\end{revblock}

\paragraph{Games.}
\begin{revblock}
Games shows broadly positive but more moderate gains.
SASRec improves from 12.90 to 13.57 in HR@10 ($+5.3\%$) and from 7.15 to 7.53 in NDCG@10 ($+5.3\%$).
BERT4Rec improves from 9.73 to 10.53 in HR@10 ($+8.2\%$) and from 5.18 to 5.60 in NDCG@10 ($+8.0\%$).
BSARec improves from 13.74 to 14.13 in HR@10 ($+2.8\%$) and from 7.62 to 7.89 in NDCG@10 ($+3.4\%$), the smallest relative HR@10 gain in this domain.
DIFF improves from 11.82 to 12.28 in HR@10 ($+3.9\%$) and from 6.55 to 6.82 in NDCG@10 ($+4.2\%$).
Although many game reviews contain experiential language, only a subset is captured by our strict five-facet schema; Section~\ref{sec:facet_ablation} shows that in Games the visual and auditory facets contribute most.
\end{revblock}

\paragraph{Cross-backbone analysis.}
\begin{revblock}
All four backbones are ID-only in the base row, so absolute differences across backbones are architecture and capacity comparisons under one shared pipeline, not comparisons between an ID-only and a side-information model.
BSARec is the strongest base backbone in all five domains on both HR@10 and NDCG@10.
The controlled ASER comparison remains the within-cell difference between the frozen backbone's own ranking and the ranking produced after the bounded relational residual is applied.
\end{revblock}

\paragraph{Hit rate versus ranking quality.}
\begin{revblock}
The relative NDCG@10 gain exceeds the relative HR@10 gain in 11 of the 20 cells, so the residual is not merely inserting additional correct items into the top-$K$ set; it also reorders the candidates the backbone already retrieved.
The clearest instances are Grocery--BERT4Rec, where HR@10 improves by $+10.4\%$ while NDCG@10 improves by $+15.1\%$, and Grocery--DIFF, where HR@10 improves by $+5.0\%$ and NDCG@10 by $+8.1\%$.
This split is the one that the matched-channel control of Section~\ref{sec:channel_control} isolates: the hit-rate component is reproduced by a non-sensory channel built with the same pipeline, whereas the ranking-quality component is not.
\end{revblock}

\paragraph{Summary.}
\begin{revblock}
The main results establish two points.
Structured, evidence-grounded item-side records, integrated as a bounded residual over a frozen backbone, improve every domain--backbone cell we test on every reported metric, with no cell requiring an exception.
What the table does not establish on its own is that the improvement is sensory in origin; Section~\ref{sec:channel_control} separates that question by metric, and Sections~\ref{sec:recall_audit} and~\ref{sec:exposure} bound the completeness of the extraction and the exposure cost of reranking.
\end{revblock}

\subsection{Teacher Model Semantic Alignment}
\label{sec:teacher_alignment}

We evaluate how closely the Qwen teacher reproduces the sensory extraction behavior of GPT-5 Mini on a held-out validation split.
Table~\ref{tab:teacher_alignment} reports micro-averaged precision, recall, and F1 for complementary alignment metrics.
Exact string comparison can underestimate agreement because sensory values are open vocabulary and often differ by paraphrase.
We therefore report metrics that separately assess exact attribute--value agreement, fine-grained attribute agreement, semantic value agreement, and coarse five-facet agreement.

\paragraph{Matching unit and normalization.}
We evaluate sets of extracted attribute--value records.
Each record is represented as an attribute string and a value string.
Before comparison, we lowercase text, collapse repeated whitespace, and remove trivial punctuation.
We compute micro-averaged precision, recall, and F1 by counting true positives, false positives, and false negatives across all records in the validation split.

\paragraph{Exact match.}
Exact match counts a record as correct only when both the attribute and the value match a reference record after normalization.

\paragraph{Attribute match.}
Attribute match evaluates whether the teacher identifies the same fine-grained sensory attribute as the reference, regardless of the value text.
A prediction is counted as correct if its attribute matches any reference attribute for the same item.

\paragraph{Semantic value match.}
We measure semantic similarity between value texts using sentence embeddings and cosine similarity.
For each predicted record, we restrict candidates to reference records with the same attribute and compute cosine similarity between the predicted value and each reference value.
We count a match when the maximum cosine similarity exceeds a threshold.
Table~\ref{tab:teacher_alignment} reports semantic value match at thresholds of 0.8, 0.7, and 0.6.

\paragraph{Facet-class match.}
We also report a coarse five-facet match that maps each extracted attribute--value record to one of the five sensory facets: visual, tactile, auditory, olfactory, and gustatory.
A prediction is counted as matched when its mapped facet class matches that of the corresponding reference.

\begin{table}[H]
\centering\small
\setlength{\tabcolsep}{4pt}
\renewcommand{\arraystretch}{1.05}
\begin{tabular}{lccc}
\toprule
Metric & Precision & Recall & F1 \\
\midrule
Exact match & 0.4165 & 0.4398 & 0.4278 \\
Attribute match & 0.7557 & 0.7609 & 0.7583 \\
Semantic value match, 0.8 & 0.5157 & 0.5444 & 0.5297 \\
Semantic value match, 0.7 & 0.5728 & 0.6047 & 0.5883 \\
Semantic value match, 0.6 & 0.6120 & 0.6461 & 0.6286 \\
Facet-class match & 0.8523 & 0.8379 & 0.8450 \\
\bottomrule
\end{tabular}
\caption{Teacher alignment between Qwen and GPT-5 Mini on the validation split. Exact match requires exact attribute and value equality after normalization. Attribute match evaluates agreement on the fine-grained extracted attribute only. Semantic value match evaluates value agreement using cosine similarity thresholds of 0.8, 0.7, and 0.6. Facet-class match evaluates agreement after mapping each extracted record to one of the five coarse sensory facets.}
\label{tab:teacher_alignment}
\end{table}

As shown in Table~\ref{tab:teacher_alignment}, attribute match is substantially higher than exact match, indicating that the Qwen teacher usually identifies the correct fine-grained sensory attribute while differing in the surface form of the value.
Semantic value match increases as the threshold is relaxed, which is consistent with disagreements driven by paraphrasing and synonym choice rather than incorrect sensory grounding.
Facet-class match achieves the highest F1 score, suggesting that extracted values are typically grounded in the correct coarse sensory class even when the fine-grained value wording differs.
These results support using the Qwen teacher for offline annotation and motivate distillation into a compact student encoder that avoids LLM inference during recommendation.

\subsection{Extraction Recall}
\label{sec:recall_audit}

\begin{revblock}
The diagnostics above are precision-oriented: the human seed-label audit in Table~\ref{tab:seed_audit} (pair precision 0.88 and evidence faithfulness 0.90) and an evidence-grounding rate of 0.948 over the 60{,}493 records that the seed model extracted for the full Beauty evaluation catalog, in which 11{,}806 of the 12{,}101 items yield at least one record.
Neither measures how much of the sensory content present in the reviews the pipeline fails to extract.
We therefore conduct a recall audit.

The audit uses an independent local LLM auditor (Qwen3-30B-A3B-Instruct-2507) under a two-stage protocol designed to avoid circularity.
In the first stage the auditor enumerates sensory mentions in open vocabulary, blind to our thirteen-attribute schema and to all pipeline output, with the target defined neutrally as any span reporting a product property perceivable by sight, touch, smell, taste, or hearing.
In a second, separate call the enumerated mentions are mapped onto our taxonomy, so that our schema never defines what counts as a mention.
Bare-valence spans, which our vocabulary excludes, are recorded rather than dropped during enumeration, so that denominators with and without them come from the same annotation.
The audit covers 750 items in total, across the capture panel and the auditor-reliability measurements of hallucination, self-consistency, and saturation, and required 4{,}401 auditor generations over its enumeration, mapping, and verification passes.
The capture panel reported here draws on a sample of 310 items from the seed annotation corpus; 293 yielded at least one enumerated mention. The panel is computed over these 293 items, which contain a total of 3{,}070 mentions.

\par\smallskip
{\centering
\captionof{table}{LLM-audited recall of the extraction pipeline. The panel is computed over the 293 scored items of a 310-item sample drawn from the seed annotation corpus, covering 3{,}070 enumerated mentions. Intervals are 95\% bootstrap confidence intervals.}
\label{tab:recall_audit}
\footnotesize
\setlength{\tabcolsep}{7pt}
\renewcommand{\arraystretch}{1.12}
\begin{tabular}{lcc}
\toprule
Measurement & Recall & 95\% CI \\
\midrule
Strict (attribute and evidence span match)     & 0.499 & [0.472, 0.526] \\
Lenient (many-to-one match permitted)          & 0.511 & [0.485, 0.539] \\
Facet level                                    & 0.541 & [0.517, 0.567] \\
Item level (attribute captured somewhere in the item) & 0.907 & --- \\
\midrule
After record cleaning                          & 0.496 & [0.470, 0.524] \\
After 512-token truncation                     & 0.457 & [0.429, 0.485] \\
Facet-presence labels                          & 0.967 & [0.953, 0.979] \\
Dense supervision targets                      & 0.333 & [0.311, 0.354] \\
\bottomrule
\end{tabular}
\par}
\smallskip

\noindent
The informative quantity is the gap between 0.499 and 0.907.
Half of all individual mentions are not matched under a strict span-and-attribute criterion, but 90.7\% of attributes are captured somewhere in the same item, so the loss is overwhelmingly instance loss rather than content loss: the pipeline has usually already recorded the same property from a different span.
Only 1.2\% of mentions fall in a facet that the pipeline missed entirely for that item.
Ranking the sources of attrition, the largest is the twelve-record per-item cap of Section~\mbox{\ref{sec:method_supervision}}, where the miss rate is 0.528 at the cap against 0.386 below it; the three-pair cap on dense supervision targets costs a further 12.5 percentage points; 512-token truncation costs 3.9 points; and record cleaning costs 0.3 points, or eight records, which is negligible.

We also examined whether the facet bank reproduces the audited mentions and report that measurement with an explicit caveat because the aggregate figure is not a recall measure.
Visual and tactile facets are predicted as present for essentially the entire catalog; a bank that always predicts facet presence would score 1.0, and a shuffled placebo already reaches 0.846. Thus, the aggregate rate mostly measures positive-prediction frequency.
The informative form is the confusion table (923 true positives, 48 false negatives, 244 false positives, 985 true negatives; precision 0.791, specificity 0.802) and the restriction to the three discriminative facets, where the match rate is 0.830 over 249 mentions, decomposing into olfactory 0.961, gustatory 0.584, and auditory 0.128.
The auditory value has an identified cause rather than being an artifact of sparsity: auditory content is extracted but misrouted to visual or tactile attributes.
For one Beauty item, the pipeline extracted the span \emph{``fizzes and releases rose petals''} but labeled it as a visual attribute; auditory coverage in the Beauty bank is only 242 of 12{,}101 items.
This mechanism explains why the auditory facet contributes little in Beauty but is among the most influential facets in Games in Section~\ref{sec:facet_ablation}.

These recall estimates have two important limitations.
First, no human anchor was collected, so these are LLM-audited recall estimates and the auditor's own bias cannot be corrected; the auditor's measured hallucination rate is 3.37\% and its self-consistency is 0.93.
Second, auditor saturation was not reached: a third enumeration pass still added 19.0\% new mentions, exceeding our design criterion of less than 10\%. The denominator is therefore incomplete, and every recall figure above likely overestimates true recall.
The denominator is also restricted to content our taxonomy can represent.
On the seed annotation corpus, which is drawn from the Amazon Reviews 2018 supervision source rather than from the evaluation domains, the unconstrained seed model proposed attributes outside our inventory for 9.2\% of all extracted attributes (2{,}602 of 28{,}146), principally weight (1{,}422 instances), comfort (1{,}051), and temperature (129); this content is not counted as missed here.
\end{revblock}

\subsection{Sensitivity to Extraction Quality}
\label{sec:extraction_sensitivity}

\begin{revblock}
Because the item-side signal originates from an LLM, it is necessary to know how much of the downstream gain depends on the quality of that extraction and on the thresholds that filter it.
We measure 14 operating points along three axes on Beauty, using the BSARec backbone and the relational integration described in Section~\ref{sec:method_integration} with a fixed seed. The ID-only baseline achieves an HR@10 of 0.09310 and an NDCG@10 of 0.05697.
The first axis varies the bank-side presence gate while holding the student fixed; the second and third axes require retraining the student, so their reference point differs slightly from that of the first axis (+0.00447 with the fixed student and +0.00461 after retraining at the same nominal setting).

\par\smallskip
{\centering
\captionof{table}{Sensitivity of the downstream gain to the presence threshold on the bank-side presence gate (student fixed). Bank slots are counted over the Beauty catalog.}
\label{tab:sens_threshold}
\footnotesize
\setlength{\tabcolsep}{7pt}
\renewcommand{\arraystretch}{1.12}
\begin{tabular}{lcccc}
\toprule
Threshold & Bank slots & HR@10 & $\Delta$HR@10 & $\Delta$NDCG@10 \\
\midrule
0.3 & 35{,}806 & 0.09717 & +0.00407 & +0.00199 \\
0.4 & 34{,}095 & 0.09699 & +0.00389 & +0.00204 \\
0.5 (selected) & 32{,}843 & 0.09757 & +0.00447 & +0.00228 \\
0.6 & 31{,}974 & 0.09806 & \textbf{+0.00496} & \textbf{+0.00243} \\
0.7 & 31{,}303 & 0.09739 & +0.00429 & +0.00220 \\
0.8 & 30{,}540 & 0.09717 & +0.00407 & +0.00234 \\
\bottomrule
\end{tabular}
\par}
\smallskip

\par\smallskip
{\centering
\captionof{table}{Sensitivity to the confidence margin applied at extraction time (student retrained at every point). Teacher-record retention falls from 90.1\% at margin $0.00$ to 57.5\% at $+0.20$.}
\label{tab:sens_margin}
\footnotesize
\setlength{\tabcolsep}{7pt}
\renewcommand{\arraystretch}{1.12}
\begin{tabular}{lcccc}
\toprule
Margin & Bank slots & HR@10 & $\Delta$HR@10 & $\Delta$NDCG@10 \\
\midrule
$-0.20$ & 32{,}369 & 0.09806 & +0.00496 & +0.00214 \\
$-0.10$ & 32{,}183 & 0.09708 & +0.00398 & +0.00171 \\
$0.00$ (selected) & 32{,}854 & 0.09771 & +0.00461 & +0.00194 \\
$+0.10$ & 31{,}864 & 0.09744 & +0.00434 & +0.00180 \\
$+0.20$ & 30{,}074 & 0.09641 & +0.00331 & +0.00160 \\
\bottomrule
\end{tabular}
\par}
\smallskip

\par\smallskip
{\centering
\captionof{table}{Sensitivity to injected label noise, in which a fixed fraction of teacher records is routed to a randomly chosen wrong facet (student retrained at every point).}
\label{tab:sens_noise}
\footnotesize
\setlength{\tabcolsep}{7pt}
\renewcommand{\arraystretch}{1.12}
\begin{tabular}{lcccc}
\toprule
Injected flips & Facet accuracy & Bank slots & $\Delta$HR@10 & $\Delta$NDCG@10 \\
\midrule
0\% (control) & 1.000 & 32{,}854 & +0.00461 & +0.00194 \\
10\% & 0.899 & 31{,}437 & +0.00465 & +0.00215 \\
25\% & 0.749 & 31{,}504 & +0.00407 & +0.00185 \\
50\% & 0.497 & 40{,}772 & +0.00478 & +0.00230 \\
\bottomrule
\end{tabular}
\par}
\smallskip

\noindent
Along the bank-side presence gate, the HR@10 gain remains between +0.00389 and +0.00496 while bank coverage changes by 15\%; the gain is positive at every point.
The selected threshold is not the highest-performing value in the tested range: 0.6 yields a marginally larger gain (+0.00496 versus +0.00447), so the reported number lies on a plateau rather than at a tuned peak, slightly below the best observed point.

Along the extraction-side confidence gate the response is flat on the permissive side and degrades monotonically on the restrictive side, losing 28\% of the HR@10 gain and 18\% of the NDCG@10 gain at margin $+0.20$, where teacher-record retention has fallen to 57.5\%.
The plateau is one-sided, and the operating point lies at its right-hand edge.
Because teacher confidence scores are quantized to eight discrete values, the gate behaves as a step function; therefore, a teacher with a different confidence calibration could enter the degrading regime without any change to the nominal threshold.
The risk is over-filtering rather than under-filtering, and a practitioner replacing the teacher should verify the retention rate for the new teacher rather than reuse the threshold value.

Under injected label noise the gain survives even when half of all teacher records are routed to the wrong facet (+0.00478 HR@10, marginally above the noiseless control), and degradation is not monotone.
This result does not establish robustness to extraction errors in general: cross-facet flips relocate evidence rather than delete it, so the total quantity of sensory evidence attached to an item is preserved and only its facet assignment is corrupted; at 50\% the slot count rises to 40{,}772, which is the injection of false positives rather than a recovery of signal.
The results instead suggest that the gain depends on the presence of sensory evidence for each item rather than on its fine-grained facet assignment. This interpretation is consistent with the misrouting behavior documented above, in which the downstream gain persists despite facet misrouting.
\end{revblock}

\subsection{Temporal-Control Evaluation}
\label{sec:temporal_control}

\begin{revblock}
The student supervision corpus is drawn from Amazon Reviews 2018 while recommendation is evaluated on Amazon Reviews 2014, so review text written after the evaluation period could in principle contribute to the item representations.
The item-disjointness rule of Section~\ref{sec:method} removes direct textual overlap between supervision and evaluation items, but it does not by itself establish temporal precedence of the supervision text.
As an empirical control, we rebuilt the sensory channel under a conservative temporal restriction and re-evaluated three representative domains.

Both corpora carry per-review timestamps, which makes the control feasible.
The last interaction in the 2014 benchmark is dated 2014-07-23, and we retain a supervision item only if it has at least one review dated on or before that cutoff.
The filter removes 48.9\% of the supervision training split (5{,}278 of its 10{,}337 items remain).
All other pipeline choices remain unchanged: the same schema, teacher labels, cleaning procedure, student architecture and hyperparameters, split construction, item-disjointness rule, and bounded relational integration are used, and the frozen ID-only BSARec checkpoints from the main table are reused without retraining.
Bank coverage shrinks accordingly: supported item-facet slots drop by 6.8\% in Beauty (32{,}843 to 30{,}618), 9.2\% in Grocery (28{,}321 to 25{,}728), and 2.3\% in Video Games (28{,}243 to 27{,}580).
Table~\mbox{\ref{tab:temporal_control}} reports the result.
\end{revblock}

\begin{table}[t]
\centering
\caption[Temporal-control evaluation]{\rev{Temporal-control evaluation with BSARec. Temporal ASER rebuilds the sensory channel from supervision items that have at least one review dated on or before the 2014-07-23 evaluation cutoff (51.1\% of the supervision training split); the backbone checkpoints are identical to the main table. The improvement over the ID-only base persists in all three domains.}}
\label{tab:temporal_control}
\begin{revtable}
\begin{tabular}{lcccccc}
\hline
& \multicolumn{3}{c}{\textbf{HR@10}} & \multicolumn{3}{c}{\textbf{NDCG@10}} \\
\textbf{Domain} & Base & ASER & Temporal & Base & ASER & Temporal \\
\hline
Beauty      & 0.0931 & 0.0976 & 0.0971 & 0.0570 & 0.0593 & 0.0585 \\
Grocery     & 0.0975 & 0.0999 & 0.1028 & 0.0565 & 0.0579 & 0.0588 \\
Video Games & 0.1374 & 0.1413 & 0.1412 & 0.0762 & 0.0789 & 0.0789 \\
\hline
\end{tabular}
\end{revtable}
\end{table}

\begin{revblock}
The improvement over the ID-only base persists in all three domains under the temporal restriction, and in Grocery the temporally restricted channel is slightly stronger than the unrestricted one.
The persistence of the gain under this restriction indicates that the observed improvement does not depend entirely on unrestricted post-cutoff review text.
The filter operates at the item level: an item is retained if any of its reviews predates the cutoff, and the annotated text for a retained item may still include later reviews.
A strictly review-level filter retains only 3.9\% of the supervision corpus and would therefore confound temporal control with severe data scarcity.
The experiment is a sensitivity control, not a proof of the absence of temporal leakage.
\end{revblock}

\subsection{Student Validation}
\label{sec:student_validation}

\begin{revblock}
Table~\mbox{\ref{tab:student_validation}} reports the diagnostic performance of the finalized facet-aware student used to construct the released facet banks, evaluated on its held-out supervision split (545 examples) with the same presence threshold of 0.5 that is used at bank construction.
In addition to the aggregate score, the table gives the per-facet decomposition: positive support, precision, recall, and F1 for each of the five sensory facets.
The same evaluation pass yields a macro precision of 0.8461, macro recall of 0.7777, macro F1 of 0.7967, micro F1 of 0.8909, and macro AUROC of 0.9216; the cosine similarity between the active-facet embedding and the teacher target is 0.6696, and the teacher-confidence Brier score is 0.0480.
All values are computed using the released evaluation code and student checkpoint.
\end{revblock}

\begin{table}[t]
\centering
\caption[Per-facet presence scores of the facet-aware student]{\rev{Per-facet presence scores of the released facet-aware student on the supervision validation split (545 examples, presence threshold 0.5). Support is the number of positive examples per facet. Macro F1 is 0.7967 and micro F1 is 0.8909.}}
\label{tab:student_validation}
\begin{revtable}
\begin{tabular}{lrrrr}
\hline
\textbf{Facet} & \textbf{Support} & \textbf{Precision} & \textbf{Recall} & \textbf{F1} \\
\hline
visual    & 447 & 0.9032 & 0.9597 & 0.9306 \\
tactile   & 413 & 0.8692 & 0.9492 & 0.9074 \\
auditory  &  64 & 0.7647 & 0.4062 & 0.5306 \\
olfactory &  61 & 0.8222 & 0.6066 & 0.6981 \\
gustatory &  91 & 0.8713 & 0.9670 & 0.9167 \\
\hline
\end{tabular}
\end{revtable}
\end{table}

\begin{revblock}
The decomposition shows that the facets are not equally reliable and that sparsity alone does not determine reliability.
Two sparsely represented facets behave in opposite ways.
Gustatory has relatively low support (91) yet the second-highest F1 (0.9167), which is consistent with the relatively distinctive vocabulary of taste descriptions.
Auditory is comparably sparse (64) and is clearly the weakest facet (F1 0.5306), with precision (0.7647) far exceeding recall (0.4062): the student produces few false-positive auditory predictions but misses most true auditory instances.
The auditory deficit is consistent with two independent observations reported elsewhere in this paper: the extraction audit of Section~\ref{sec:recall_audit} documents auditory content that is extracted but routed to visual or tactile attributes, and the per-facet ablation of Section~\ref{sec:facet_ablation} finds the auditory channel to have a negligible effect in Beauty.
That same ablation, however, finds auditory the second-most important facet in Video Games, so a weak presence classifier for a facet does not imply that the facet channel is useless downstream; its utility is domain-dependent.
Olfactory falls between these extremes (F1 0.6981) and is likewise limited mainly by recall.
\end{revblock}

\subsection{Exposure Diversity and Long-Tail Coverage}
\label{sec:exposure}

\begin{revblock}
Accuracy metrics do not reveal how a reranking method redistributes exposure across the catalog.
We therefore measure exposure-side metrics on all 20 domain--backbone cells, reproducing the reported accuracy to the stated precision in every cell.
An item's exposure is the number of times it appears in the top-$K$ recommendation lists over all test users.
Catalog coverage is the fraction of the catalog that appears in at least one list; the Gini coefficient is computed over the per-item exposure counts; the long-tail share is the fraction of recommendation slots held by items outside the Pareto head, defined as the smallest set of most popular items covering 80\% of training interactions; and novelty is the mean self-information $-\log_2 p(i)$ of recommended items, in bits, where $p(i)$ is the item's smoothed share of training interactions.

\par\smallskip
{\centering
\captionof{table}{Exposure metrics at $K=10$. The right-hand column reports how many of the 20 domain--backbone cells move in the stated direction. The upper block gives the Beauty--BSARec cell as a representative example; the lower block gives a quantity that is meaningful only as a grid average because, in the Beauty--BSARec cell alone, the top-1\% exposure share rises slightly, from 0.2019 to 0.2048.}
\label{tab:exposure}
\footnotesize
\setlength{\tabcolsep}{6pt}
\renewcommand{\arraystretch}{1.12}
\begin{tabular}{lccc}
\toprule
Metric & Base & Ours & Cells \\
\midrule
\multicolumn{4}{l}{\textit{Beauty--BSARec cell}} \\
Catalog coverage ($\uparrow$ better) & 0.8334 & 0.7788 & 16/20 fall \\
Gini of exposure ($\downarrow$ better) & 0.7789 & 0.8122 & 16/20 rise \\
Long-tail (Pareto) share ($\uparrow$ better) & 0.1420 & 0.1120 & 19/20 fall \\
Novelty ($\uparrow$ better) & 12.223 & 12.142 & --- \\
\midrule
\multicolumn{4}{l}{\textit{Mean over the 20 cells}} \\
Exposure share of top-1\% popular items ($\downarrow$ better) & 0.2601 & 0.2501 & 14/20 fall \\
\bottomrule
\end{tabular}
\par}
\smallskip

\noindent
The method concentrates exposure.
Catalog coverage falls, exposure inequality rises, and the long-tail share falls in 19 of 20 cells at cutoffs up to $K=20$ and in 18 of 20 cells at $K=50$.
The direction is identical at every cutoff: catalog coverage falls in 16 of the 20 cells at $K=1$, $5$, $10$, $20$, and $50$, with mean changes of $-0.018$, $-0.031$, $-0.030$, $-0.023$, and $-0.007$, respectively. Thus, the result is not an artifact of the reporting cutoff.
The novelty change shrinks as $K$ grows and is $-0.002$ on average at $K=50$.
This exposure concentration constitutes a limitation of the method.

Two observations qualify the finding without removing it.
First, the exposure that is redistributed moves toward the middle of the popularity distribution rather than toward the head.
In the Beauty--BSARec cell, the slot share changes by $-2.26$ points for items with at most five interactions, $-1.32$ points for 6--10, $+0.40$ points for 11--20, $+2.58$ points for 21--50, and $+0.60$ points for 51 or more; averaged over the 20 cells, the exposure share held by the most popular 1\% of items falls from 0.2601 to 0.2501, with decreases in 14 of the 20 cells.
Second, the recommendation lists do not stagnate.
Although the trust region restricts reranking to the backbone's own top 50 or top 100 candidates, an average of 30.0\% of the slots in the final top 10 contain items that the ID backbone did not place in its own top 10. This proportion ranges from 17.2\% to 43.7\% across cells and reaches 33.5\% in the Beauty--BSARec cell, indicating that the concentrated exposure is redistributed rather than merely reinforced.
\end{revblock}

\subsection{Is the Gain Sensory-Specific?}
\label{sec:channel_control}

\begin{revblock}
A structured item-side channel could improve recommendation for two different reasons: because its content is sensory, or because any evidence-grounded extraction imposes useful structure on review text.
To separate these explanations, we build a sentiment channel that is matched to the sensory channel in every respect except vocabulary and compare the two directly.
The sentiment channel uses 50 descriptors organized into five aspects (quality, value, efficacy, usability, and satisfaction) that occupy the same five channel slots as the sensory facets.
Lexical overlap with the sensory vocabulary is zero by construction: bare valence and undifferentiated quality statements are exactly what the sensory vocabulary excludes, so the sentiment channel is built from the material the sensory channel discards.
Both channels share the same seed model, the same six-field record schema, the same catalog, the same supervision code path, the same student hyperparameters, the same bank-construction procedure, the same split construction, and the same relational integration described in Section~\ref{sec:method_integration}.
The sentiment bank is denser: on Beauty, it activates 44{,}274 slots compared with 30{,}520 for the Beauty-only sensory bank of Panel~A, because reviewers state opinions more often than they describe percepts; the five-domain sensory bank activates 32{,}843 slots at the same threshold (Table~\mbox{\ref{tab:sens_threshold}}).

We measure the comparison along both axes: backbones at fixed domain, and domains at fixed backbone.
On the backbone axis, all four backbones are evaluated on Beauty, and the sensory student is retrained on Beauty alone so that it does not carry the five-domain training advantage.
The sentiment student for this axis is built from the same Beauty items and the same train/validation partition: of the 2{,}069 training and 109 validation items of the sensory student, seed annotation with the sentiment vocabulary succeeded and produced at least one in-vocabulary record for 2{,}008 and 105 respectively, and those items form the sentiment supervision set.
The item text is byte-identical between the two channels, so within this axis the only difference is the extraction vocabulary.
On the domain axis, BSARec is evaluated on all five domains, with a sentiment student trained on the same items and the same split as the five-domain sensory student, which annotated 10{,}848 of those 10{,}882 items successfully; 705 of the annotated items yielded no record inside the sentiment vocabulary and are dropped, leaving 9{,}633 training and 510 validation items. This panel therefore holds the supervision items themselves fixed.
Because the two axes use different sensory students by design, the Beauty--BSARec entry differs between panels.

\par\smallskip
{\centering
\captionof{table}{Sensory versus matched sentiment channel. Values are changes against each backbone's own ID-only baseline under identical integration. Panel A varies the backbone on Beauty with a Beauty-only sensory student; Panel B varies the domain with BSARec and the five-domain students. Both panels reuse the frozen ID-only backbone checkpoints of Table~\mbox{\ref{tab:performance}}, so each entry is the change produced by adding a channel to the same base ranking; only changes are reported because the object of comparison is the difference between the two channels, not the absolute level. The better channel in each comparison is in bold.}
\label{tab:channel_control}
\footnotesize
\setlength{\tabcolsep}{5pt}
\renewcommand{\arraystretch}{1.12}
\begin{tabular}{lcccc}
\toprule
& \multicolumn{2}{c}{$\Delta$HR@10} & \multicolumn{2}{c}{$\Delta$NDCG@10} \\
\cmidrule(lr){2-3}\cmidrule(lr){4-5}
& Sensory & Sentiment & Sensory & Sentiment \\
\midrule
\multicolumn{5}{l}{\textit{Panel A: backbone axis (Beauty)}} \\
SASRec   & \textbf{+0.00541} & +0.00470          & \textbf{+0.00246} & +0.00190 \\
BERT4Rec & +0.00434          & \textbf{+0.00554} & +0.00284          & \textbf{+0.00299} \\
BSARec   & +0.00429          & \textbf{+0.00532} & \textbf{+0.00311} & +0.00244 \\
DIFF     & \textbf{+0.00358} & +0.00322          & \textbf{+0.00181} & +0.00166 \\
\midrule
\multicolumn{5}{l}{\textit{Panel B: domain axis (BSARec)}} \\
Beauty  & +0.00447          & +0.00447          & \textbf{+0.00228} & +0.00201 \\
Toys    & \textbf{+0.00505} & +0.00448          & \textbf{+0.00236} & +0.00201 \\
Sports  & +0.00230          & \textbf{+0.00244} & \textbf{+0.00121} & +0.00098 \\
Games   & \textbf{+0.00391} & +0.00214          & \textbf{+0.00261} & +0.00164 \\
Grocery & \textbf{+0.00238} & +0.00143          & \textbf{+0.00139} & +0.00133 \\
\bottomrule
\end{tabular}
\par}
\smallskip

\noindent

For HR@10, the two channels are indistinguishable.
On the backbone axis, each channel obtains the larger gain in two comparisons; on the domain axis, the sensory channel obtains the larger gain in three comparisons, the sentiment channel in one, and one comparison is an exact tie.
Three of the nine differences exceed the run-to-run variability of $\pm 0.001$ that we measure for this pipeline (Table~\mbox{\ref{tab:seed_stability}}; Section~\mbox{\ref{sec:stage4_ablation}}): BERT4Rec ($-0.00121$) and BSARec ($-0.00103$) favor the sentiment channel and Games ($+0.00177$) favors the sensory channel, so the exceedances do not share a direction.
Both channels improve every backbone and every domain over the ID-only baseline, by +0.00143 to +0.00554.
These results indicate that entry into the top-$K$ set follows from structured, evidence-grounded extraction as such, rather than from the sensory content of the extracted attributes.

On NDCG@10 the sensory channel yields a larger gain in eight of the nine comparisons, three on the backbone axis and five on the domain axis.
The margins in those eight range from +0.00006 on Grocery to +0.00097 on Games and are often within the observed run-to-run variability; the evidence therefore lies in the consistency of direction rather than the magnitude of any individual comparison.
The exception is BERT4Rec on Beauty, where the sentiment channel is higher by 0.00015, an order of magnitude below that variability. Under a two-sided sign test over the nine comparisons, eight agreeing directions give $p=0.039$; because the nine cells share one pipeline and one bank family they are not fully independent, so we read this as a summary of consistency rather than as an independent significance test.
Sensory descriptors such as \emph{rose} or \emph{matte} provide shared experiential properties that are absent from generic evaluative descriptors such as \emph{well-made} or \emph{overpriced}.
The largest NDCG@10 difference on the domain axis occurs in Games (+0.00097), where the visual and auditory facets are also among the most influential in the leave-one-facet-out analysis of Section~\ref{sec:facet_ablation}.

Taken together, these results indicate that structured, evidence-grounded attribute extraction drives the hit-rate gains regardless of the channel, whereas the sensory channel shows a consistent advantage in ranking quality across eight of the nine backbone--domain comparisons. Because individual margins are often within the observed run-to-run variability, this result is interpreted from the consistency of direction rather than the magnitude of any single comparison.
Two limitations apply.
Each cell is a single seed, so the hit-rate comparison cannot resolve differences at the $\pm 0.001$ scale.
The matched sentiment channel is itself competitive under this protocol; whether the two channels are complementary when combined is left to future work.
\end{revblock}

\subsection{Comparison with Text-Enhanced Baselines}
\label{sec:text_baselines}

\begin{revblock}
We train and evaluate the complete FineRec model~\citep{zhang2024finerec} as a recommender in its own right, within our evaluation pipeline, together with a matched raw-text control and the ID-only rows for every backbone.

We ported Eq.~(1)--(10) of FineRec from the paper and the official release.
An independent re-derivation of the equations agrees with our implementation to a maximum absolute deviation of $3.0\times10^{-8}$ on the user path and $1.2\times10^{-7}$ on the item path, and the Eq.~(9) window and the Eq.~(10) scoring rule match.
Our implementation follows the official release at three points where it differs from the paper: the loss is the release's full-softmax cross entropy rather than the BCE of Eq.~(11), the merge stage uses the release's bottleneck MLP rather than scoring directly on the concatenation of Eq.~(10), and attribute embeddings are learned rather than BERT-initialized.
One deviation is induced by our data: the attribute inventory is our thirteen sensory attributes rather than site metadata such as brand or price.
This is deliberate, so that both systems observe the same LLM-extracted attribute--opinion signal and the comparison isolates the architecture rather than the text source.

FineRec's text path is a learned $\mathrm{Linear}(768 \rightarrow d)$ projection, with PCA used only for initialization; in the selected configuration ($d=16$), that path contains 12{,}304 trainable parameters, and the complete FineRec model contains 6{,}223{,}736.
The formulation used in the official release for the same path is a free per-opinion embedding table over the 276{,}601 opinion phrases of the graph, which would hold 4{,}425{,}617 parameters on its own; we report the projection variant because it lets gradients reach all 768 input dimensions instead of the 39.1\% of variance a fixed 16-component PCA retains.
The default attention is the softmax-cosine form, which is also the form selected on validation, and attributes are restricted to those whose coverage is at least 5\% of both the user side and the item side, retaining eleven of thirteen.
Hyperparameters were selected on validation NDCG@10 over a twelve-configuration grid (attribute dimension $\{16,32,64\}$ $\times$ learning rate $\{10^{-3}, 5\times10^{-4}\}$ $\times$ two attention forms, softmax attention over cosine similarity and ReLU-ratio attention over cosine similarity, with at most 40 epochs and patience 5).
The selected configuration is $d=16$ with learning rate $10^{-3}$ and softmax-cosine attention, reaching validation NDCG@10 of 0.042737 and HR@10 of 0.078031 at epoch 17.
Validation NDCG@10 across the whole grid spans 0.0402--0.0427, a relative spread of 6\%, so the hyperparameter choice does not determine the outcome.
The best epoch across the 12 configurations ranges from 5 to 38; for 11 of them, the patience window closed before the 40-epoch cap, so training ended through early stopping rather than upon reaching the budget. Only the configuration whose best epoch is 38 could have been curtailed by the cap.
That configuration is not the selected one.

\par\smallskip
{\centering
\captionof{table}{Text-enhanced baseline comparison on the Beauty test split under a single leave-one-out protocol with full-catalog ranking and no negative sampling. FineRec receives the same LLM-extracted attribute--opinion signal as our method. Base rows are ID-only.}
\label{tab:text_baselines}
\footnotesize
\setlength{\tabcolsep}{7pt}
\renewcommand{\arraystretch}{1.12}
\begin{tabular}{llcc}
\toprule
Model & Item-side signal & HR@10 & NDCG@10 \\
\midrule
FineRec                 & LLM attribute--opinion graph & 0.05022 & 0.02640 \\
\midrule
SASRec (base)           & ID only & 0.08760 & 0.05415 \\
BERT4Rec (base)         & ID only & 0.06283 & 0.03394 \\
DIFF (base)             & ID only & 0.09113 & 0.05527 \\
BSARec (base)           & ID only & 0.09310 & 0.05697 \\
\midrule
BSARec + relational (ours) & Frozen facet bank & \textbf{0.09757} & \textbf{0.05926} \\
\bottomrule
\end{tabular}
\par}
\smallskip

\noindent
We reimplemented FineRec within our evaluation pipeline so that it is trained and evaluated under exactly the same leave-one-out split and full-catalog ranking protocol as every other model reported here, and we provided it with the same LLM-extracted attribute--opinion signal used by our method.
Its hyperparameters were selected on validation NDCG@10 over a twelve-configuration grid spanning attribute dimension, learning rate, and attention type.
Because the published FineRec results use a different preprocessing pipeline and sampled-negative evaluation, the relevant comparison is strictly within our shared evaluation pipeline: under an identical split, an identical ranking protocol, and an identical text signal, relational integration of a frozen facet bank is more effective than per-attribute graph propagation.

The second control isolates representation rather than architecture.
We encode the same source text with a general-purpose sentence encoder, expose it as a single-facet bank, and apply the same relational integration used for the sensory bank, so that the only difference is whether the item-side signal is structured into facets.

\par\smallskip
{\centering
\captionof{table}{Raw-text embedding control on Beauty. Both channels use the same source text and the same relational integration; only the representation differs. Changes are against each backbone's ID-only baseline.}
\label{tab:rawtext_control}
\footnotesize
\setlength{\tabcolsep}{6pt}
\renewcommand{\arraystretch}{1.12}
\begin{tabular}{lcccc}
\toprule
& \multicolumn{2}{c}{$\Delta$HR@10} & \multicolumn{2}{c}{$\Delta$NDCG@10} \\
\cmidrule(lr){2-3}\cmidrule(lr){4-5}
Backbone & Raw text & Sensory & Raw text & Sensory \\
\midrule
SASRec   & +0.00300 & \textbf{+0.00541} & +0.00071 & \textbf{+0.00246} \\
BERT4Rec & +0.00063 & \textbf{+0.00434} & +0.00036 & \textbf{+0.00284} \\
BSARec   & +0.00246 & \textbf{+0.00429} & +0.00142 & \textbf{+0.00311} \\
DIFF     & +0.00183 & \textbf{+0.00358} & +0.00072 & \textbf{+0.00181} \\
\bottomrule
\end{tabular}
\par}
\smallskip

\noindent
The structured channel yields the larger gain in all eight comparisons, four backbones by two metrics.
Unstructured text is not without effect: it improves every backbone over the ID-only baseline.
However, in every comparison, it recovers only part of the gain, and on BERT4Rec, it recovers almost none of it (+0.00063 compared with +0.00434 on HR@10).
This comparison indicates that the gain cannot be explained solely by the addition of textual information; structuring the item-side signal provides a consistent additional benefit.
Finally, every backbone in the main comparison carries an ID-only row, including DIFF (HR@10 0.09113, NDCG@10 0.05527), so no reported gain is measured against a baseline that already receives side information.
\end{revblock}

\subsection{Per-Facet Ablation}
\label{sec:facet_ablation}

\begin{revblock}
The five facets are not equally populated, and it is reasonable to ask whether the sparse ones contribute to the downstream gain.
We therefore rebuild the item-level bank five times per domain, each time removing one facet, and rerun the relational integration with everything else fixed.
Table~\ref{tab:facet_ablation} reports the change in HR@10 relative to the full bank in the same domain, so that a negative entry means the removed facet was contributing.

\par\smallskip
{\centering
\captionof{table}{Leave-one-facet-out with BSARec. Entries are $\Delta\Delta$HR@10 against the full-bank result for that domain, so negative values indicate a facet whose removal costs performance. The full-bank gain over the ID-only baseline is given in the header.}
\label{tab:facet_ablation}
\footnotesize
\setlength{\tabcolsep}{8pt}
\renewcommand{\arraystretch}{1.12}
\begin{tabular}{lccc}
\toprule
Removed facet & Beauty (+0.00447) & Games (+0.00391) & Grocery (+0.00238) \\
\midrule
Visual     & \textbf{$-$0.00188} & \textbf{$-$0.00181} & +0.00205 \\
Tactile    & \textbf{$-$0.00103} & $-$0.00050          & +0.00048 \\
Auditory   & $-$0.00013          & \textbf{$-$0.00142} & +0.00075 \\
Olfactory  & +0.00022            & $-$0.00041          & +0.00082 \\
Gustatory  & $-$0.00058          & +0.00004            & +0.00151 \\
\bottomrule
\end{tabular}
\par}
\smallskip

\noindent
Facet contributions vary by domain.
Removing the visual or tactile facet produces the largest decreases in Beauty, consistent with a domain in which color, finish, and texture dominate the described experience, whereas removing the visual or auditory facet has the largest effect in Games.
In particular, removing the auditory facet has a negligible effect in Beauty ($-$0.00013, within measurement noise) but reduces HR@10 by 0.00142 in Games, the second-largest effect in that domain.
This domain dependence motivates retaining sparse facets even when their contribution is small in some categories: it is the union of the five facets that allows a single schema to transfer across domains.
These results do not indicate that every facet is required in every domain.
As a robustness check we repeated the ablation with SASRec on Beauty, where every facet removal moves HR@10 by less than 0.0006, so the per-facet structure visible with BSARec is not uniformly recoverable across backbones.

Grocery provides a contrasting result.
Removing any individual facet does not reduce HR@10 there: all five removals are neutral to slightly positive, and the expectation that gustatory evidence would dominate a food domain is not borne out (removing it \emph{improves} HR@10 by 0.00151).
Grocery is also among the domains with the smallest overall gain from the sensory channel (+0.00238 HR@10 with BSARec); only Sports is smaller, at +0.00230.
That all five reduced banks slightly exceed the full bank indicates that no single facet is necessary for the aggregate gain in this domain; the present experiments do not identify the cause of this pattern, and we treat it as a limitation.
The Beauty result for the auditory facet has a documented mechanism, reported in Section~\ref{sec:recall_audit}: auditory content in Beauty reviews is frequently extracted but routed to a visual or tactile attribute, which depresses auditory coverage in that domain specifically.
\end{revblock}

\subsection{Component Ablation of the Bounded Relational Integration}
\label{sec:stage4_ablation}

\begin{revblock}
Stage~4 is the recommendation-time method, so we ablate its components directly using Beauty with BSARec, the frozen backbone checkpoint and final facet bank from the main table, and the same data split. For every variant, the residual weight \mbox{$\lambda_{\mathrm{rel}}$} and trust-region size \mbox{$N$} are reselected using validation NDCG@10 exactly as in the full method (seed 42).
The variants are: (A) the full integration; (B) no quality bias, removing the evidence-quality term from history attention; (C) no recency bias; (D) mean history pooling, replacing the learned facet-history attention with a uniform mean over valid history items while retaining the facet projections, facet weights, relational scoring, standardization, and trust region; (E) uniform facet weights in place of the learned ones; (F) no trust region, applying the same standardized residual to the entire candidate catalog; and (G) the frozen ID-only base.
Table~\mbox{\ref{tab:stage4_ablation}} reports the result.
\end{revblock}

\begin{table}[t]
\centering
\caption[Component ablation of the bounded relational integration]{\rev{Component ablation of the bounded relational integration on Beauty with BSARec (seed 42). Deltas are against the full integration (A). Differences smaller than 0.001 are within the reproduction noise measured across seeds and are not interpreted.}}
\label{tab:stage4_ablation}
\begin{revtable}
\begin{tabular}{lrrrr}
\hline
\textbf{Variant} & \textbf{HR@10} & \textbf{$\Delta$} & \textbf{NDCG@10} & \textbf{$\Delta$} \\
\hline
A. Full integration      & 0.0976 & --      & 0.0593 & --      \\
B. No quality bias       & 0.0976 & +0.0000 & 0.0594 & +0.0001 \\
C. No recency bias       & 0.0970 & $-$0.0006 & 0.0587 & $-$0.0005 \\
D. Mean history pooling  & 0.0964 & $-$0.0012 & 0.0590 & $-$0.0002 \\
E. Uniform facet weights & 0.0970 & $-$0.0006 & 0.0584 & $-$0.0009 \\
F. No trust region       & 0.0976 & +0.0000 & 0.0593 & +0.0000 \\
G. Base (ID-only)        & 0.0931 & $-$0.0045 & 0.0570 & $-$0.0023 \\
\hline
\end{tabular}
\end{revtable}
\end{table}

\begin{revblock}
A three-seed analysis, in which the full integration and the two variants with the largest effects (D and E) were rerun using seeds 42, 43, and 44, supports three observations.
First, the gain arises from the facet-level relational scoring itself: every variant remains at least 0.0033 above the ID-only base on HR@10, so no single refinement of the attention or weighting is responsible for the improvement.
Second, the individual attention refinements are within reproduction noise.
The only seed-42 difference exceeding 0.001, mean pooling at \mbox{$-0.0012$} HR@10, does not persist across seeds: over the three seeds the full integration reaches 0.0971\,$\pm$\,0.0005 and mean pooling reaches 0.0972\,$\pm$\,0.0009, so the experiment does not provide evidence that the learned history attention outperforms a uniform mean.
The learned facet weights are the only component whose removal is directionally consistent across all six seed-level comparisons: the full integration is higher in three of three seeds on both metrics, by +0.0007 HR@10 and +0.0005 NDCG@10 on average. The magnitude remains below our 0.001 resolution and constitutes a tendency rather than an established effect.
Third, removing the trust region leaves the result indistinguishable from the full integration (+0.0000 on both metrics): the trust region is a scale-control mechanism that bounds where the standardized residual may act, not a source of ranking accuracy, and we retain it for that reason.
Finally, this ablation does not evaluate polarity or negation: in the final Stage~4, polarity is stored in the facet bank to support auditing and inspection and does not enter recommendation-time scoring.
\end{revblock}

\subsection{Runtime Efficiency Analysis}
\label{sec:runtime_efficiency}

A key motivation of \textsc{ASER} is to avoid invoking an LLM or a text encoder during recommender training and
inference. The teacher LLM and the facet-aware student are used only during offline preprocessing to construct a
frozen item-level facet-bank table. At runtime, the downstream recommender consumes item IDs and the frozen bank is
consulted only to \rev{rescore} candidates that the recommender itself has already ranked.

\begin{revblock}
We describe the cost profile of the bounded relational integration of Section~\ref{sec:method_integration} in terms of the operations the design performs.

\textbf{Trainable parameters.} The sequential backbone is frozen and structurally unchanged, so the integration adds
\emph{zero} trainable parameters inside the recommender: the ID-only row and the ASER row of every cell of
Table~\mbox{\ref{tab:performance}} are produced by the same checkpoint. The only trainable component is the sensory metric,
which lives outside the recommender and holds $246{,}095$ parameters for $|\mathcal{F}|=5$, $D=768$ and $d_p=64$,
namely five $64 \times 768$ projections, five $64$-dimensional facet queries, and fifteen scalars (a quality scale, a
recency scale and a facet weight per facet). This is one metric per backbone--domain cell, and it is trained once,
offline, over frozen inputs.

\textbf{Storage.} The frozen facet bank is a precomputed cache rather than an online text encoder. Its dominant
component stores five canonical facet embeddings of 768 dimensions per item, so it scales as
$|\mathcal{I}| \times 5 \times 768$, plus per-facet confidence, polarity, and coverage scalars. The bank is loaded for
lookup and is not part of the trainable recommender.

\textbf{Runtime work.} Serving a query adds three operations to ordinary embedding-based scoring: a table lookup for
at most $50$ history entries in the bank, five $64 \times 768$ projections followed by one attention pass per
facet to form the user prototype, and $|\mathcal{F}|=5$ inner products per scored candidate. Only the
$N \le 100$ trust-region candidates need to receive the residual. No language model, teacher, student, or text encoder
is invoked at recommendation time, which is the efficiency property the design is meant to preserve.

\textbf{Scope.} The parameter and storage statements above are read directly from the released configuration and are
exact; per-query wall-clock latency and throughput were not benchmarked, so the runtime cost is characterized
structurally in terms of the operations listed above rather than through a timing measurement.
\end{revblock}
\FloatBarrier
\subsection{Limitations}
\begin{revblock}
The empirical findings are subject to several limitations.

\textbf{What the gain is attributable to.} The hit-rate improvement is not specific to sensory content. A control channel built with the same seed model, the same six-field schema, the same extraction and distillation code, and the same supervision budget, but a non-sensory vocabulary of fifty opinion descriptors with zero lexical overlap, matches our channel on HR@10 across nine comparisons (four backbones on Beauty, five domains with BSARec): the backbone axis splits two to two, the domain axis three to one with one tie, and most differences lie inside the reproduction variation of roughly 0.001 that we measure directly. The sensory-specific evidence is on ranking quality, where the sensory channel yields the larger gain in eight of the nine comparisons, with the one exception 0.00015 below the control and therefore inside reproduction noise. The gain is also not uniform across domains: in Grocery removing any individual facet does not reduce HR@10 under leave-one-facet-out even though the domain gains overall.

\textbf{Exposure concentration.} We measured exposure-side metrics for all 20 cells. The relational correction concentrates exposure: catalog coverage falls in 16 of 20 cells, the Gini coefficient of exposure rises in 16 of 20, and the long-tail share falls in 19 of 20 at cutoffs up to $K{=}20$ and in 18 of 20 at $K{=}50$. We report this as a negative result. Two observations qualify but do not negate this finding. First, exposure shifts toward the middle of the popularity distribution rather than toward the head: averaged over the 20 cells, the exposure share of the most popular 1\% of items falls, while the 21st-to-50th popularity band gains 2.58 percentage points of recommendation slots in the Beauty--BSARec cell. Second, an average of 30.0\% of the items in the reranked top 10 were not in the ID backbone's original top 10, so the lists do not stagnate. A deployment that is sensitive to catalog coverage would require an explicit diversity constraint, which we do not provide.

\textbf{Recall is LLM-audited and has no human anchor.} Our recall figures come from an independent language model that first enumerated sensory mentions in open vocabulary, blind to our schema and to all pipeline output, and only then mapped them to our taxonomy in a separate call. On a 310-item sample drawn from the seed annotation corpus, of which 293 items yielded at least one mention, it produced 3{,}070 mentions, of which the pipeline captures 49.9\% under a strict span-and-attribute match (95\% confidence interval 0.472 to 0.526), 54.1\% at facet level, and 90.7\% at item level, with only 1.2\% of mentions falling in a facet the pipeline missed entirely for that item. Three caveats bound these numbers. No human annotation was collected, so we cannot correct for the auditor's own bias and report the figures strictly as LLM-audited recall. The denominator is incomplete: a third enumeration pass still added 19.0\% new mentions, short of our 10\% saturation criterion, so the recall values are optimistic estimates rather than upper bounds. The auditor's own hallucination rate is 3.37\% and its self-consistency is 0.93. The largest identified source of attrition is our own design, namely the twelve-record per-item cap, whose miss rate is 0.528 at the cap against 0.386 below it.

\textbf{The schema omits a family of sensations.} Our thirteen-attribute inventory is precision-first and closed. Under unconstrained prompting on the seed annotation corpus, the seed model proposed attributes outside that inventory for 9.2\% of all extracted attributes (2{,}602 of 28{,}146), dominated by weight, comfort, and temperature, and the qualitative gap is a somatosensory family that our five facets do not host: burning, tingling, stinging, and warmth. These are exactly the descriptions that matter for categories such as topical cosmetics and spiced food, so the reported results should be read as covering the five classical facets rather than the full space of experiential description. The schema is extensible within the same framework, as the control channel demonstrates, but we do not evaluate an extended one.

\textbf{The upstream extraction pipeline is not bit-exactly reproducible.} The teacher LoRA adapters, the merged teacher weights, and the seed annotations are not redistributed: the seed annotations are outputs of a proprietary hosted language model, and the terms of use governing those outputs restrict redistribution of the annotations and of models derived from them. Appendix~\mbox{\ref{app:prompt_format}} documents the sensory schema and the seed-labeling prompt and request format, and Appendix~\mbox{\ref{app:training_details}} together with Table~\mbox{\ref{tab:teacher_lora_hparams}} reports the teacher LoRA configuration. All recommendation results reported in this work are reproducible from the released facet banks, recommender-side code, configurations, seeds, and evaluation scripts, and the released student checkpoint additionally supports generating facet banks for new item text. We therefore distinguish downstream recommendation reproducibility, which the release provides, from exact reconstruction of the upstream extraction pipeline, which it does not; anyone rebuilding the pipeline from raw text must substitute their own instruction-following model at the seed stage and would obtain a different, though not necessarily worse, supervision set.

\textbf{The threshold plateau is one-sided and the selected operating point lies near its edge.} We measured 14 operating points along three axes. For the bank-side presence gate, the gain remains between +0.0039 and +0.0050 HR@10 while bank coverage changes by 15\%, and the selected value is not the maximum. Thus, the reported number represents a point on a plateau rather than a tuned peak. For the extraction-side confidence gate, the response is flat on the permissive side but degrades monotonically on the restrictive side, losing 28\% of the relative HR@10 gain at a margin of +0.20, where teacher-record retention falls from 90.1\% to 57.5\%. The risk is therefore over-filtering, not under-filtering. Because teacher confidence scores are quantized, the gate is a step function, and a teacher with a different confidence calibration could enter the degrading regime without any change to the nominal threshold. Practitioners transferring this pipeline should check the retention rate rather than copy the threshold. The trust-region grid has the same one-sided character: the two selected values of $\lambda_{\mathrm{rel}}$ are the two largest offered, and $N=100$ is the grid maximum. Thus, the reported operating point lies at the upper edge of the search space.

\textbf{The bank is built from the evaluation catalog's own review text.} The frozen facet bank of a domain is produced by encoding the item-side review text of that domain, and under leave-one-out a share of that text belongs to the interactions being predicted: in Beauty, 22.5\% of the review texts entering the bank are some user's test or validation interaction. The encoder sees no user identifier, no interaction order, and no label, so the setting remains the static item-content setting of Section~\ref{sec:method}, and the temporal control of Section~\ref{sec:temporal_control} constrains supervision rather than this text. We did not run a bank-side control in which held-out reviews are withheld from bank construction, so the size of this exposure is disclosed rather than measured, and a deployment that must exclude it would rebuild the bank from interactions preceding the prediction point.

\textbf{Most cells use a single seed.} Seventeen of the 20 domain--backbone cells are reported from a single seed. Three representative cells, BSARec on Beauty, Toys, and Video Games, were run with three seeds each: all nine runs improve both metrics, with seed-to-seed standard deviations of at most 0.0010 in HR@10 and 0.0004 in NDCG@10, in every case below the corresponding mean effect. This supports the direction of the result but not fine-grained comparisons, and differences on the order of 0.001 in HR@10 between single-seed configurations should not be interpreted as meaningful.

\textbf{Scope of the evaluation and cross-model comparison.} In-domain evaluation covers five English-language Amazon 2014 categories under 5-core filtering. Absolute values in this paper are comparable only within this shared experimental setup: every model is evaluated using one preprocessing pipeline and one full-catalog ranker, so these values should not be compared directly with published values obtained using different preprocessing or sampled-negative evaluation. Negative sampling appears only in the training loss of the causal backbones; no evaluation result in this paper uses sampled negatives.
\end{revblock}

\section{Conclusion}

We presented \textsc{ASER}, a sensory-aware sequential recommendation framework that connects review language with item-side representation learning.
Instead of using review text as an unconstrained embedding source, \textsc{ASER} extracts structured sensory attributes from product reviews, distills them into a compact facet-aware student encoder, and \rev{integrates} the resulting sensory embeddings into standard sequential recommenders.
This design allows recommendation models to use information about how products look, feel, smell, taste, or sound, while avoiding LLM inference at recommendation time.

The main contribution of this work is threefold.
First, we introduce a sensory-only extraction schema that converts unstructured review text into evidence-grounded attribute--value records with polarity, negation, and confidence metadata.
Second, we propose a facet-aware student distillation framework that decomposes teacher supervision into visual, tactile, auditory, olfactory, and gustatory facets, enabling reusable item-level sensory representations.
Third, \rev{we show that these representations can be integrated into diverse sequential recommenders through a bounded relational mechanism that leaves the backbone frozen and applies a validation-selected, magnitude-bounded correction, without changing the backbone architecture.}
\rev{Across five Amazon domains and four representative backbones, all trained within a common experimental pipeline and evaluated under one leave-one-out full-catalog protocol, the augmented models improve over their matched non-sensory counterparts in all 20 domain--backbone pairs for both HR@10 and NDCG@10 and in all 120 reported metric cells, with average relative gains of 6.1\% in HR@10 and 6.4\% in NDCG@10.}
\rev{These results indicate that review-derived attribute representations provide complementary information beyond matched non-sensory inputs and behavioral interaction patterns. They do not, by themselves, establish that the complementary information is sensory.}

Our analyses further clarify when and why sensory augmentation is useful.
\rev{Matched text controls show that the gains do not come merely from adding text-derived features: unstructured embeddings of the same review text are weaker than the structured representation on every backbone and metric we compare.} \rev{A matched control channel, constructed with the same seed model, schema, pipeline, and supervision budget but a non-sensory vocabulary, then separates the claim by metric. On HR@10 the two channels are indistinguishable across nine backbone and domain comparisons, with most differences inside the reproduction variation we measure, so the hit-rate gain is attributable to structured, evidence-grounded extraction rather than to sensory content. On NDCG@10 the sensory channel yields higher values in eight of the nine matched comparisons, the single exception falling an order of magnitude below the run-to-run variability, indicating that the sensory-specific evidence is concentrated in ranking quality.}
\rev{The component ablation of the integration stage shows that the gain arises from the facet-level relational scoring itself, and that the trust region functions as scale control: the sensory correction is expressed in units of the base score and reorders only candidates the backbone already considers plausible, so noisy sensory evidence cannot displace the collaborative ranking.}
\rev{Thus, \mbox{\textsc{ASER}} improves ranking performance and additionally supports auditability through facet-level representations derived from evidence-grounded supervision, subject to the caveat that this evidence is audited but incomplete.}

More broadly, this work demonstrates that structured outputs from language models can serve as an intermediate representation layer for recommendation systems.
\rev{Rather than using LLMs as online recommenders or opaque text encoders, \mbox{\textsc{ASER}} uses them to create auditable supervision that can be distilled into efficient item-side embeddings: on the Beauty evaluation catalog, 94.8\% of retained records are supported by the evidence span they cite, and an independent, open-vocabulary LLM audit finds that the pipeline captures 90.7\% of item-level sensory content.}
Future work can extend this idea to richer attribute spaces, user-facing explanation interfaces, and feedback mechanisms where users express preferences over specific sensory or functional attributes.
\rev{Another promising direction is to raise extraction recall rather than precision alone. The same audit finds that only 49.9\% of individual sensory mentions are captured under a strict span-and-attribute match, and the largest single source of attrition is our own per-item record cap, so verification or multi-pass labeling should be paired with a larger extraction budget while preserving evidence grounding.}
\rev{To support reproducibility, we release the accompanying code and research data through public repositories, as detailed in the Data availability statement.}
\rev{Overall, \mbox{\textsc{ASER}} shows that distilling structured, evidence-grounded attributes from reviews is a practical way to make sequential recommendation more semantically grounded and auditable, and that within that family the sensory channel provides a consistent advantage in ranking quality together with a direct connection to perceptual product descriptions.}

\section*{Data availability}
\rev{The released artifacts are organized according to their role in reproducibility. The facet banks are the primary artifacts, because they are the objects directly consumed by the recommender; the student checkpoint enables the construction of new banks; and the teacher LoRA adapters, the merged teacher weights, and the seed annotations are not redistributed, because they derive from outputs of a proprietary hosted model whose terms of use restrict redistribution.}

The datasets used in this study are publicly available from the Amazon Reviews benchmark collections. \rev{To support reproducibility, the implementation, configurations, and experiment
scripts are released in a public GitHub repository}%
\footnote{Code repository: \url{https://github.com/ycyoon/sensory-seqrec-release}}%
\rev{, and the research data associated with this study are archived separately}%
\footnote{Research data archive: \url{https://doi.org/10.5281/zenodo.22008657}}\rev{.}
\rev{The archived artifacts carry SHA-256 checksums recorded in a manifest, together with a script that verifies them.}

\section*{CRediT authorship contribution statement}

Yeo-Chan Yoon: Conceptualization, Methodology, Software, Validation,
Formal analysis, Writing -- original draft, Writing -- review and editing, Supervision.

Chanjun Park: Methodology, Validation, Writing -- review and editing.

Kyuhan Koh: Investigation, Validation, Writing -- review and editing.

\section*{Declaration of competing interest}

The authors declare that they have no known competing financial interests or
personal relationships that could have appeared to influence the work reported
in this paper.

\section*{Declaration of generative AI and AI-assisted technologies in the manuscript preparation process}

During the preparation of this work, the authors used ChatGPT and Claude to check for factual inaccuracies, evaluate logical consistency, and translate text between Korean and English. After using these tools, the authors reviewed and edited the content as needed and take full responsibility for the content of the published article.

\ifshowauthors



\fi

\bibliographystyle{model1-num-names}

\bibliography{cas-refs}



\clearpage
\appendix
\section*{Appendix}
\addcontentsline{toc}{section}{Appendix}

\section{Teacher JSON Output Example}
\label{app:json_example}

This section provides an example teacher output in the sensory JSON schema used throughout the paper.
The example illustrates the six-field record format for each extracted attribute and shows how evidence and negation are represented.

\begin{lstlisting}[caption={Teacher output example in sensory JSON format.},label={lst:json_example}]
{
  "items": [
    {
      "id": 0,
      "attributes": [
        {
          "attribute": "color",
          "value": "matte black",
          "evidence": "matte black finish",
          "polarity": "positive",
          "negated": false,
          "confidence": 0.93
        },
        {
          "attribute": "sound",
          "value": "quiet click",
          "evidence": "quiet click when pressed",
          "polarity": "positive",
          "negated": false,
          "confidence": 0.88
        },
        {
          "attribute": "scent",
          "value": "none",
          "evidence": "odorless",
          "polarity": "unknown",
          "negated": true,
          "confidence": 0.91
        }
      ]
    }
  ]
}
\end{lstlisting}

\section{Prompt and Input Format for Sensory Extraction}
\label{app:prompt_format}
\label{app:prompt_guidelines}

We use GPT-5 Mini to generate seed labels for sensory extraction.
The teacher is trained to imitate this behavior.

\subsection{System Prompt}
\label{app:prompt_system}

\begin{lstlisting}[caption={System prompt used for GPT-5 Mini seed labeling.},label={lst:prompt_system}]
You extract only sensory relevant product attributes for a recommender system.
Return only valid JSON and do not add any explanation text.
Sensory relevant means how the product looks, feels, smells, tastes, sounds, or the immediate user experience.
Do not extract non sensory information such as brand, compatibility, contents lists, age ranges, player counts, country of origin, warranty, or shipping.

The attribute field must be one of:
color, pattern, shape, graphics, brightness, glossiness, transparency, finish, design,
texture, scent, flavor, sound.

The value field is open vocabulary and should be short and concrete.
Keep each value short, ideally six words or fewer.
The evidence field should copy a short span from the input text and should be at most 120 characters.
If longer, truncate.

The polarity field is one of:
positive, negative, neutral, unknown.

For explicit absence cases such as odorless, unscented, or no smell, set negated to true, set value to none, and set polarity to unknown.


If no sensory attributes are present, return an empty list.

Output schema:
{"items":[{"id":0,"attributes":[{"attribute":...,"value":...,"evidence":...,"polarity":...,"negated":...,"confidence":0..1}]}]}
\end{lstlisting}

\subsection{Batched User Request Format}
\label{app:prompt_request}

\begin{lstlisting}[caption={Batched request format for teacher inference.},label={lst:prompt_request}]
{
  "items": [
    {
      "id": 0,
      "text": "Title line\nCategory line\nDescription line\nReview 1\nReview 2\n..."
    }
  ],
  "rules": {
    "max_attributes_per_item": 20,
    "drop_vague": true,
    "avoid_duplicates": true
  }
}
\end{lstlisting}

\FloatBarrier
\section{Student and Teacher Training Details}
\label{app:training_details}
\label{app:student_teacher_details}

To improve reproducibility, we report the implementation details of the facet-aware student and the teacher LoRA fine-tuning setup.
The student is implemented as \rev{a facet-aware model} that consists of a DeBERTa-v3-Small encoder and additional facet-specific adapters and prediction heads.
\begin{revblock}
The encoder checkpoint is \texttt{microsoft/deberta-v3-small}, and the frozen target encoder used to build the dense supervision targets of Section~\ref{sec:method_supervision} is the sentence encoder \texttt{sentence-transformers/all-mpnet-base-v2} (\url{https://huggingface.co/sentence-transformers/all-mpnet-base-v2}), which is used only during supervision construction and is not invoked at bank-construction or recommendation time.
\end{revblock}
Table~\ref{tab:student_model_size} summarizes the measured parameter counts, and Table~\ref{tab:student_training_hparams} reports the loss weights, thresholds, and optimization hyperparameters used in student training.
The facet weights in Table~\ref{tab:student_training_hparams} are global
loss-mixing coefficients for the shared all-domain student, not
inverse-frequency class weights. Positive/negative imbalance in evidence
detection and presence prediction is already addressed by focal BCE and
weighted BCE. We therefore do not set $\lambda_f$ by inverse frequency.
Instead, the weights are chosen to stabilize multi-domain optimization.
Visual and tactile facets receive larger weights because they are broadly
observed across domains and provide stable supervision. Olfactory receives
an intermediate weight because it is informative in Beauty and Grocery but
less frequent elsewhere. Auditory and gustatory receive conservative
global weights because auditory supervision is sparse and often ambiguous,
whereas gustatory supervision is highly concentrated in Grocery.
Up-weighting these facets globally can bias the shared encoder toward
domain-specific lexical patterns or introduce high-variance gradients.
Domain-adaptive facet weighting is left for future work.
In particular, gustatory evidence is dominant in Grocery but much rarer elsewhere, so we use a conservative global weight to avoid over-specializing the shared student to a single domain.
Table~\ref{tab:teacher_lora_hparams} reports the LoRA configuration used for the teacher model.

\begin{table}[htbp]
\centering
\small
\setlength{\tabcolsep}{5pt}
\renewcommand{\arraystretch}{1.08}
\caption{Measured parameter counts of the facet-aware student.}
\label{tab:student_model_size}
\begin{tabular}{lr}
\toprule
Component & Parameters \\
\midrule
Facet-aware student encoder & 154{,}715{,}417 \\
Encoder & 141{,}304{,}320 \\
Adapters and prediction heads & 13{,}411{,}097 \\
\bottomrule
\end{tabular}
\end{table}

\begin{table}[htbp]
\centering
\small
\setlength{\tabcolsep}{5pt}
\renewcommand{\arraystretch}{1.08}
\caption{Facet-aware student loss and optimization hyperparameters.
The facet weights $\lambda_f$ are global loss-mixing coefficients for
all-domain student training, not inverse-frequency class weights.}
\label{tab:student_training_hparams}
\begin{tabular}{lll}
\toprule
Hyperparameter & Value & Description \\
\midrule
$\lambda_{\mathrm{visual}}$ & 1.0 & Facet loss weight \\
$\lambda_{\mathrm{tactile}}$ & 1.0 & Facet loss weight \\
$\lambda_{\mathrm{auditory}}$ & 0.4 & Facet loss weight \\
$\lambda_{\mathrm{olfactory}}$ & 0.8 & Facet loss weight \\
$\lambda_{\mathrm{gustatory}}$ & 0.4 & Facet loss weight \\
$\tau_{\mathrm{pres}}$ & 0.5 & Presence threshold \\
Focal BCE $\alpha$ & 0.25 & Token evidence loss \\
Focal BCE $\gamma$ & 2.0 & Token evidence loss \\
weighted BCE positive weight & 1.0 & Presence loss \\
weighted BCE negative weight & 0.35 & Presence loss \\
Phase A epochs & 2 & Evidence localization and presence training \\
Phase B epochs & 4 & Full multi-task training \\
Batch size & 64 & Student training \\
Encoder learning rate & $2\times10^{-5}$ & Student encoder \\
Head learning rate & $1\times10^{-4}$ & Adapters and prediction heads \\
Warmup ratio & 0.1 & Linear warmup \\
\bottomrule
\end{tabular}
\end{table}

For the facet-presence objective, we use a weighted BCE:
\[
\mathcal{L}_{\mathrm{pres}}
=
- w_{+} y \log p
- w_{-} (1-y) \log (1-p),
\]
where $w_{+}=1.0$ and $w_{-}=0.35$.
This implementation uses positive and negative class weights to control asymmetry.

\begin{table}[htbp]
\centering
\small
\setlength{\tabcolsep}{4pt}
\renewcommand{\arraystretch}{1.08}
\caption{Teacher LoRA fine-tuning configuration.}
\label{tab:teacher_lora_hparams}
\begin{tabular}{@{}p{0.30\columnwidth}p{0.22\columnwidth}p{0.38\columnwidth}@{}}
\toprule
Hyperparameter & Value & Description \\
\midrule
LoRA rank $r$ & 16 & Low-rank adaptation dimension \\
LoRA alpha & 32 & LoRA scaling factor \\
LoRA dropout & 0.05 & LoRA dropout rate \\
Target modules &
\begin{tabular}[t]{@{}l@{}}
\texttt{q\_proj}, \texttt{k\_proj}, \texttt{v\_proj} \\
\texttt{o\_proj}, \texttt{gate\_proj} \\
\texttt{up\_proj}, \texttt{down\_proj}
\end{tabular}
& Attention and feed-forward projection modules \\
\bottomrule
\end{tabular}
\end{table}

The student is trained in two phases.
Phase A emphasizes evidence localization and facet-presence prediction so that the model first learns to identify grounded sensory cues.
Phase B then optimizes the full multi-task objective, including facet presence, polarity, confidence regression, and facet-embedding alignment.
For the token-level evidence objective, we use focal binary cross-entropy with $\alpha=0.25$ and $\gamma=2.0$ to address the sparsity of positive evidence tokens.
For facet presence, we use a weighted binary cross-entropy loss with positive and negative class weights.

\clearpage

\end{document}